\documentclass[journal]{IEEEtran}
\ifCLASSINFOpdf
   \usepackage[pdftex]{graphicx}
   \graphicspath{{../Figures/}}
   \DeclareGraphicsExtensions{.pdf,.jpeg,.png}
\else
\fi
\usepackage[table]{xcolor}

\usepackage{amsmath}
\usepackage{array}

\ifCLASSOPTIONcompsoc
 \usepackage[caption=false,font=normalsize,labelfont=sf,textfont=sf]{subfig}
\else
  \usepackage[caption=false,font=footnotesize]{subfig}
\fi

\usepackage{stfloats}
\usepackage{tabularx,booktabs}

\usepackage{multirow}

\usepackage{amssymb}
\usepackage{array}
\usepackage{xcolor}
\usepackage{cite}
\begin{document}

\title{Attend and Guide (AG-Net): A Keypoints-driven Attention-based Deep Network for Image Recognition}
%
%
%

\author{Asish~Bera$^\ast$,~\IEEEmembership{~Member,~IEEE},
        Zachary~Wharton$^\ast$, 
        Yonghuai~Liu,~\IEEEmembership{~Senior Member,~IEEE}, \\
        Nik~Bessis,~\IEEEmembership{~ Senior Member,~IEEE},
        and~Ardhendu Behera$^\dagger$,~\IEEEmembership{~Member,~IEEE}
\thanks{A. Bera, Z. Wharton, Y. Liu, N. Bessis and A. Behera are with the Department of Computer Science, Edge Hill University, UK.}
\thanks{$\ast$ Both authors contributed equally to this work}%
\thanks{$\dagger$ Corresponding author, beheraa@edgehill.ac.uk}%
\thanks{Manuscript received Month 00, 2020; revised Month 00, 2020.}}
\markboth{Journal of \LaTeX\ Class Files,~Vol.~00, No.~0, April~2020}%
{Shell \MakeLowercase{\textit{et al.}}: Bare Demo of IEEEtran.cls for IEEE Journals}
\maketitle
\begin{abstract}
This paper presents a novel keypoints-based attention mechanism for visual recognition in still images. Deep Convolutional Neural Networks (CNNs) for recognizing images with distinctive classes have shown great success, but their performance in discriminating fine-grained changes is not at the same level. We address this by proposing an end-to-end CNN model, which learns meaningful features linking fine-grained changes using our novel attention mechanism. It captures the spatial structures in images by identifying semantic regions (SRs) and their spatial distributions, and is proved to be the key to modelling subtle changes in images. We automatically identify these SRs by grouping the detected keypoints in a given image. The ``usefulness'' of these SRs for image recognition is measured using our innovative attentional mechanism focusing on parts of the image that are most relevant to a given task. This framework applies to traditional and fine-grained image recognition tasks and does not require manually annotated regions (e.g. bounding-box of body parts, objects, etc.) for learning and prediction. Moreover, the proposed keypoints-driven attention mechanism can be easily integrated into the existing CNN models. The framework is evaluated on six diverse benchmark datasets. The model outperforms the state-of-the-art approaches by a considerable margin using Distracted Driver V1 (Acc: 3.39\%), Distracted Driver V2 (Acc: 6.58\%), Stanford-40 Actions (mAP: 2.15\%), People Playing Musical Instruments (mAP: 16.05\%), Food-101 (Acc: 6.30\%) and Caltech-256 (Acc: 2.59\%) datasets.
\end{abstract}


\IEEEpeerreviewmaketitle

\section{Introduction}
%
\IEEEPARstart{V}{isual} recognition using still images is a challenging problem and is widely studied by computer vision researchers \cite{YaoKL11, guo2014survey}. Recent advancements of Convolutional Neural Networks (CNN) are very successful in achieving high accuracy in image-based action recognition \cite{ZhaoMY16, YanSLZ18, Lavinia2019NewCF}, object detection \cite{he2017mask}, machine translation \cite{BahdanauCB14}, and other multimedia content analysis tasks. Though impressive solutions have been devised using deep models for human action recognition, it is yet a challenging task to discriminate various fine-grained activities like playing a violin vs a guitar, using a phone vs talking to passengers while driving, etc. It can be regarded as a more challenging problem when multiple actions appear in a single image, such as walking and talking over the phone \cite{TWang2019}. Existing approaches on visual image categorization are often based on the contextual information within a region enclosing the person/object in focus \cite{GkioxariGM15, gkioxari2015actions, mallya2016learning}. A major limitation is that pre-annotated bounding-boxes or object/part detectors are essential to determine the region of interest. Moreover, region-based approaches are also adapted with CNN to improve action recognition accuracy \cite{ZhaoMY16}, and fine-grained visual recognition \cite{JiangMLL20}.

Before the deep learning era, visual recognition methods are dominated by the keypoints-driven engineered features such as the Scale Invariant Feature Transform (SIFT) \cite{Lowe04}, Speeded-Up Robust Features (SURF) \cite{bay2006surf}, etc. Generally, these keypoints often refer to salient locations in the image and are invariant to image rotation, scale, translation, illumination, distortion and so forth. These hand-crafted feature representation techniques are widely investigated along with the bag-of-words and discriminative machine learning approaches until the great success has been attained with the deep architectures \cite {sharma2012discriminative}, \cite{sun2009action}. 
These robust and invariant local descriptors have shown expressive feature representation capability, influencing high recognition accuracy. These keypoints-driven methods have ruled the world of computer vision before the recent advances and popularity of deep CNNs. As a result, modern research direction is focused on the deep feature extraction techniques due to their more powerful representation and generalization capabilities for visual recognition involving large datasets \cite{gu2018recent}. Nevertheless, we find that earlier keypoints-driven works are still helpful in guiding the modern CNNs in achieving higher accuracy. Thus, our aim is to get the best of both approaches i.e. to further enhance the power of deep models by guiding them using salient regions, defined by traditional salient keypoints in a bottom-up fashion. For a given image, the goal is to identify a set of semantic regions (SRs) and their importance to guide the potential of deep feature representation mechanism, resulting in improved recognition performance. 
\begin{figure*}[h]
  \centering
\includegraphics[width=0.58\textwidth] {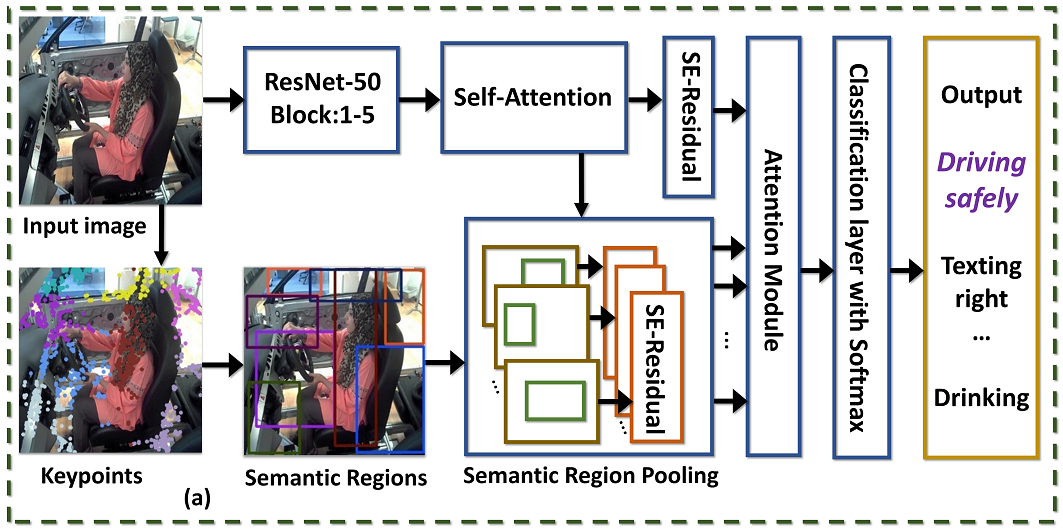}
\includegraphics[width=0.34\textwidth]  {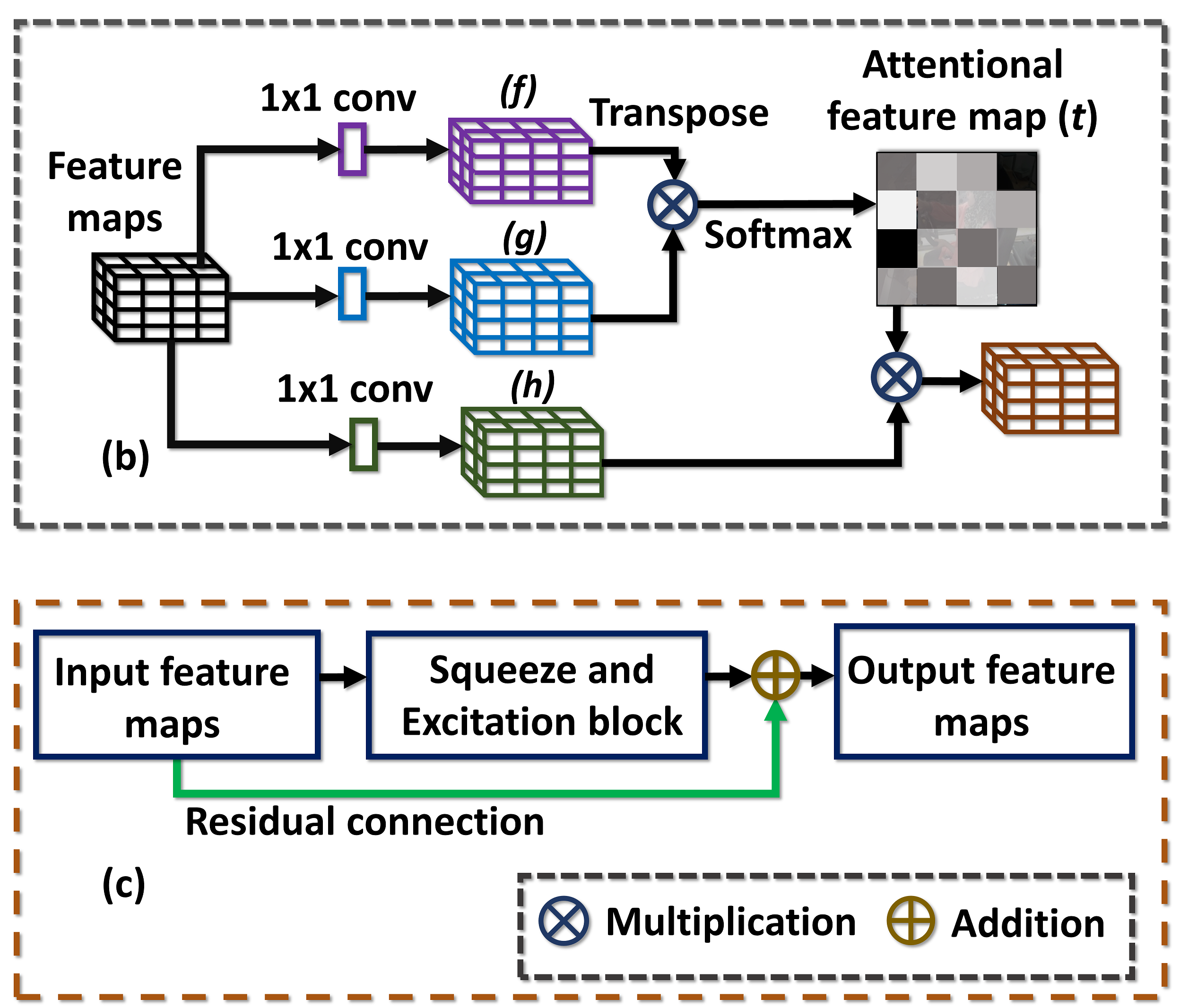}
\caption{Proposed keypoint-driven attention based visual recognition model (AG-Net): (a) Used for recognizing different fine-grained activities in still images (e.g., the input image is classified as \emph{driving safely} activity). (b) Detailed Self-Attention block adapted from SAGAN \cite{H.Zhang2018}. The $\otimes$ denotes matrix multiplication operation, and Softmax operation is applied row-wise. (c) The proposed residual connection (SE-Residual denoted with $\oplus$) to the Squeeze-and-Excitation block used for pooling features from semantic regions (SRs).}
 \label{fig:teaser}
 \vspace{-1em}
\end{figure*}

The advent of attention mechanisms has proven their superior performances in machine translation, object detection, image captioning, and other tasks \cite {BahdanauCB14,VaswaniSPUJGKP17}. It is inspired by the human visual search mechanism and emphasizes on crucial regions/features, which are more informative rather than considering the whole scene since the global image-context may incorporate background noise that causes performance degradation. The use of SRs has been explored in \cite{GkioxariGM15, gkioxari2015actions, mallya2016learning} to improve human action recognition by detecting and localizing various body parts and/or objects in an image. However, the application of attention mechanism is relatively less explored in literature where several SRs influence learnable contextual cues for fine-grained visual recognition. The proposed approach innovatively addresses this via a novel deep CNN that incorporates the attention mechanism into the identified SRs to advance the visual recognition task. 

Our model is called attend and guide network (AG-Net) and various key components of the network are shown in Fig. \ref{fig:teaser}a, illustrating the conceptual steps in recognizing safe-driving activity. It does not require region annotations and/or off-the-shelf object/part detectors to facilitate feature learning task. Our approach automatically finds significant SRs based on salient SIFT-keypoints \cite{Lowe04}. These keypoints identify distinct spatial locations of the image, which are useful to generate SRs. Our SRs contain substantial contextual information at different granularity. They are passed through attentional maps for identifying their importance in recognizing images. We leverage that regional information provides useful contextual cues influencing various image recognition tasks, particularly, driving-related distractions activities \cite{AbouelnagaEM17, H.M.Eraqi, BeheraK18, BahetiGT18}, daily/regular human actions (Stanford-40) \cite {YaoJKLGF11, liu2018loss, LiuGQWL19 }, people playing musical instruments (PPMI-24) \cite {YaoF10,YanSZ17, ZhangLPXS16, Hu2013recognising}, food classification (Food-101) \cite{JiangMLL20, BossardGG14,MinJLRJ19}, and  extremely diverse and more general object categories (Caltech-256) \cite {griffin2007caltech,zhang2019multi,ge2019weakly}. The major contributions of this paper are summarized as follows: 
\begin{enumerate}
    \item A novel method for generating SR proposals has been presented to attain local to global contextual information involving smaller patches to larger patches to the whole image. The SRs are formed and localized using salient keypoints and a Gaussian Mixture Model (GMM). 
\item A novel attention module has been proposed to efficiently recognize images from these SRs by \textit{learning to attend} each SR by its importance towards classification decision. It can be easily integrated with the state-of-the-art (SotA) CNNs to improve their performance.
\item The proposed model has been trained in an end-to-end fashion and the comprehensive experimental results on six diverse benchmark datasets show significant improvements over the SotA approaches.
\end{enumerate}
The rest of this paper is organized as follows. Section \ref{rel_work} summarizes related works on visual recognition using still images. Section \ref{proposed} describes the proposed framework. The datasets used for experiments are briefed in Section \ref{datasets}. The experimental results are presented and analyzed in Section \ref{experiments}, and an in-depth ablation study is discussed in \ref{Ablation}. Finally, the conclusion is drawn in Section \ref{conclusion}.
\section{Related Work} \label{rel_work}
Several existing methods follow holistic cues for visual recognition, including deep CNNs. Recently, attention-based mechanisms have been applied in several different ways to improve the performance of the existing deep CNNs. We have summarized the related studies linking the proposed approach. 
\subsection {Human Action Recognition} 
Contexts, body-pose, body-parts, and recent attention-based techniques have shown promising successes in human action recognition from still images. The context-based approach incorporates the person interacting with an object, generally, specified within the candidate region(s).
The R*CNN model \cite {GkioxariGM15} describes that action in different candidate parts contains informative contextual cues for making a decision. A set of secondary region proposals with pre-annotated bounding-boxes are used to learn regional contexts. Deep features of a region are combined with the Vector of Locally Aggregated Descriptors (VLAD) encoding technique in \cite{YanSZ17}. It identifies local and global contexts in still images. In \cite{ZhangLPXS16}, VLAD is applied to learn local features, and generalized max-pooling is incorporated for action classification. In this regard, the VLAD \cite {ZhangLPXS16} and saliency map \cite {sharma2012discriminative} based approaches extract SIFT-based local descriptors. On the contrary, Zhang et al. \cite{4CWCDL16} validate that input bounding-box is not crucial to determine an action. Instead, an action mask is generated in their proposal. A pose-object interaction exemplar is configured in \cite{Hu2013recognising} to describe an action where human-object interaction takes place. Similarly, in \cite{yao2012recognizing}, the mutual context between objects and human poses is explored in such a manner that an object can facilitate the recognition of an action. 
In the other direction, action recognition using the various body-parts is emphasized due to their discriminative effectiveness. Gkioxari et al. \cite{GkioxariGM14} consider three body parts, along with a provided bounding-box regression. The keypoints i.e., predefined landmarks on each of the body-part are clustered into poselets for part-action detection. Zhao et al. \cite{ZhaoMY16} describe semantic body-part actions from a single image. The mid-level semantic part actions are defined from seven body parts, which are useful for action recognition by a part action network. Recently, visual attention based deep models are developed for action recognition from still images. An attention network is presented in \cite{YanSLZ18} with the scene-level and region-level contextual cues along with a target person's bounding-box. The context-aware appearance features play a substantial role to modulate attention in \cite{TWang2019}.
\subsection {Driving Action Recognition} 
Driver action recognition is a subset of human action/activity recognition. It is a challenging task for intelligent vehicles to monitor 
various secondary activities (e.g. texting messages, eating or drinking), which are bound to happen during autonomous driving when vehicles are in control. This is also applicable for developing intelligent features for Advanced Driver-Assistance Systems (ADAS) for monitoring safe driving activities. Zhao et al. \cite{ZhaoIET12}  propose a method that uses the contourlet transform and random forest classifier for recognizing four activities (safe driving, operating shift lever, eating, and talking on the phone). In \cite{XingLZWNCVW18}, seven driving activities are experimented using a feed forward neural network. Abouelnaga et al. \cite{AbouelnagaEM17} develop a genetically weighted ensemble (GAWE) method for recognizing various driving activities. It consists of five deep networks and is computationally expensive for self-driving cars due to the limited computational capacity in embedded devices. To alleviate network heaviness, a simplified deep architecture is presented to improve accuracy in \cite{BahetiGT18}. The DenseNet 
is adapted in in \cite{BeheraK18} using latent body poses to distinguish secondary in-vehicle driver activities. The latent poses are extracted through the part affinity fields, which is pre-trained on the MS-COCO dataset. A fusion of three deep architectures, namely, ResNet, 
Inception module, and hierarchical recurrent neural network is presented in \cite{Alotaibi2019} to assess the performance of driver’s activity.  
\subsection {Food and Generic Object Recognition} 
Food category discrimination is a fine-grained visual recognition task  \cite{LiuXWL16, MinJLRJ19, CuiSSHB18, MartinelFM18, PandeyDMP17}. An ensemble of the AlexNet, ResNet, and GoogLeNet is proposed in \cite{PandeyDMP17}. A pipeline parallelism framework, namely GPipe performs well using the AmoebaNet-B \cite{GPipeFood}. It employs a batch splitting pipeline to enhance efficiency for a large-scale network. In WIde-Slice Residual Network (WISeR) \cite{MartinelFM18}, a slice convolution layer is introduced to represent the vertical trait of food. In addition, a large residual learning method has been followed for generic food classification. Recently, a multi-scale fusion and multi-view feature aggregation (MSMVFA) method is proposed in \cite {JiangMLL20} by considering contextual information of ingredient to aggregate high-level semantic features, mid-level attributes, and deep visual features into a unified representation. In \cite{CuiSSHB18}, visual similarity is measured between source and target domain using the Earth Mover’s Distance to improve transfer learning performance. Transfer learning is applied to the source domain and calibrated on target domain for classification. In \cite{CuiZWLLB17}, a deep kernel pooling framework is implemented to capture higher-order feature representation in the form of kernels. A Fully Convolutional Attention Network (FCAN) is presented for fine-grained visual recognition in \cite{LiuXWL16}. The attention module is formulated into a Markov decision process for rewarding estimation at each step of attention. 

For the generic image classification, a codebookless model is proposed in \cite{ wang2016towards} as an alternative to the bag-of-words based methods. It uses a single Gaussian model to represent the whole image, and then a two-step metric is applied for matching Gaussian models. The advancement of CNNs has significantly improved the recognition accuracy of large-scale image classification \cite{szegedy2016rethinking,zoph2018learning,he2016deep,huang2017densely,simonyan2014very}. To investigate the functionality of different intermediate layers in the CNNs, a visualization-based method is presented that provides insight into individual feature maps \cite{zeiler2014visualizing}. It uses a multi-layer deconvolutional network to map feature activation back to the input image pixels. A coherent parameter regularization approach is described in \cite{xuhong2018explicit} for transfer learning using ResNet architecture for generic object recognition. The $L^2$ regularizer effectively improves the performance for inductive transfer-learning task. In \cite {ge2019weakly}, Mask R-CNN and conditional random field are utilized for object detection and instance segmentation in a weakly supervised manner. A bidirectional LSTM is used for complementary context-encoding from selected part proposals. A multi-view image classification approach  using the visual, semantic and view consistency is presented in \cite{zhang2019multi}. It linearly combines the outputs of multiple views to enhance classification accuracy.  

\begin{figure*}
\centering
\subfloat[Generic image recognition consisting distinctive categories]{ 
\begin{minipage}[c][1\width]{0.23\textwidth}
	   \centering
\includegraphics[width=0.48\textwidth] {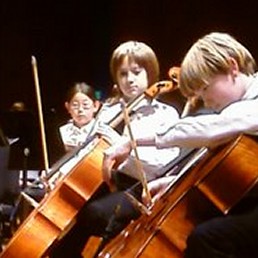} 
\includegraphics[width=0.48\textwidth]  {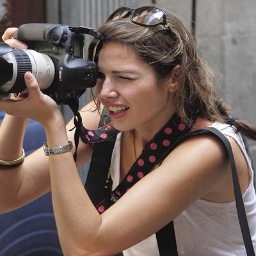}\\ \vspace{1mm}
\includegraphics[width=0.48\textwidth]  {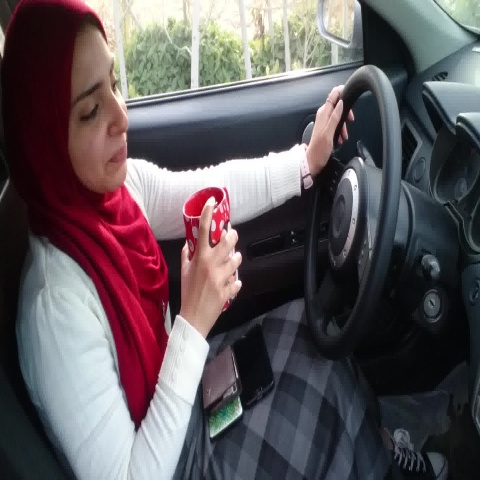}
\includegraphics[width=0.48\textwidth]  {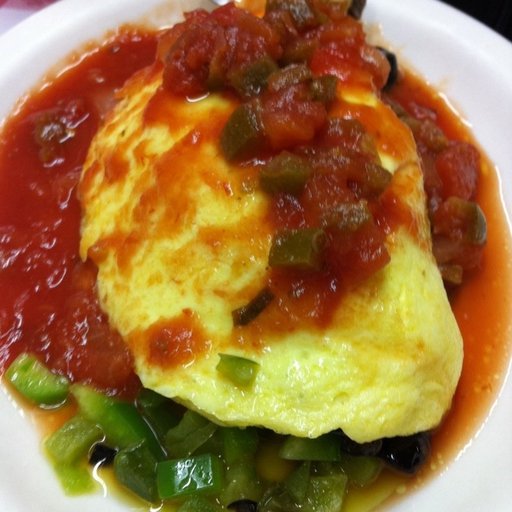}
	\end{minipage}
	}\hfill
\subfloat[Different fine-grained driving activities performed by the same person]{ 
\begin{minipage}[c][1\width]{0.23\textwidth}
	   \centering
\includegraphics[width=0.48\textwidth] {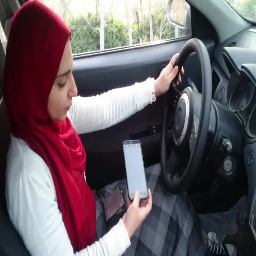} 
\includegraphics[width=0.48\textwidth]  {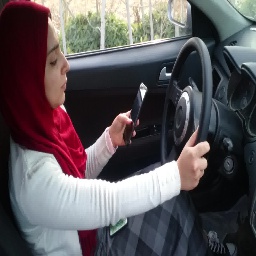}\\ \vspace{1mm}
\includegraphics[width=0.48\textwidth]  {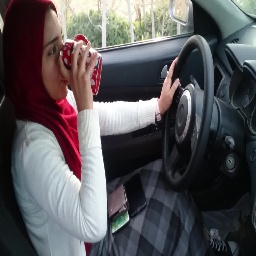}
\includegraphics[width=0.48\textwidth]  {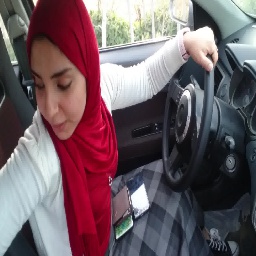}
	\end{minipage}
	}\hfill
\subfloat[Fine-grained action of `playing' (top) vs `with violin' by distinct people] {
\begin{minipage}[c][1\width]{0.23\textwidth}
	   \centering
\includegraphics[width=0.48\linewidth] {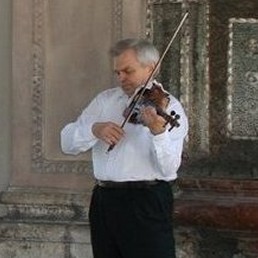}
\includegraphics[width=0.48\textwidth]  {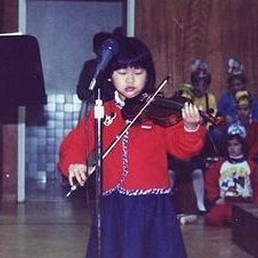} \\ \vspace{1mm}
\includegraphics[width=0.48\textwidth]  {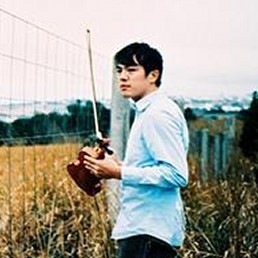}
\includegraphics[width=0.48\textwidth]  {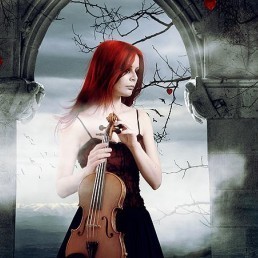}
\end{minipage}
	}\hfill
\subfloat[Same fine-grained action `with cello' involving different people] { 
\begin{minipage}[c][1\width]{0.23\textwidth}
	   \centering
\includegraphics[width=0.48\linewidth] {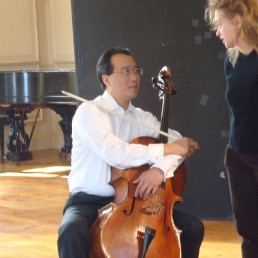}
\includegraphics[width=0.48\textwidth]  {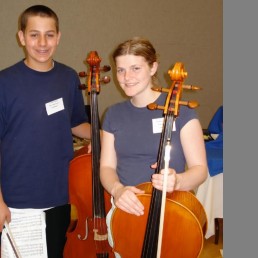}\\ \vspace{1mm}
\includegraphics[width=0.48\textwidth]  {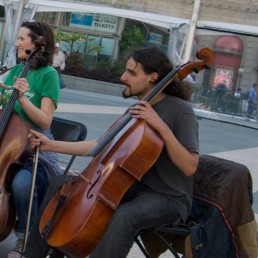}
\includegraphics[width=0.48\textwidth]  {Figures/Norm_With_Cello_071_0.jpg}
\end{minipage}
}
\caption{Traditional image recognition with distinctive classes (a) versus different fine-grained variations in images containing human actions/activities (b-d). The examples representing the driving actions are from AUC-V2 \cite{H.M.Eraqi}, taking photos from Stanford-40 \cite{YaoJKLGF11}, interactions with musical instruments from PPMI-24 \cite{YaoF10}, and food dish from Food-101 \cite{BossardGG14} datasets.
}
\label{fig:variation}
\vspace{-1em}
\end{figure*}

In the above-mentioned three different types of visual recognition tasks, it is observed that many models are designed to capture contextual information at different granularity (e.g. patch to image level) to discriminate various visual categories. The contextual information is often captured by considering salient regions that often focus on bounding-boxes enclosing body-parts and objects. Such approaches require either manual annotations of the target bounding-boxes or object detectors to localize such SRs. This is time-consuming, laborious and often noisy due to human errors. Moreover, the pre-trained object detectors might not have seen the targeted object categories and the samples in the target dataset might have drawn from different distributions to that of the dataset on which the detectors are trained. This results in detection of noisy bounding-boxes. To avoid this, we propose a novel approach that implicitly detects the bounding-box containing SR and is based on well-known keypoints where the semantic content is locally rich. Moreover, we use this information to extract hierarchical contextual information by pairing the detected SRs in a unique way (local patches to global image). This contextual information influences the classification decision made by our deep architecture by \textit{learning to select} these SRs automatically. It is carried out using our novel attention mechanism that considers the importance of subtle changes within an SR for a given class. It is also worth to mention that most of the existing approaches focus on either traditional image recognition with distinctive image categories \cite{krizhevsky2012imagenet,simonyan2014very,he2016deep} (see Fig. \ref{fig:variation}a) or fine-grained image recognition involving subordinate categories (Fig. \ref{fig:variation} b-d). However, it is unclear how a given model developed for a particular recognition task (e.g. species of birds, models of cars, etc.) can be easily adapted to other tasks (e.g. human-objects interactions) without compromising the recognition performance. The proposed AG-Net aims to achieve this via our novel SRs based attention mechanism and can be applied to various visual recognition problems. This is demonstrated by evaluating on diverse benchmark datasets, and our model outperforms the SotA approaches.       

\section{Proposed Approach} \label{proposed}
The overview of our proposed AG-Net is shown in Fig. \ref{fig:teaser}a. The model takes an image and processes it in two parallel streams. One stream focuses on the high-level feature map representation of the image using a base CNN with additional self-attention, squeeze-and-excitation block with the residual connection. The other stream is concentrated on identification of the SRs based on the automatically detected SIFT keypoints. These SRs are then processed to extract the corresponding feature maps. These feature maps are passed to our novel attentional module that \textit{learns to attend} each SR by its importance in making classification decision in the following classification layer. Moreover, we also propose an innovative learnable pooling approach that combines the Global Average Pooling (GAP) and Global Max Pooling (GMP), and establishes correspondences between pooled feature maps and image categories. Thus, both GMP and GAP can be used in an effective way since they are complementary to each other. 
\begin{figure}
\centering
\subfloat{ \includegraphics[width=0.115\textwidth] {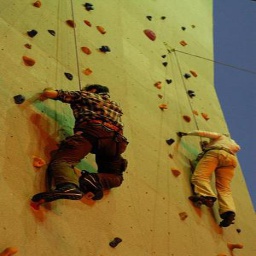}
\includegraphics[width=0.115\textwidth]  {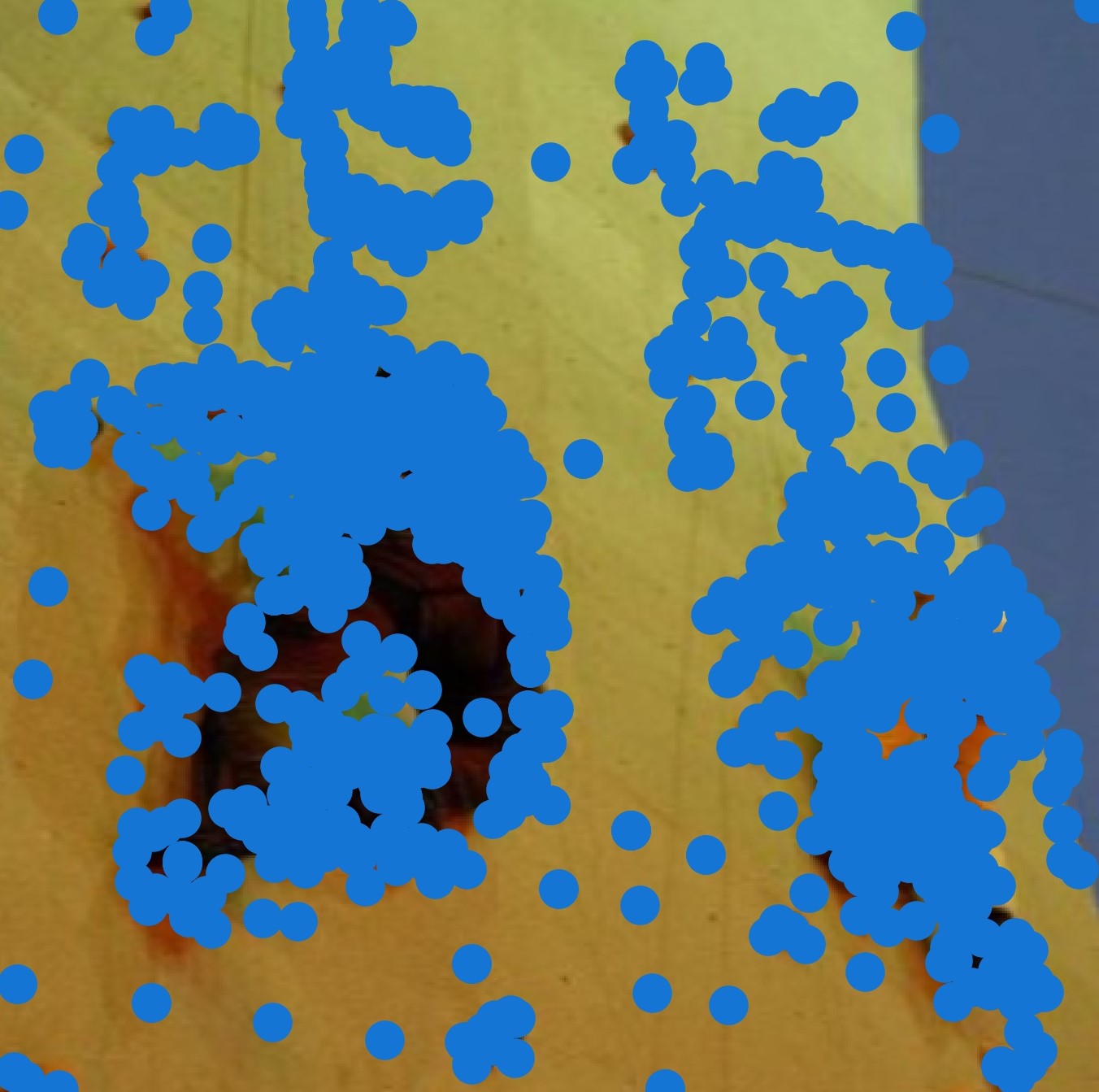} 
\includegraphics[width=0.115\textwidth]  {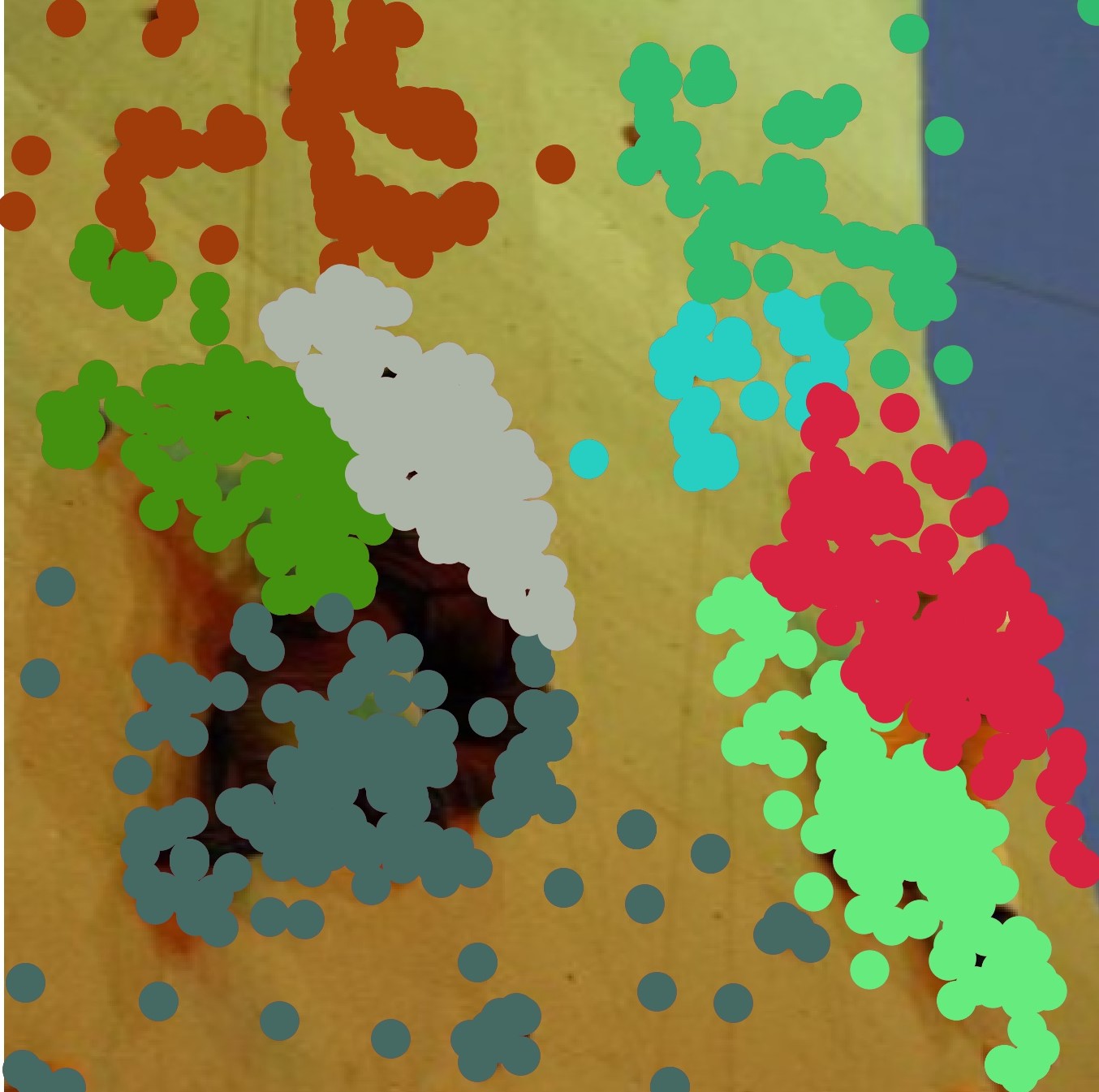}
\includegraphics[width=0.115\textwidth]  {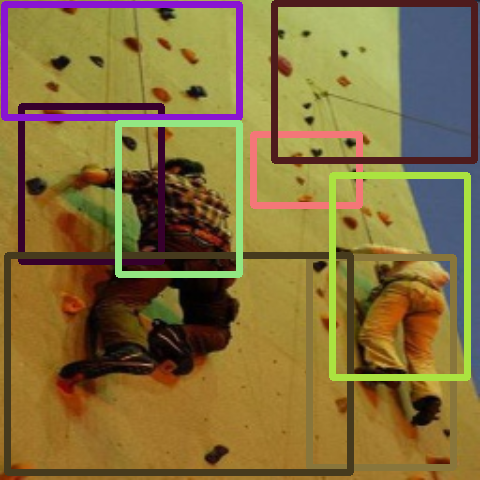}
}
\caption{Example images for climbing action from Stanford-40 dataset \cite{YaoJKLGF11}. Using our keypoints-based clustering for detecting primary SRs. Eight SRs are detected: original image $\Rightarrow$ detected SIFT keypoints $\Rightarrow$ clustered keypoints $\Rightarrow$ bounding boxes enclosing SRs (left to right). Best view in color.}
\label{fig:Primary_8SRs}
\vspace{-1em}
\end{figure}
\begin{figure}
\centering
\subfloat {\includegraphics[width=0.16\textwidth] {Figures/Drinking_V2.jpg} 
\includegraphics[width=0.16\textwidth]  {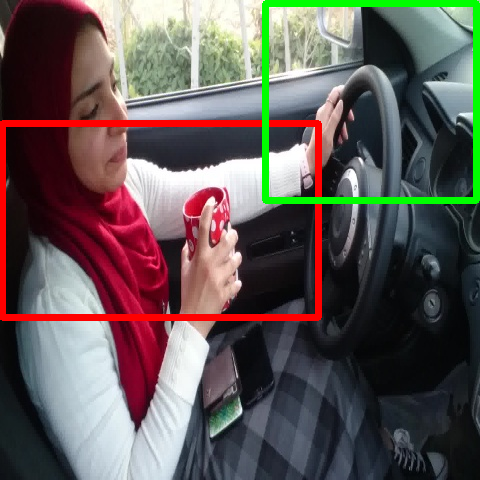}
\includegraphics[width=0.16\textwidth]  {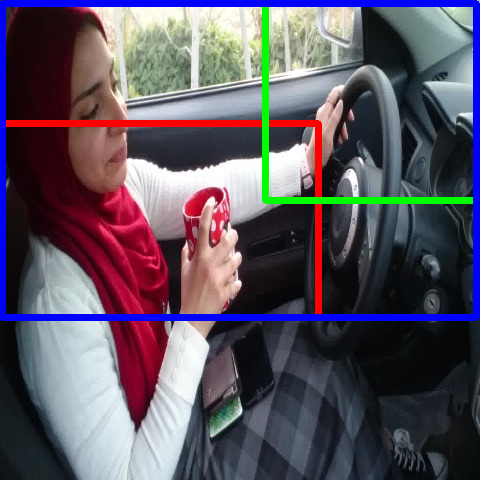}
}
\caption{Drinking image example from AUC-V2 dataset \cite{H.M.Eraqi}. Detection of two primary SRs with which a secondary SR is generated. Original image $\Rightarrow$ detected two primary SRs (\textcolor{red}{red} and \textcolor{green}{green}) $\Rightarrow$ Secondary SR (\textcolor{blue}{blue}) with the primary SRs (left to right). The primary SRs signify partial information such as driving (\textcolor{green}{green}) and drinking (\textcolor{red}{red}), but, those SRs do not reflect the action drinking while driving which is properly described in the secondary SR. 
}
\label{fig:PRs_SRs}
\vspace{-1em}
\end{figure}
\subsection{Salient Regions Generation}\label{sec:sr}
SRs are the informative parts of the images and play a significant role in effective visual recognition. Our aim is to detect these SRs automatically and guide the network by focusing on those that contain most discriminative information so that the network is more attentive to them. We explore the automatic detection of SRs by using well-known SIFT \cite{Lowe04} keypoints, which are distinctive and invariant to rotation, translation, scale, illumination and distortion. 
A GMM \cite{bishop2006pattern} is then applied to the positions of these keypoints in the image plane to group them into $\kappa$ clusters. The bounding-box enclosing each cluster is our primary SRs (Fig. \ref{fig:Primary_8SRs}). The goal is to group the salient keypoints, which are spatially close to each other and their combined locations are represented as a SR. From Fig. \ref{fig:Primary_8SRs}, it is evident that some of the detected clusters (i.e. SRs) are located far from the person(s) in focus. These SRs might not be helpful in providing discriminative cues in visual recognition. It is also observed that in highly textured images, these SRs often partially cover the person(s) in focus executing action(s). Therefore, we derive a secondary set of SRs by considering unique pairs from these primary SRs to capture the larger region containing contextual information in a fine-to-coarse manner, i.e. from local (smaller) to global (larger) salient regions/patches. We achieve this by merging all unique pairs of SRs (Fig. \ref{fig:PRs_SRs}). For example, there will be a $\kappa(\kappa - 1)/2$ possible unique pairs from $\kappa$ primary SRs. Therefore, the final number of possible SRs is $R = \kappa + \kappa(\kappa - 1)/2$, which are used by our attention mechanism to decide their usefulness. Our SR generation method differs from the common bounding box proposals used in object detection. In object detection, the goal is to predict the bounding box of the target more accurately. Whereas, our goal is to recognize the visual content by developing a rich feature descriptor that can ensemble features from multiple local regions so that the subtle variations in images can be described and discriminated as a misalignment of the local pattern. The object detection algorithms usually suggest thousands (e.g. EdgeBoxes \cite{zitnick2014edge} 1K, R-CNN \cite{he2017mask} 2K) of bounding box proposals and this could be a problem for our approach since our model complexity is linked to the number of regions, i.e. more regions imply higher complexity since our model learns joint relationships between them while generating the target feature descriptor. Therefore, we focus on fewer regions and is inspired by \cite{gkioxari2015actions,GkioxariGM15,mallya2016learning} in which parts/objects detectors are used to localize different body parts and objects which are used for actions and/or human-objects interaction classifications. In our case, we follow a very simple yet effective approach for on-the-fly generation of SRs by considering the spatially distributed salient keypoints without requiring any object/part detectors, which are often computationally expensive, and the available pre-trained models might not have seen objects appearing on the target datasets. 
\subsection {Attention Model} 
The proposed attention consists of two types: 1) \textit{self-attention} (also known as \emph{intra-attention}) focusing on high-level convolutional features from the ResNet-50 (Fig. \ref{fig:teaser}a), and 2) \textit{inter-attention} that captures the ``importance" of a given SR with respect to all other SRs and the whole image.
\subsubsection{Self (intra)-attention}
It was introduced in Self-Attention Generative Adversarial Networks (SAGAN) to enable both the generator and the discriminator to better model relationships between spatial regions \cite{H.Zhang2018}.
Inspired by SAGAN, we apply the self-attention layer to the convolutional feature from the ResNet-50 (Fig. \ref{fig:teaser}a \& Fig. \ref{fig:teaser}b). Our aim is to explicitly learn the relationship between high-level convolutional features, even they are located far apart in the feature map. This is complementary to the convolutions to capture the spatial structure. To achieve this, let  $\mathbf{x}$ be the high-level feature map for a given input image $I$. In our AG-Net, the feature $\mathbf{x}$ is the output of the base CNN (ResNet-50) with the resolution of width $W$, height $H$ and channels $C$. The attention concepts \cite{VaswaniSPUJGKP17} of \textit{key}  $f(\mathbf{x})=\mathbf{W}_{f}\mathbf{x}$, \textit{query} $g(\mathbf{x})=\mathbf{W}_{g}\mathbf{x}$ and \textit{value} $h(\mathbf{x})=\mathbf{W}_{h}\mathbf{x}$ are computed from $\mathbf{x}$ using three separate $1\times 1$ convolutions as shown in Fig. \ref{fig:teaser}b. Afterwards, we apply the dot-product attention to compute the self-attention feature map $\mathbf{s}=\{\mathbf{s}_1, \mathbf{s}_2,..., \mathbf{s}_j,...,\mathbf{s}_{W\times H}\} \in \mathbb{R}^ {W\times H\times C}$ as:          
\begin{equation} 
t_{i,j} = \text{Softmax}(f(\mathbf{x}_{i})^{T}g(\mathbf{x}_{j})), \text{ and } \mathbf{s}_{j} = {\sum_{i=1} ^ {W\times H} t_{i,j} h(\mathbf{x}_{i})}\\
\end{equation}
where $t_{i,j}$ is one entry in the attention map (Fig. \ref{fig:teaser}b), mentioning how much attention should be given to the $i^{th}$ position when synthesizing the $j^{th}$ position in $\mathbf{x}$. The weight matrices ${\mathbf{W}_f\in \mathbb{R}^ {C'\times C}}$, ${\mathbf{W}_g\in \mathbb{R}^ {C'\times C}}$ and ${\mathbf{W}_h\in \mathbb{R}^ {C'\times C}}$ are computed using $1\times1$ convolution as suggested in \cite{H.Zhang2018}. For memory efficiency, $C' = \lfloor C/8\rfloor$ is used in all our experiments. 
The output $\mathbf{s}=\{\mathbf{s}_1, \mathbf{s}_2,..., \mathbf{s}_j,...,\mathbf{s}_{W\times H}\} \in \mathbb{R}^ {W\times H\times C}$ is multiplied with a learnable scalar parameter $\delta$ to produce the final output $\mathbf{o}_j = \delta \mathbf{s}_j + \mathbf{x}_j$. 
The scalar parameter $\delta$ is initialized to zero and allows the network to rely on the local cues at first and then gradually learns to assign more weight to the global evidence. Afterwards, the output $\mathbf{o}$ of the self-attention layer is used for modelling inter-attention as described below. 
\subsubsection{Inter-attention} 
Our aim is to allow the model to \textit{learn to attend} each SR (Section \ref{sec:sr}) by its importance in decision making. It is achieved by introducing an inter-attention layer (Attention module in Fig. \ref{fig:teaser}a), which focuses on how much to attend a given SR conditioned on all other SRs and the whole image $I$. To attain this, each SR is first represented as the corresponding feature $f_r$. The generated SRs are of different spatial sizes. Thus, we use bilinear pooling on each SR to compute the corresponding fixed feature map $f_r$ (e.g. spatial resolution of $7\times 7$). The pooled feature $f_r$ is then passed through a Squeeze-and-Excitation (SE) block (Fig. \ref{fig:teaser}c) to enhance its representational capacity by modelling interdependencies between the channels \cite{HuSS18}. The SE block learns to apply global information to emphasize only on discriminative pooled features from an SR. The SE block's output is connected to its input via a residual path (namely, SE-Residual), which can refine contextual information for improving learning capability. In our experiment, it is observed that the image recognition performance is improved by introducing this residual path, which smooths the feature propagation in the network and also enhances channel-wise feature recalibration \cite{he2016identity}.

There are $R$ generated SRs (Section \ref{sec:sr}) from the image $I$, resulting in $R+1$ regions (whole image $I$ is also a region).
The corresponding $R+1$ bilinearly pooled features are represented as $f_r={f_1,f_2,\dots,f_{R+1}}$. Then, these features are processed with the respective SE-Residual block (Fig. \ref{fig:teaser}c) to produce the desired outputs $\mathtt{f}_r={\mathtt{f}_1,\mathtt{f}_2,\dots,\mathtt{f}_{R+1}}$ to be considered for computing inter-attention. 
Our inter-attention module learns an attention matrix $M$ representing the similarity of a given SR ($r$) with respect to all the other SRs ($r'$) in $I$. The component $m_{r,r'} \in M$  denotes the similarity between the features $\mathtt{f}_r$ and $\mathtt{f}_{r'}$ representing SRs $r$ and $r'$, respectively. The inter-attention element $ m_{r,r'}$ is computed as:
\begin{equation}  
\begin{split}
m_{r,r'} &=\sigma (W_m u_{r,r'}+{b_m}) \\
u_{r,r'} &=\tanh({W_u}{\mathtt{f}_r}+{W_u'}{\mathtt{f}_{r'}}+{b_u} )
\end{split}
\label{eq:att}
\end{equation}
where $r, r' = 1, \cdots, R+1$; $\sigma $ is the element-wise sigmoid function; ${W_u}$ and ${W_u'}$ are the weight matrices for the respective feature maps $\mathtt{f}_r$ and ${\mathtt{f}_{r'}}$; 
${W_m}$ is the weight matrix corresponding to their non-linear combination; ${b_u}$ and ${b_m}$ are the corresponding bias vectors. The component $m_{r,r'}$ is responsible in conveying how much to attend the feature $\mathtt{f}_r$ representing an SR in focus conditioned on all the other surrounding SRs ($\mathtt{f}_{r'}$) in the image $I$. The element $m_{r,r'}$ is multiplied with the respective feature maps $\mathtt{f}_{r'}$ ($r' \in [1,R+1]$) and aggregated over all SRs to produce the required attentional feature map $\alpha_r=\sum_{r'=1}^{R+1}m_{r,r'}\mathtt{f}_{r'}$ for the SR $r$. 
Now, we have obtained the attentional feature $\alpha_r$ for each $r\in [1, R+1]$. Next, the goal is to combine all the regional attentional features to produce the final feature for the whole image $I$. We achieve this by learning their importance scores as weights $w=\{w_1,w_2,\dots ,w_{R+1}\}$ and is computed as: 
\begin{equation} 
 \mathbf{\hat{f}}=\sum_{r=1} ^ {R+1}  \alpha_{r}w_r, \text{ where } w_r =\text{Softmax}(W_{\alpha} \alpha_r + b_{\alpha})
\label{eqn:sum}
\end{equation}
where the weight matrix $W_\alpha$ and the bias $b_\alpha$ are learnable parameters. The score $w_r$ for each SR represented with $\alpha_r$ is computed via probability distribution using a Softmax function. This approach is similar to the attention-based approach used to solve machine translation problem \cite{BahdanauCB14} in which the model automatically searches for parts of a source sentence that are relevant to predict a target word. The main difference is that we do not consider the sequential information. The final feature map $\mathbf{\hat{f}}$ obtained as an output is a
high-level encoding of the entire image $I$ and is used as an input to our classification layer to solve the image recognition problem.
\subsection{Classification}\label{sec:classification}
It takes as the input feature map $\mathbf{\hat{f}}$ ($width\times height\times channels$) and provides the output probability vector of length $\mathcal{C}$ representing the probabilities of $\mathcal{C}$ classes for a given classification task. Normally, this is done using a Global Average Pooling (GAP) or Global Max Pooling (GMP) to aggressively summarize the presence of a feature in an image and replaces the traditional fully-connected layer \cite{he2016deep,huang2017densely}. 
It can be  seen as a structural regularizer that explicitly enforces feature $\mathbf{\hat{f}}$ to
be confidence maps of $\mathcal{C}$ classes without any parameters to optimize and thus, overfitting is avoided. Often, the selection of GAP or GMP is done empirically and is dependent on the task in hand. Moreover, the GAP equally aggregates all the values as the final output, which favors the saliency, but it cannot directly indicate which inputs are the true prediction. It also gives too much focus on frequently occurring patches in the input image, whereas GMP is the opposite. It only selects the largest value as the final result and also loses the other useful information for accurate recognition. Thus, a combination of both can capture complementary information. To this end, we use a weighted summation of GMP and GAP by learning their importance. The method is similar to (\ref{eqn:sum}) but considers only pooled GMP ($\mathbf{\hat{f}}_{\text{GMP}}$) and GAP ($\mathbf{\hat{f}}_{\text{GAP}}$) features:
\begin{equation} 
 F = \omega \mathbf{\hat{f}}_{\text{GMP}} + (1 - \omega) \mathbf{\hat{f}}_{\text{GAP}},\text{ and } \omega = \text{Softmax}(W_\omega \mathbf{\hat{f}}_{p}  + b_{\omega}) 
\label{eqn:pool}
\end{equation}
where $p\in \{\text{GMP}, \text{GAP}\}$, and the weight matrix $W_\omega$ and bias $b_{\omega}$ are learnable parameters. The final feature vector $F$ is then passed through the $\mathtt{Softmax}$ to generate output vector representing class probabilities of length $\mathcal{C}$.
\subsection{Model Implementation and Learning}\label{sec:learning}
AG-Net is implemented using TensorFlow and Keras. The GMM \cite{bishop2006pattern} is initialized with k-means. It uses full co-variance matrix with regularization of {$10^{-6}$} and runs 100 iterations with a convergence threshold of {$10^{-3}$}. We evaluate AG-Net using ResNet-50 \cite{he2016deep}, DenseNet-121 \cite{huang2017densely}, Inception-V3 \cite{szegedy2016rethinking}, NASNet-Mobile \cite{zoph2018learning}, and VGG-16 \cite{simonyan2014very} as a backbone network. The backbone CNNs are initialized with the pre-trained ImageNet weights. The end-to-end AG-Net is trained with the default image size of $224\times 224$. We follow the standard data augmentation techniques of random translation ($\pm 15\%$ of width and height in horizontal and vertical directions), random rotation ($\pm 15$ degrees) and random scaling (ranging from 0.85 to 1.15) to enhance the generalization capability of the model. The Stochastic Gradient Descent (SGD) optimizer with a momentum of 0.99 is applied to minimize the categorical cross-entropy loss function.
\begin{equation} 
  L_i=-\sum_{c=1}^{\mathcal{C}} {y_{i,c} log(p_{i,c})} 
 \label{eq:cross_entropy}
\end{equation}
where $p_{i,c}$ represents the prediction probability of class $c$, ${y_{i,c}}$ is the actual class-label of the $i^{th}$ image, and $\mathcal{C}$ is the number of classes. During training, the loss in (\ref{eq:cross_entropy}) is accumulated (i.e., $L = \frac{1}{N}\sum_{i=1}^{N} L_i$) over all the training images $I=\{I_i|i=1\dots N\}$ with the respective labels $y=\{y_i|i=1\dots N\}$, where $N$ is the total number of training images. For the SGD, an adaptive learning rate with an initial value of {$10^{-5}$} is applied and is reduced further by a decay factor of 0.1 after every 25 epochs. The model is trained with a batch size of 8 for 50 epochs using an NVIDIA Titan V GPU (12 GB).
\section{Dataset description} \label{datasets}
The effectiveness of our AG-Net is evaluated using six diverse datasets consisting of fine-grained human actions (e.g. driver activities), human-objects interactions (e.g. playing musical instruments), different food images, and extremely diverse and more general object categories. Moreover, the size of these datasets ranges from small ($\sim$5K ) to medium ($\sim$10K,$\sim$14K and $\sim$17K) to large ($\sim$31K and 101K).   

\noindent\textbf{Distracted Driver V1 (AUC-V1)} \cite{AbouelnagaEM17}: It consists of 12,977 training and 4,331 testing images. The images are collected from 31 persons (22 males and 9 females) from 7 different countries: Egypt (24), Germany (2), USA (1), Canada (1), Morocco (1), Palestine (1), and Uganda (1). These datasets comprise 10 activities: C0: driving safely, C1: texting right, C2: talking on the phone-right, C3: texting left, C4: talking on the phone-left, C5: operating radio, C6: drinking,  C7: reaching behind, C8: hair and makeup, and C9: talking to passenger.

\noindent\textbf{Distracted Driver V2 (AUC-V2)} \cite{H.M.Eraqi}: It is an enhanced version of the aforesaid dataset. The images are gathered from 44 persons (29 males and 15 females) from the above-mentioned 7 countries. It comprises of 12,555 training images from 38 drivers, and 1,923 testing images from the remaining 6 drivers (i.e. driver-wise split).

\noindent\textbf{Stanford-40 Action (S-40)} \cite{YaoJKLGF11}: The dataset consists of 9,532 images with 40 different human actions such as blowing bubbles, brushing teeth, fishing, gardening, etc. The dataset consists of 4,000 training and 5,532 testing images.\\
\textbf{People Playing Musical Instruments (PPMI-24)} \cite{YaoF10}: This is a challenging dataset to discriminate two fine-grained human interactions with 12 musical instruments such as flute, guitar, harp, violin, etc. Specifically, whether a person is playing a musical instrument (PPMI+) or simply holding the instrument without playing it (PPMI-). Altogether 24 interactions are presented with 4,800 images. For each action, 100 training and 100 test samples are provided for fine-grained classification.

\noindent\textbf{Food-101} \cite{BossardGG14}: It comprises 101 food classes, with a total of 101,000 real-world images. For each class, 750 training images and 250 test images are available. It contains very diverse as well as similar food classes.

\noindent\textbf{Caltech-256} \cite{griffin2007caltech}: It contains a total of 30,607 images of 256 object categories collected from the Internet. This dataset is an improvement on its predecessor, the Caltech-101, with new features such as larger classes, new and larger clutter categories, and overall increased difficulty. The categories are extremely diverse and more general, ranging from grasshopper to tuning fork to recognize frogs, cell phones, sailboats and many other categories in cluttered pictures. We follow the evaluation protocol in \cite{ge2019weakly} that uses 60 random samples per class for training and the rest for testing, and the mean and standard deviation of five runs are reported. 
\begin{table}
\centering
 \caption{Overall Performances of the AG-Net on Diverse Visual Recognition Datasets: AUC-V1 \cite{AbouelnagaEM17}, AUC-V2 \cite{H.M.Eraqi}, Stanford-40 \cite{YaoJKLGF11}, PPMI-24 \cite{YaoF10},  Food-101 \cite{BossardGG14}, and Caltech-256 \cite{griffin2007caltech}}
  \label{tab:freq}
   \begin{tabular} {l c c c c c c}
    \toprule
  Metrics &V1 &V2 &Stanford & PPMI & Food & Caltech \\ 
    \midrule
    Top-1 (\%)  & 99.70  & 96.65  &97.83 &98.20 & 99.30 & 96.89 \\ 
    Top-5 (\%)  & 100.00  & 99.89 &99.85  &99.96 & 99.87 & 99.68 \\
    mAP (\%)  & 99.63  & 95.39   &96.21  &97.35 & 98.93  & 94.57\\ 
  \bottomrule
  \end{tabular}
\label{table:overall}
\end{table}
\begin{table}[t]
\centering
  \caption{Comparison with SotA methods on Datasets AUC-V1 \cite{AbouelnagaEM17} and AUC-V2 \cite{H.M.Eraqi}. Accuracy is Given in Percentage (\%)}
  \label{tab:commands}
  \begin{tabular}{l c | l c}
    \toprule
    Method & AUC-V1 & Method & AUC-V2\\
    \midrule
     Fusion\cite {Alotaibi2019} &92.36 & VGG-16 \cite{H.M.Eraqi} & 76.13\\
    DenseNet \cite{BeheraK18} & 94.20 & ResNet-50 \cite{H.M.Eraqi} &81.70   \\ 
    Inception-V3 \cite{H.M.Eraqi} & 95.17 & Inception-V3 \cite{H.M.Eraqi} & 90.07     \\
    MVE \cite{H.M.Eraqi} & 95.77 & - & -\\
    GAWE \cite{H.M.Eraqi} & 95.98 & -& -   \\
    VGG 16-Reg.\cite {BahetiGT18} &96.31 & -&-\\
    \hline
    \textbf{AG-Net} & \textbf{99.70} &  \textbf{AG-Net} &\textbf{96.65}  \\
    \bottomrule
  \end{tabular}
  \label{table:V1-V2}
  \vspace{-1em}
\end{table}
\section{Experimental Results} \label{experiments}
For the performance assessment and comparison, we use the standard metrics of accuracy (Acc) and mean average precision (mAP), and are presented in percentage (\%). 
The overall performances of our AG-Net on six datasets, as mentioned above, are summarized in Table \ref{table:overall}.
\begin{table*}
\centering
  \caption{Comparison with SotA approaches on different datasets. Methods denoted with {$^\dagger$} used keypoints (e.g., SIFT).}
  \label{tab:commands}
  \begin{tabular}{p{2.3cm} p{0.35cm} p{0.5cm}|p{2.3cm} p{0.37cm} p{0.5cm}|p{2.3cm} p{0.6cm} p{0.60cm}|p{2.5cm} p{1.2cm}}
    \toprule
    \multicolumn{3}{c} {Stanford-40 \cite{YaoJKLGF11} }  &
      \multicolumn{3}{c}{PPMI-24 \cite{YaoF10}  }  &
      \multicolumn{3}{c}{Food-101  \cite{BossardGG14}  } &
      \multicolumn{2}{c}{Caltech-256 \cite{griffin2007caltech} }\\
    Method & Acc. & mAP & Method & Acc. & mAP & Method & top-1 & top-5 & Method & Acc.$\pm$ std  \\
    \midrule
  Concepts  \cite{RosenfeldU16b} & 83.12 &  - &   Randomized$^\dagger$\cite{yao2011combining} & - & 47.00 & Ensemble \cite{PandeyDMP17} & 72.12 &91.61 & IFK$^\dagger$ \cite{perronnin2010improving} &47.90$\pm$0.40 \\
Color fusion \cite{Lavinia2019NewCF} & 84.24 & 83.25 & Mutual context \cite{yao2012recognizing} & - & 48.00 &DeepFood \cite{liu2016deepfood} & 77.40 &93.70 & CLM$^\dagger$ \cite{wang2016towards} &53.60$\pm$0.20 \\
Action mask \cite{4CWCDL16} & - &82.64 & Saliency$^\dagger$ \cite{sharma2012discriminative} & - & 49.40  &Kernel pooling  \cite{CuiZWLLB17} &85.50 & - & FV$^\dagger$ \cite{sanchez2013image} &57.30$\pm$0.20  \\
VLAD pyrd. $^\dagger$ \cite {YanSZ17} &- &88.50  & Exemplar \cite{Hu2013recognising} &49.34 & 47.56  & FCAN \cite{LiuXWL16} & 86.50 &-  & ZF-Net \cite{zeiler2014visualizing} & 74.20$\pm$0.30   \\
RFBA net \cite{LiuGQWL19} &- & 90.92 & GMP-VLAD \cite{ZhangLPXS16} & 50.67 & 48.60 &WISeR \cite {MartinelFM18} & 90.27 &98.71  & VGG19+VGG16 \cite{simonyan2014very} & 86.20$\pm$0.30 \\
Multi-br. atn. \cite{YanSLZ18} & - & 90.70 & GSPM \cite{ZhaoMC17} & - & 51.70 &  Incept.-ResNet\cite{CuiSSHB18} & 90.40 & - & L$^2$-SP \cite{xuhong2018explicit} &87.90$\pm$0.20 \\
Part action  \cite {ZhaoMY16} &- &91.20   & Color fusion \cite{Lavinia2019NewCF} & 65.94 & 65.85 &MSMVFA\cite {JiangMLL20} & 90.59 &98.25  & VSVC \cite{zhang2019multi} &91.35$\pm$0.43 \\ 
 Human-mask \cite {liu2018loss} & - & 94.06 & VLAD pyrd.$^\dagger$\cite {YanSZ17} &- &81.30 &GPipe \cite {GPipeFood} & 93.00 & - & CPM \cite{ge2019weakly} &94.30$\pm$0.20 \\ 
      \hline
    \textbf{AG-Net} & \textbf{97.83} & \textbf{96.21}  &\textbf{AG-Net} & \textbf{98.20}  &\textbf{97.35}  &\textbf{AG-Net} & \textbf{99.30}  &\textbf{99.87} &\textbf{AG-Net} & \textbf{96.89}$\pm$\textbf{0.42} \\ 
    \bottomrule
  \end{tabular}
  \label{table:Sota_comparison}
\end{table*}
\begin{center}
\begin{table}
\centering       
  \caption{Accuracy (\%) of AG-Net using Various CNNs. For the Stanford-40 and PPMI-24, the mAP is Given in Parenthesis.}
  \label{tab:commands}
  \begin{tabular}{p{1.1 cm} p{0.4cm} p{0.45cm} p{1.52cm} p{1.52cm} p{0.45cm} p{0.6cm}}
    \toprule
    CNN & V1 & V2 &Stanf. (mAP) &PPMI (mAP) & Food & Caltech \\
    \midrule
    ResNet50 & \textbf{99.70} &94.56 &\textbf{97.83} (\textbf{96.21})  & 97.66 (96.51) &99.11 & \textbf{96.89}  \\ 
    Inception &98.60 & 93.83 & 96.85 (94.36) &96.08 (94.30) &99.22 & 96.84 \\
    DenseNet &99.69 & \textbf{96.65} & 97.32 (95.38) & \textbf{98.20} (\textbf{97.35}) &\textbf{99.30} & 96.83\\
    NASNet &99.67 & 94.40 & 95.00 (92.06) &94.75 (91.25) &99.24 &94.95 \\
    VGG-16 &99.65 & 95.16 & 95.15 (91.72) &96.54 (92.51) &98.07 & -- \\ 
      \midrule
  Prev Best &96.31  &90.07 &84.24 (94.06) &65.94 (81.30)  &93.00 &94.30\\
    \bottomrule
  \end{tabular}
  \label{table:abl_CNN}
\end{table}
 \end{center}
\vspace{-1em}
\subsection{Performances on AUC-V1 and AUC-V2}
Our AG-Net has achieved 99.70\% and 96.65\% accuracy on AUC-V1 and AUC-V2, respectively. These are significantly higher than all the SotA approaches shown in Table \ref{table:V1-V2}. Our AG-Net's accuracy is 3.39\% and 6.58\% higher than the best SotA approach \cite{BahetiGT18} on AUC-V1 and Inception-V3 \cite{H.M.Eraqi} on AUC-V2, respectively. Using AUC-V1, a fusion of three CNNs (Inception, ResNet and hierarchical RNN) is tested in \cite{Alotaibi2019}. It is worth to mention that the number of parameters of our network is also comparable to those of existing approaches. For example, in \cite{BahetiGT18}, the VGG-16 with regularization (VGG 16-Reg.) containing 140M (millions) parameters has produced the best accuracy (96.31\%), whereas our model has only 54.79M parameters. Similarly, the DenseNet is adapted to inject latent pose to enhance the recognition accuracy in \cite{BeheraK18}. Although, the number of training parameters is 18.34M, the complexity of the model is significantly higher since it uses both DenseNet and OpenPose models for training and inferencing. The model in \cite{H.M.Eraqi} is complex since it incorporates five different CNNs. Even though their majority voting ensemble (MVE) achieves 95.77\%, it is still lower than ours (99.70\%).

\begin{figure*}
\centering
\subfloat[]{\includegraphics[width=0.31\textwidth]{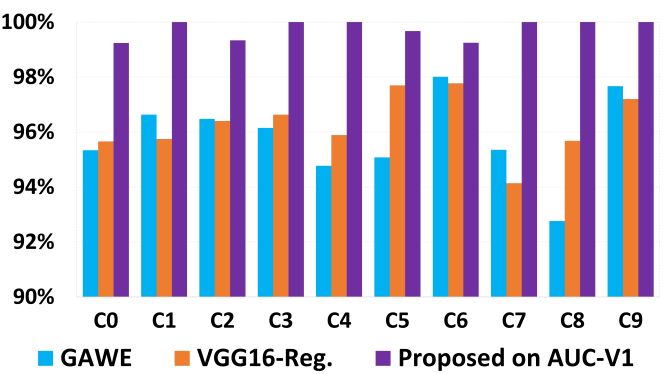} \label{fig:Class_Acc_V1} } \hfill
  \subfloat[]{\includegraphics[width=0.28\textwidth]{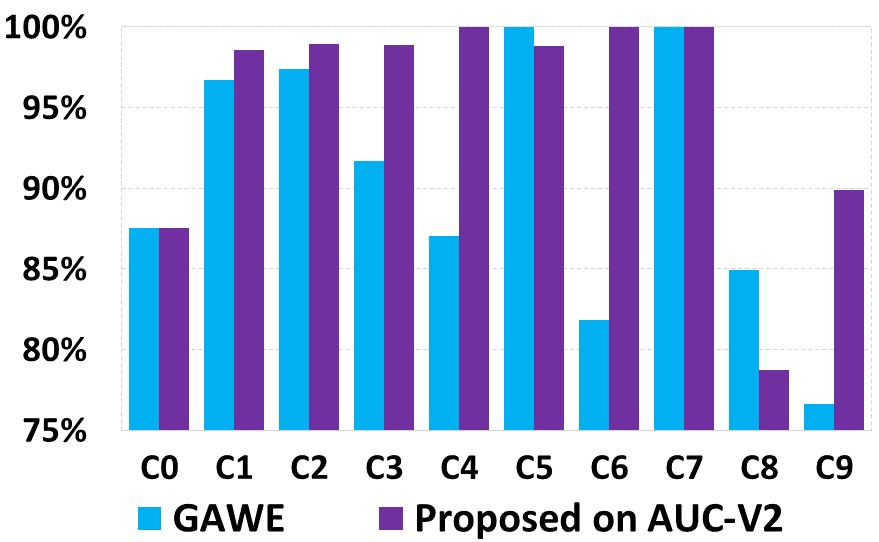} \label{fig:Class_Acc_V2}  } \hfill
\subfloat[]{\includegraphics[width=0.33\textwidth]{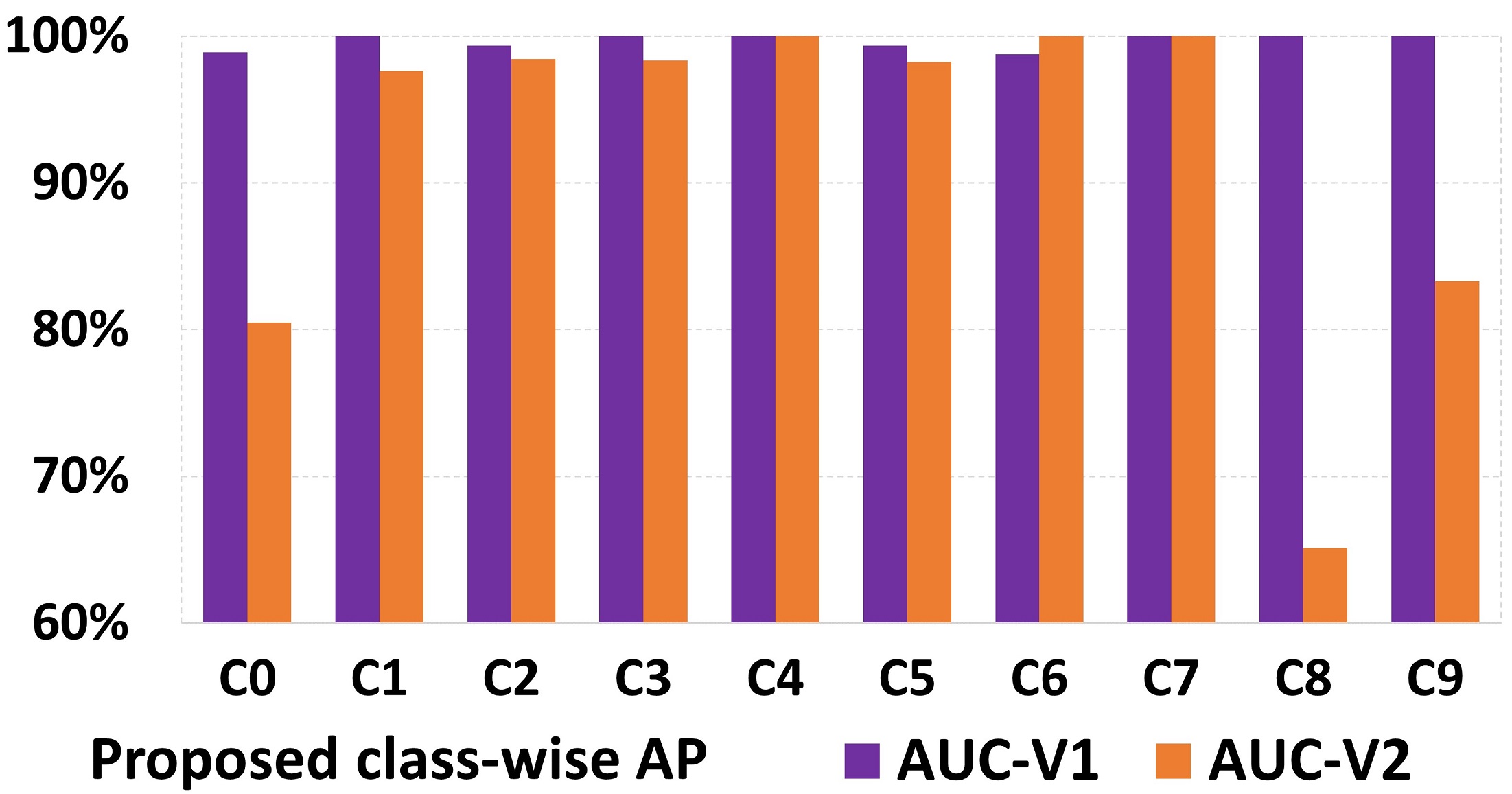} \label{fig:mAP_V1_V2} }\hfill
\caption{ (a-b) Accuracy of different methods on each driving activity. (a) Comparison with GAWE \cite{H.M.Eraqi} and VGG 16-Reg. \cite{BahetiGT18} on AUC-V1. (b) Comparison with GAWE \cite{H.M.Eraqi} on AUC-V2. (c) Proposed AP of each driving activity on AUC-V1 and AUC-V2.}
\label{fig:Class_Acc_mAP_V1_V2}
\end{figure*}

The training-testing samples of AUC-V2 are divided based on the unique drivers, cars, driving condition, and other environmental factors to represent the realistic driving scenarios. As a result, this disjoint split  causes a less correlation between training and testing samples. Thereby, the accuracy on AUC-V2 is inferior in comparison to the AUC-V1 for SotA methods. It is also the case for our AG-Net. On AUC-V2, the Inception-V3 in \cite{H.M.Eraqi} has achieved the best (90.07\%) among the existing SotA methods, and our AG-Net is 6.58\% higher. The class-wise accuracy of our AG-Net for each driving activity (C0-C9) is presented in Fig. \ref{fig:Class_Acc_mAP_V1_V2}a-b, including its 
comparison with \cite{H.M.Eraqi,BahetiGT18} using AUC-V1 and with \cite{H.M.Eraqi} using AUC-V2, respectively. The accuracies on AUC-V1 have been improved for all activities (Fig. \ref{fig:Class_Acc_mAP_V1_V2}a). On AUC-V2 (Fig. \ref{fig:Class_Acc_mAP_V1_V2}b), class-level accuracies are also improved over six actions (C1, C2, C3, C4, C6, and C9), the same over two actions (C0 and C7) and deteriorated over C5 and C8 in comparison to \cite{H.M.Eraqi}. These exceptions are C5: \emph{operating radio} (-1.18\%); and C8: \emph{hair and makeup} (-6.17\%). The most confusing one is \emph{hair and makeup}. The reason could be the absence of temporal information in still images. Also, when a driver is involved in this particular action, the head movement in either direction causing a mix-up with other activity such as \emph{talking to passenger} (C9),  \emph{talking on phone in right} (C2) or \emph{talking on phone in left} (C4). It is worth mentioning that existing works do not use the average precision (AP) metric in their evaluation. For the first time, we provide AP of every activity in both AUC-V1 and AUC-V2 datasets (Fig. \ref{fig:Class_Acc_mAP_V1_V2}c). The mean AP (mAP) is 99.63\% and 95.39\% (Table \ref{table:overall}) for AUC-V1 and AUC-V2, respectively. The confusion matrices of both datasets are shown in the supplementary document.
\begin{figure}[h]
  \centering
    \includegraphics[width=\linewidth]{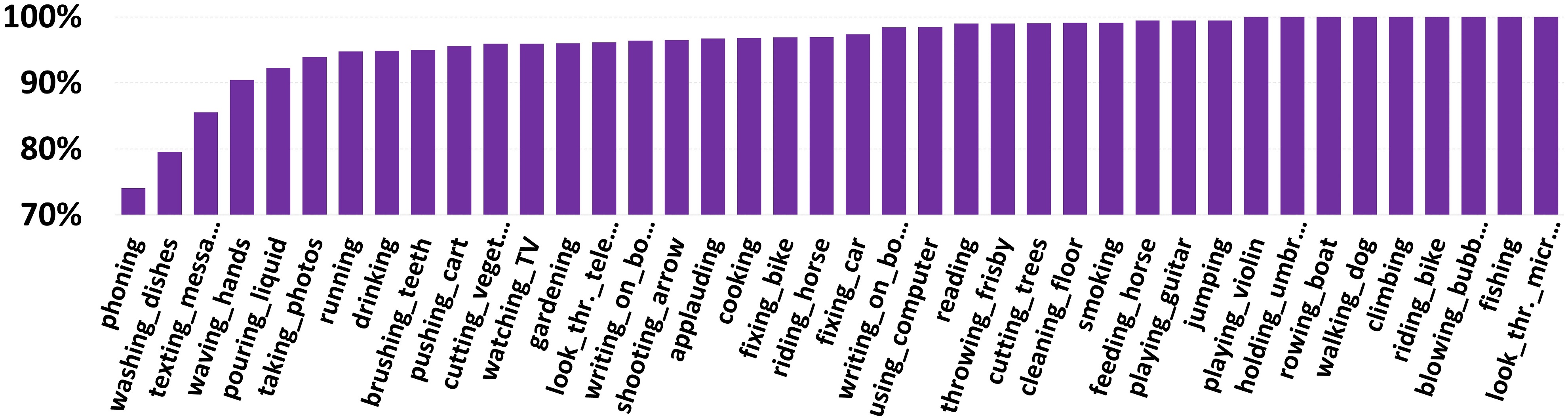}
  \caption{AP of the proposed AG-Net on each action of Stanford-40. }
 \label{fig:S_40_class}
\end{figure}
\begin{figure}[h]
  \centering
    \includegraphics[width=\linewidth]{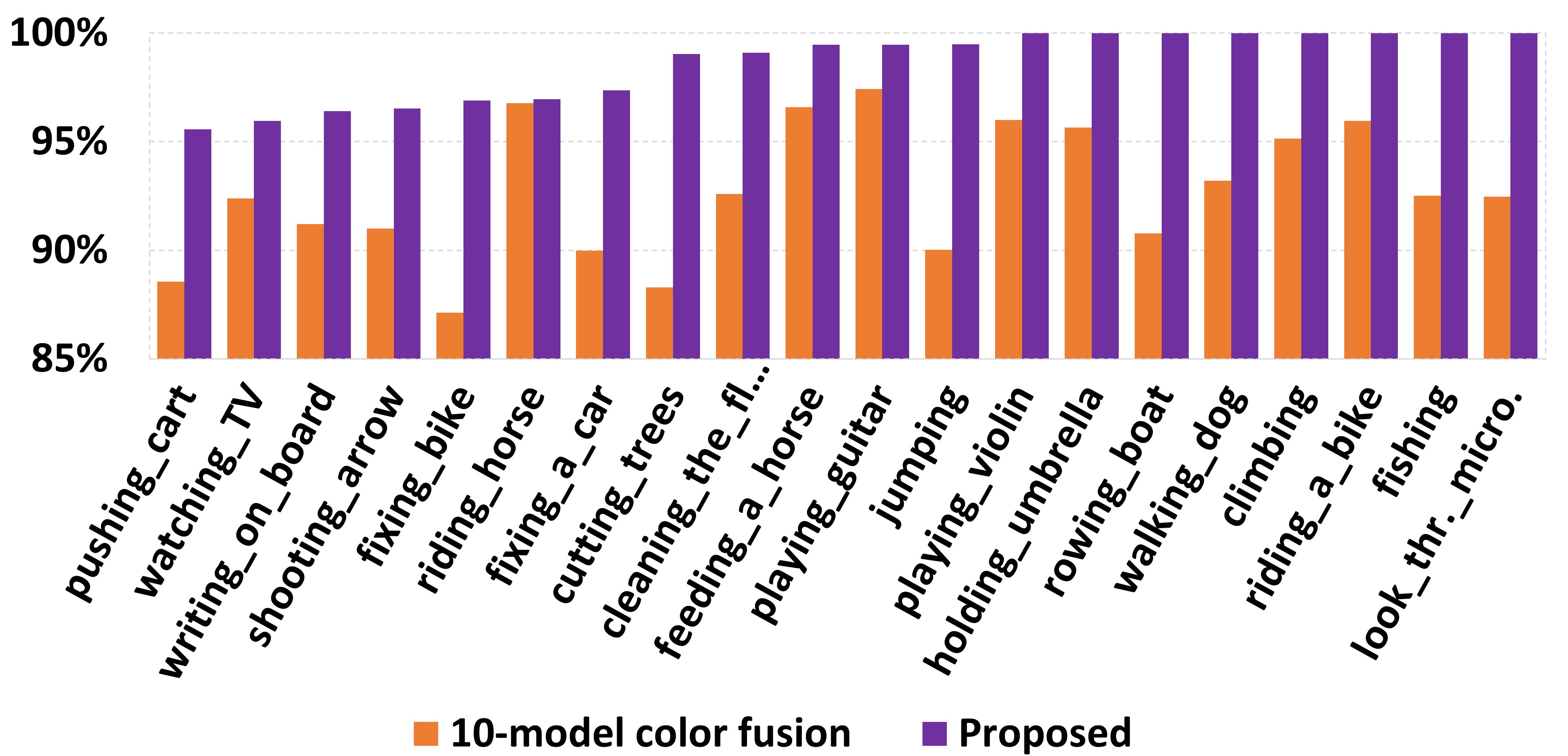}
  \caption{AP comparison on Stanford-40 using the top-20 classes as reported in 10-model color fusion \cite{Lavinia2019NewCF} with our proposed AG-Net.}
 \label{fig:S_40_AP_comparison}
\end{figure}
\begin{figure}[h]
  \centering
  \includegraphics[width=\linewidth]{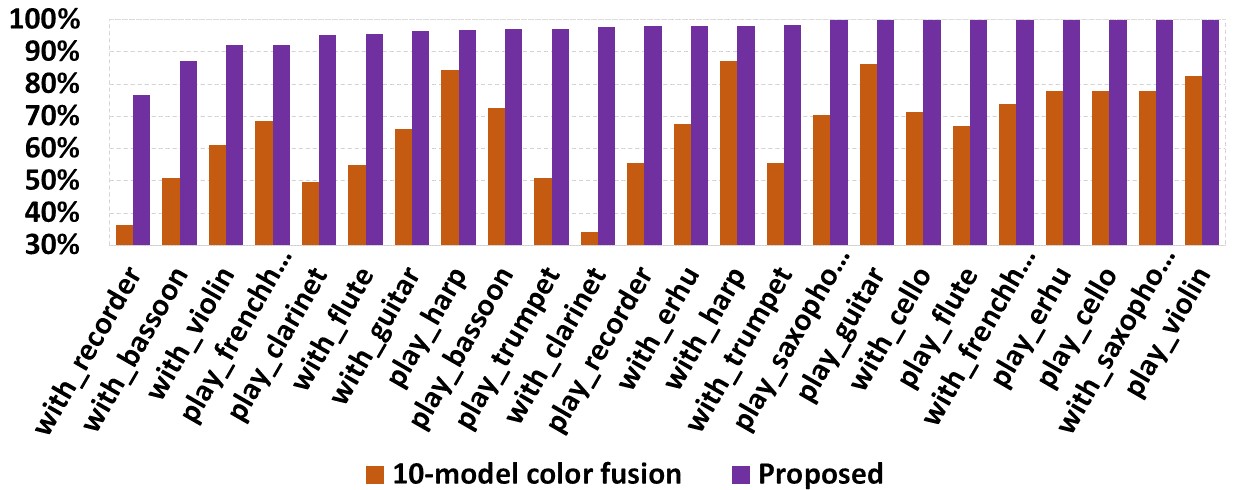} 
  \caption{AP of the proposed AG-Net on PPMI-24 is compared with 10-model color fusion \cite{Lavinia2019NewCF}.}
 \label{fig:PPMI_class}
\end{figure}
\begin{figure}[h]
  \centering
   \includegraphics[width=\linewidth]{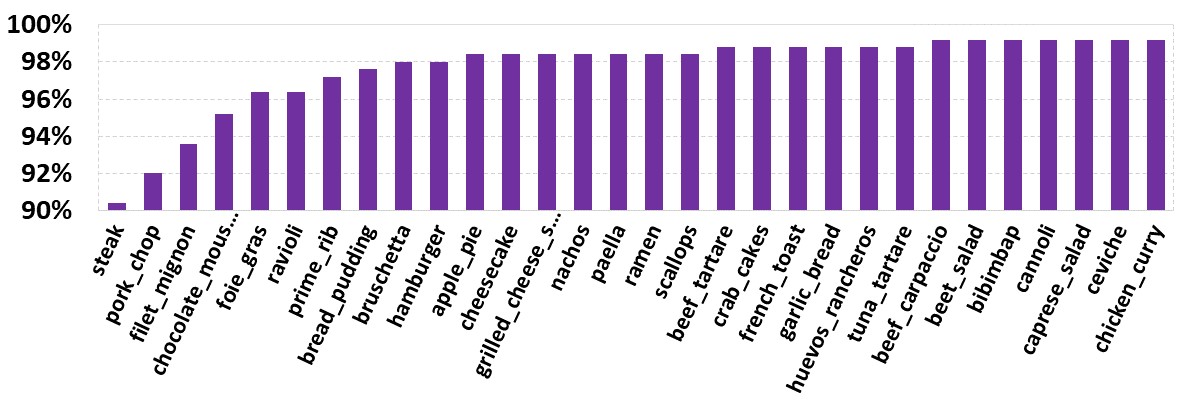}
  \caption{The worst top-1 accuracy of 30 classes of Food-101 using AG-Net.} 
 \label{fig:Food_class}
\end{figure}
\begin{figure*}
\centering
 \subfloat[blowing bubbles,  climbing, feeding a horse ]{
 \begin{minipage}[c][0.65\width]{0.31\textwidth}
	   \centering
 \includegraphics[width=0.32\linewidth] {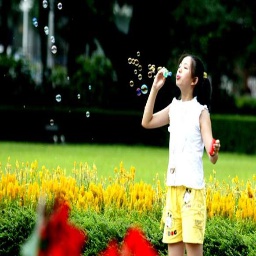} 
\includegraphics[width=0.32\linewidth] {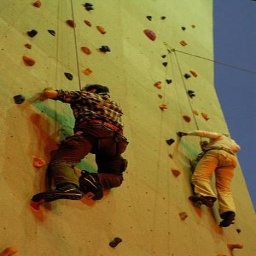}
\includegraphics[width=0.32\linewidth] {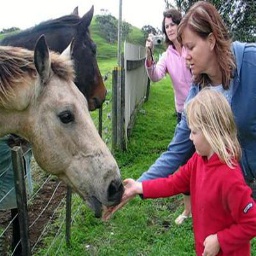} \\ \vspace{1mm}
\includegraphics[width=0.32\textwidth] {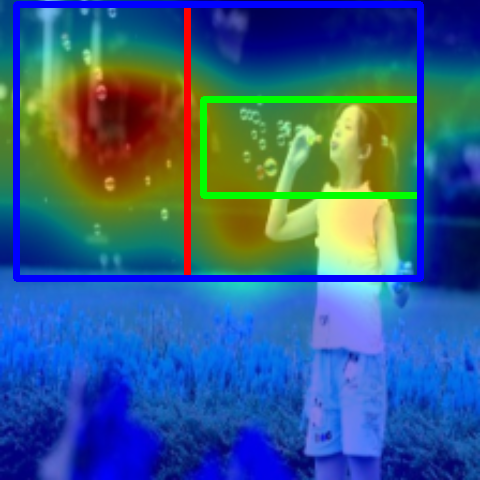}
\includegraphics[width=0.32\textwidth] {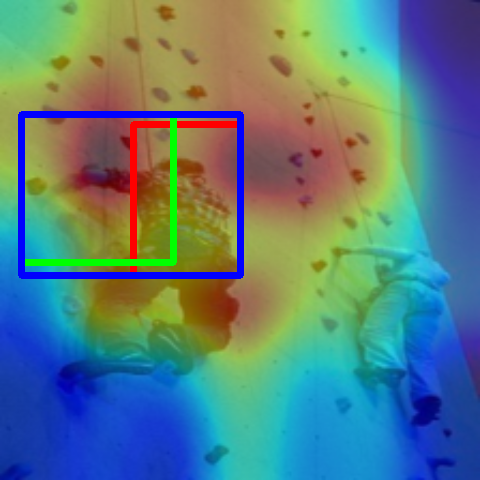}
\includegraphics[width=0.32\textwidth] {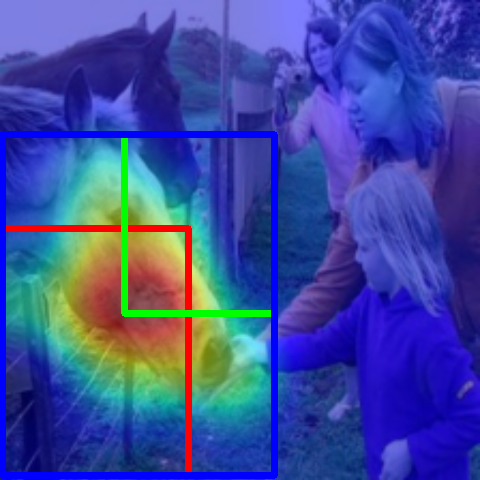}
\end{minipage}
}\hfill
\subfloat[playing Guitar, with Guitar, playing Violin] {
 \begin{minipage}[c][0.65\width]{0.31\textwidth}
	   \centering
\includegraphics[width=0.32\linewidth] {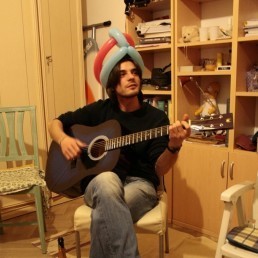} 
\includegraphics[width=0.32\linewidth] {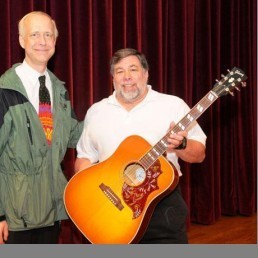}
\includegraphics[width=0.32\linewidth] {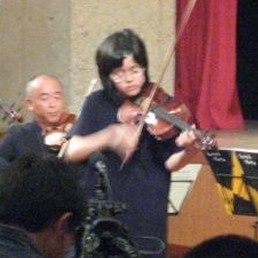}\\ \vspace{1mm}
\includegraphics[width=0.32\textwidth] {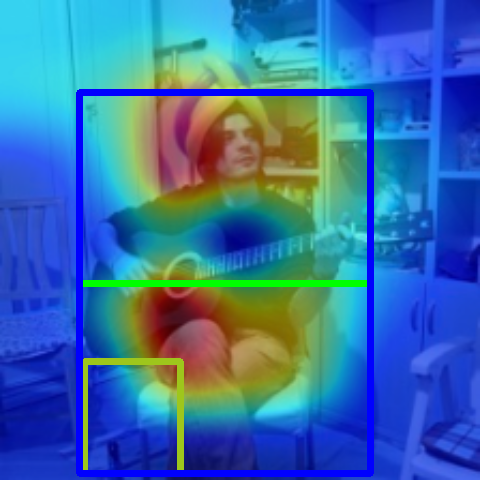}
\includegraphics[width=0.32\textwidth] {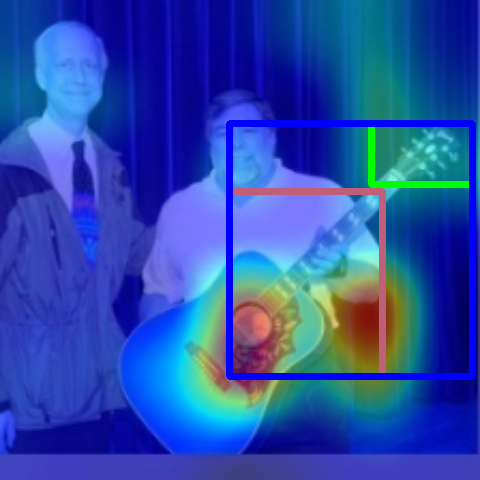}
\includegraphics[width=0.32\textwidth] {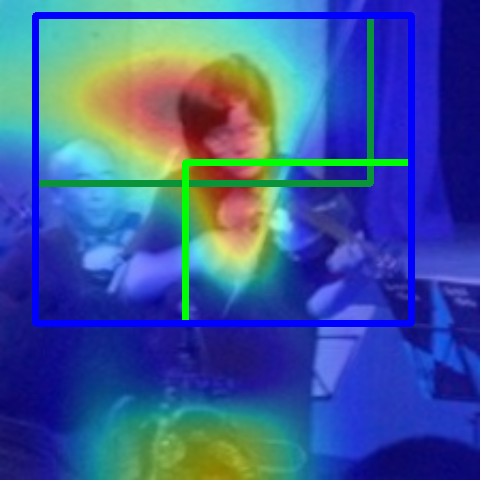}
\end{minipage}
}\hfill
\subfloat[talking left, talking to passenger, texting right]{
 \begin{minipage}[c][0.65\width]{0.31\textwidth}
\centering
\includegraphics[width=0.32\linewidth] {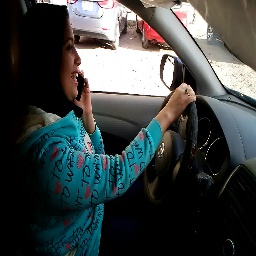}
\includegraphics[width=0.32\textwidth] {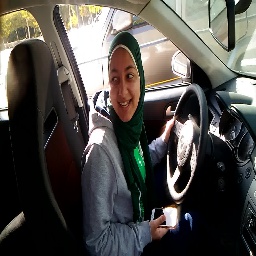}
\includegraphics[width=0.32\textwidth] {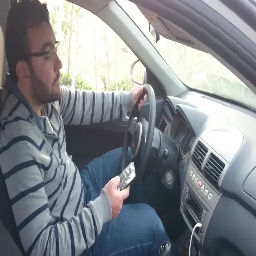}\\\vspace{1mm}
\includegraphics[width=0.32\textwidth] {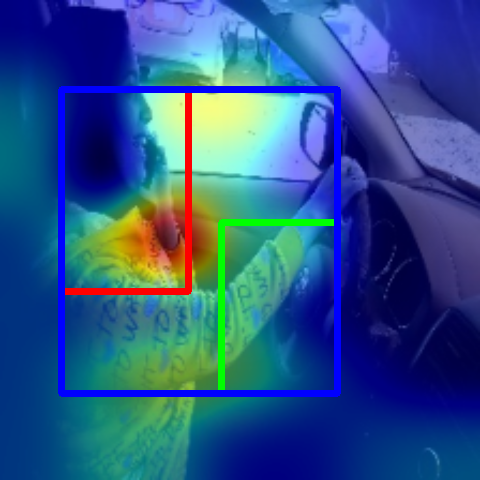}
\includegraphics[width=0.32\textwidth] {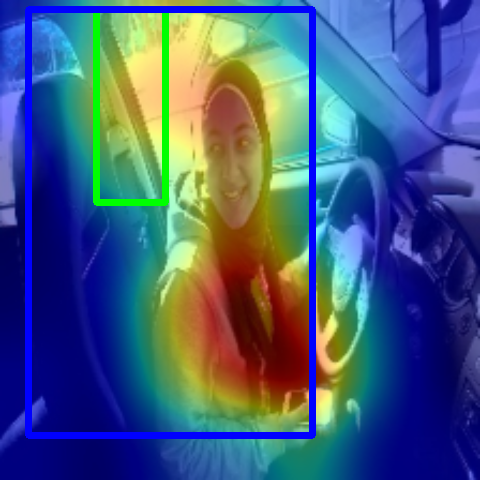}
\includegraphics[width=0.32\textwidth] {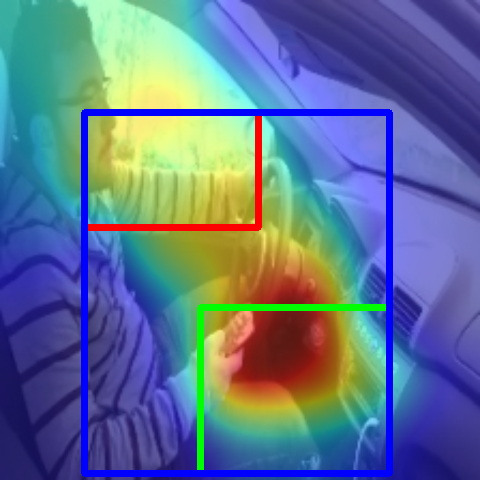}
\end{minipage}
}\hfill
 \subfloat[Stanford-40: blowing bubbles]{
 \begin{minipage}[c][0.65\width]{0.31\textwidth}
	   \centering
 \includegraphics[width=0.32\linewidth] {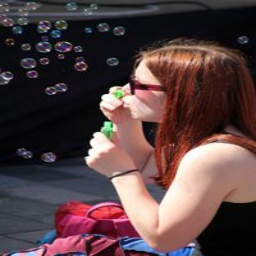} 
\includegraphics[width=0.32\linewidth] {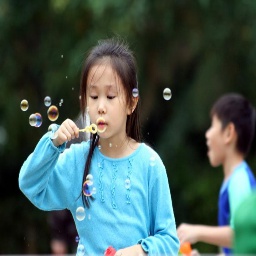}
\includegraphics[width=0.32\linewidth] {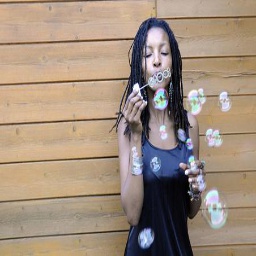} \\ \vspace{1mm}
\includegraphics[width=0.32\textwidth] {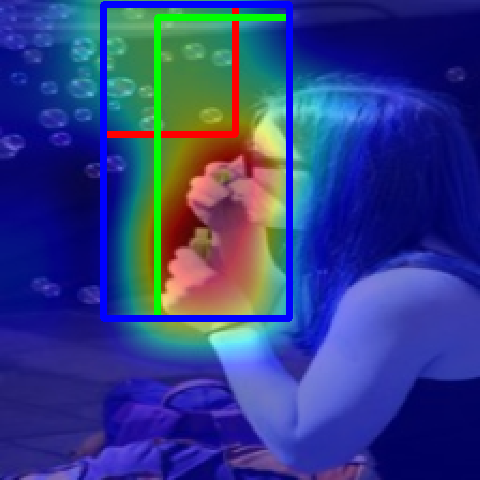}
\includegraphics[width=0.32\textwidth] {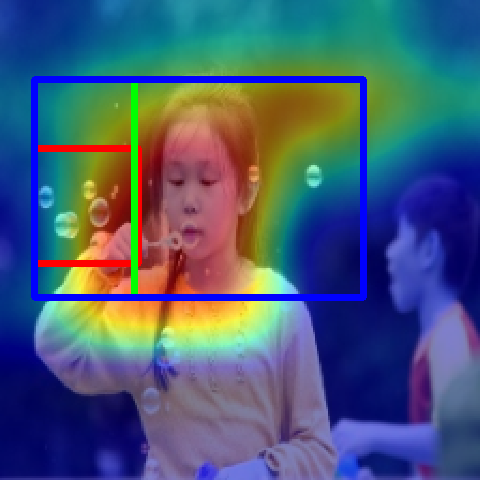}
\includegraphics[width=0.32\textwidth] {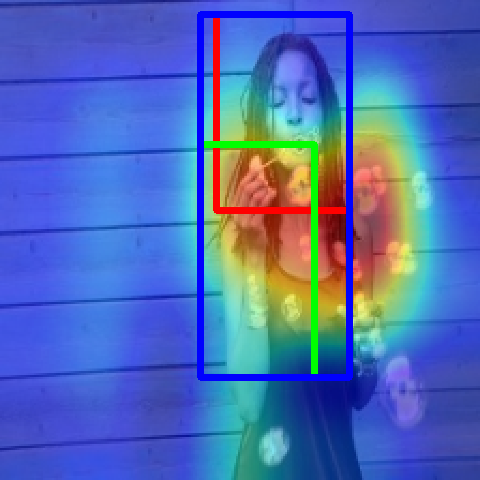}
\end{minipage}
}\hfill
\subfloat[PPMI-24: playing Guitar] {
 \begin{minipage}[c][0.65\width]{0.31\textwidth}
	   \centering
\includegraphics[width=0.32\linewidth] {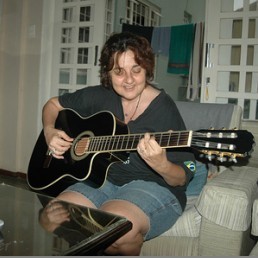} 
\includegraphics[width=0.32\linewidth] {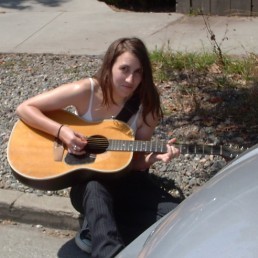}
\includegraphics[width=0.32\linewidth] {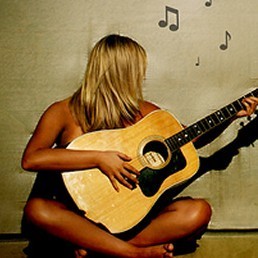}\\ \vspace{1mm}
\includegraphics[width=0.32\textwidth] {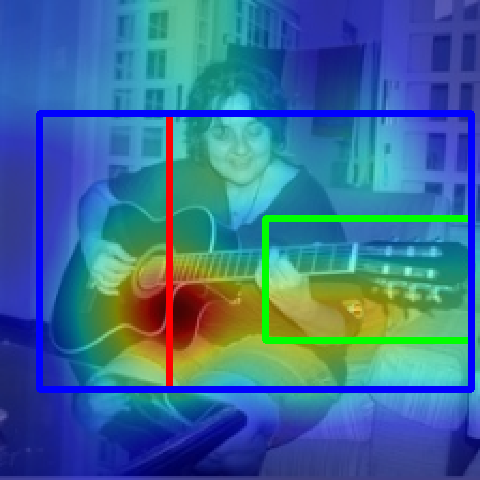}
\includegraphics[width=0.32\textwidth] {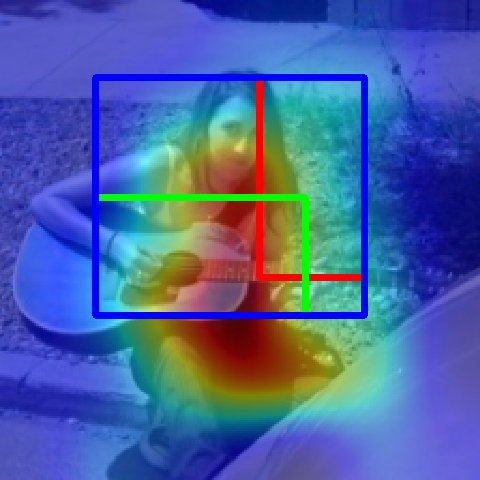}
\includegraphics[width=0.32\textwidth] {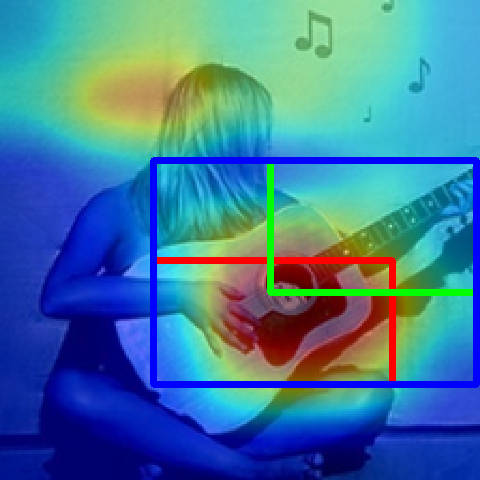}
\end{minipage}
}\hfill
\subfloat[AUC-V2: drinking]{
 \begin{minipage}[c][0.65\width]{0.31\textwidth}
\centering
\includegraphics[width=0.32\linewidth] {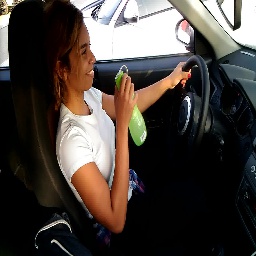}
\includegraphics[width=0.32\textwidth] {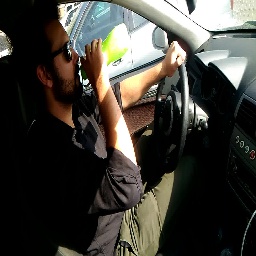}
\includegraphics[width=0.32\textwidth] {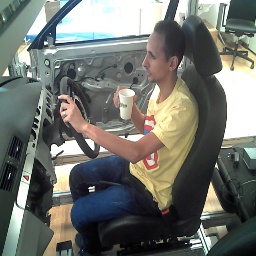}\\\vspace{1mm}
\includegraphics[width=0.32\textwidth] {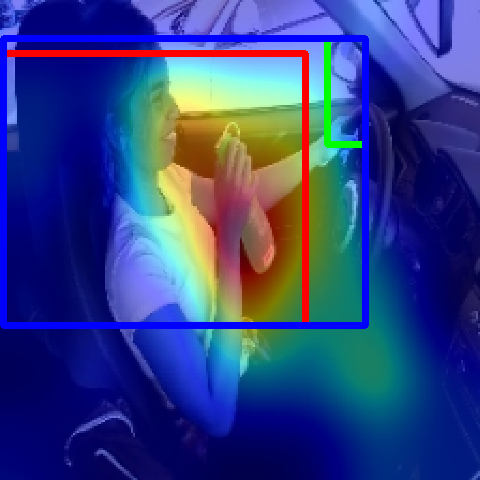}
\includegraphics[width=0.32\textwidth] {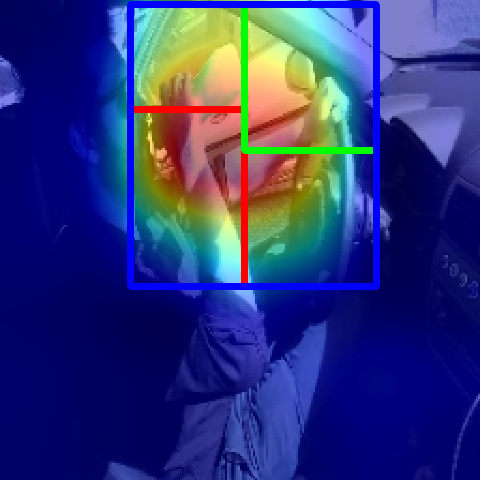}
\includegraphics[width=0.32\textwidth] {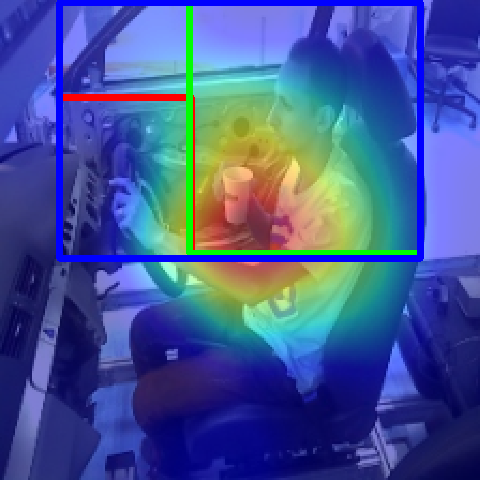}
\end{minipage}
}\hfill
\subfloat[Caltech-256: Owl, Toad, Umbrella] {
 \begin{minipage}[c][0.65\width]{0.31\textwidth}
	   \centering
\includegraphics[width=0.32\linewidth] {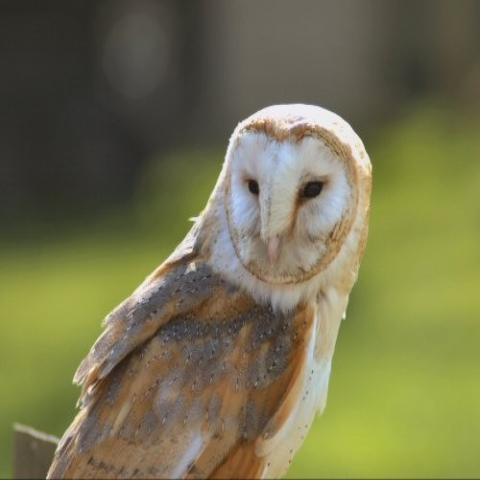} 
\includegraphics[width=0.32\linewidth] {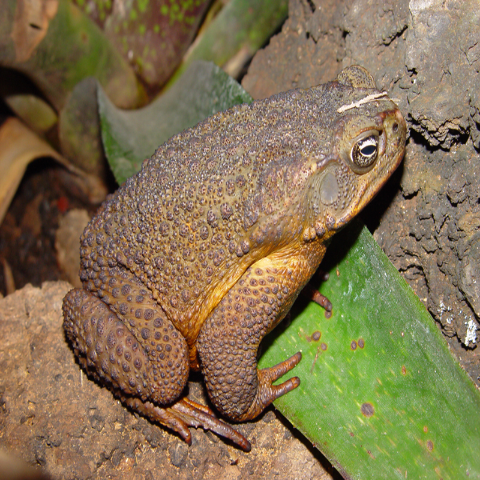}
\includegraphics[width=0.32\linewidth] {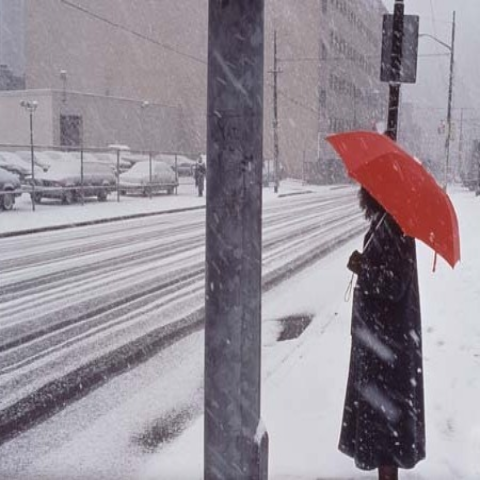} \\ \vspace{1mm}
\includegraphics[width=0.32\textwidth] {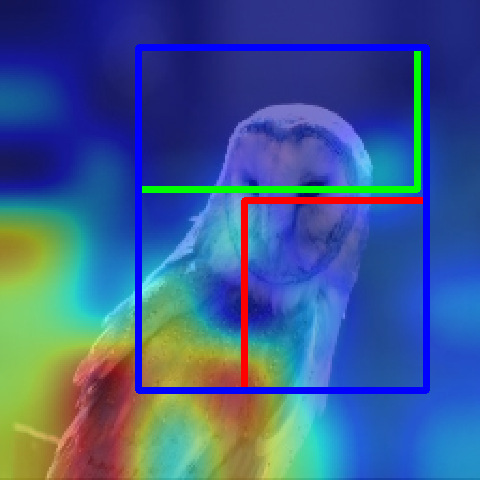}
\includegraphics[width=0.32\textwidth] {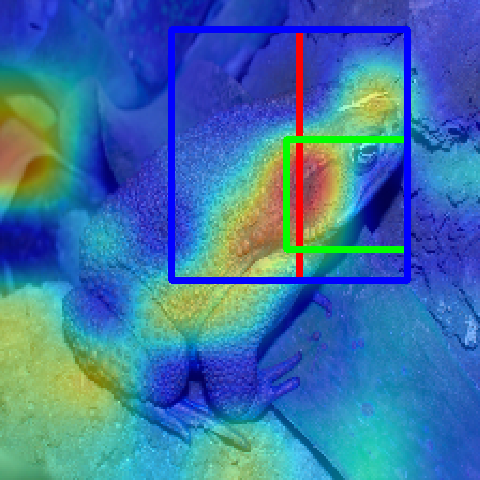}
\includegraphics[width=0.32\textwidth] {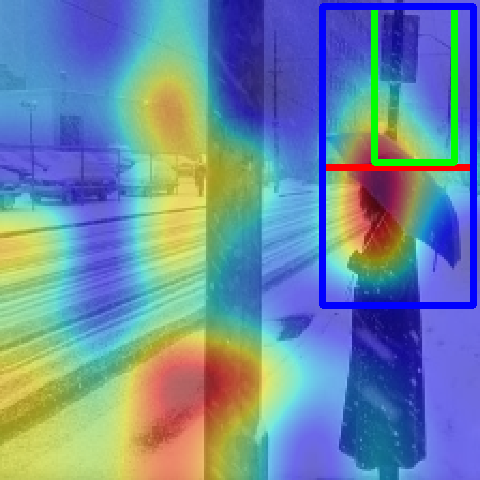}
\end{minipage}
}\hfill
\subfloat[Caltech-256: Zebra] {
 \begin{minipage}[c][0.65\width]{0.31\textwidth}
	   \centering
\includegraphics[width=0.32\linewidth] {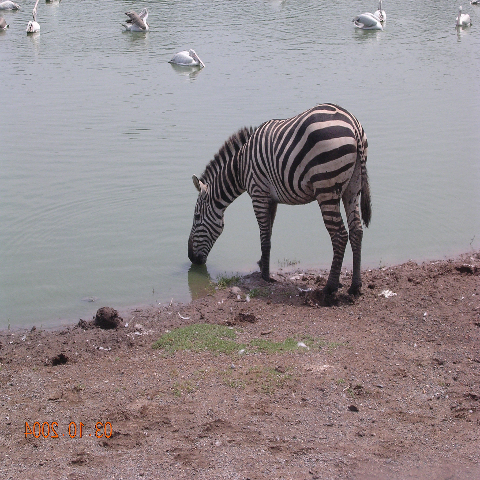} 
\includegraphics[width=0.32\linewidth] {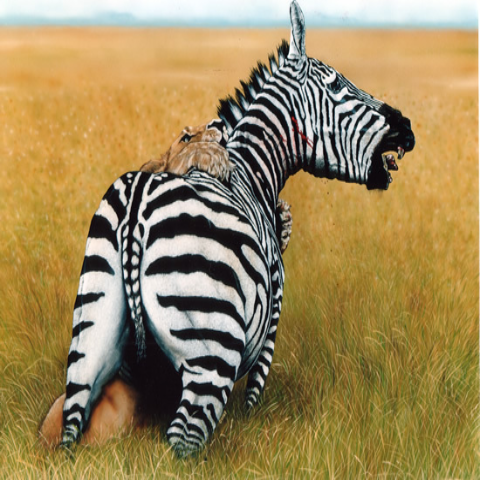}
\includegraphics[width=0.32\linewidth] {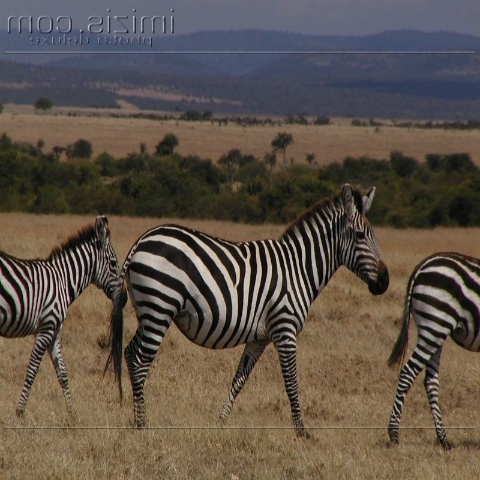} \\ \vspace{1mm}
\includegraphics[width=0.32\textwidth] {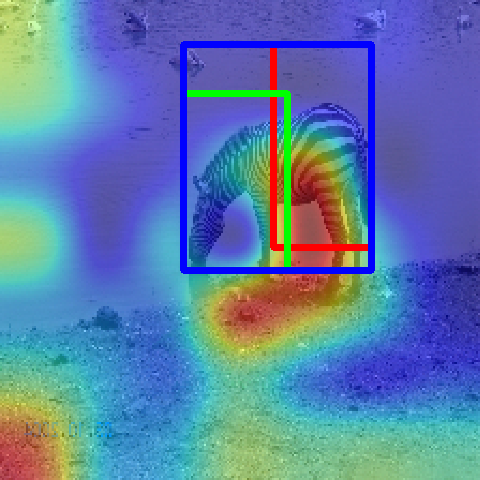}
\includegraphics[width=0.32\textwidth] {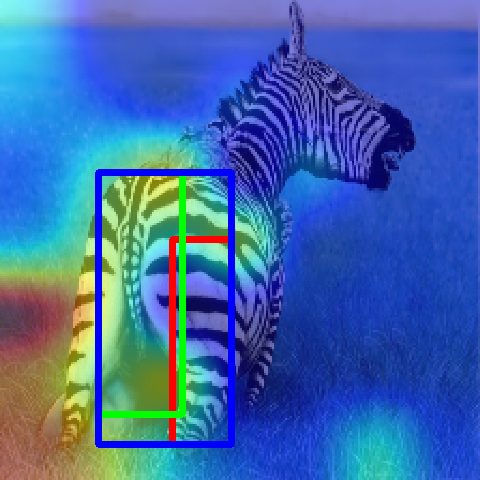}
\includegraphics[width=0.32\textwidth] {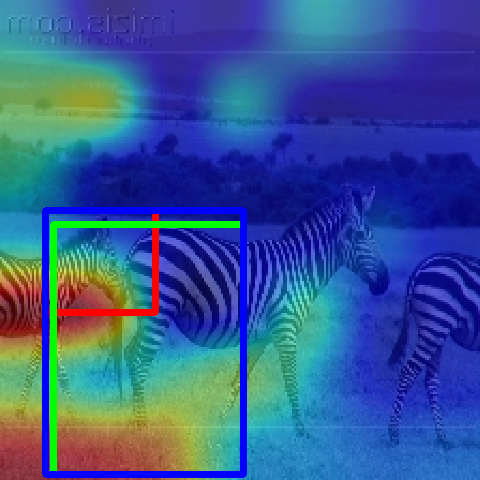}
\end{minipage}
}\hfill
\subfloat[Food-101: Ice-cream] {
 \begin{minipage}[c][0.65\width]{0.31\textwidth}
	   \centering
\includegraphics[width=0.32\linewidth] {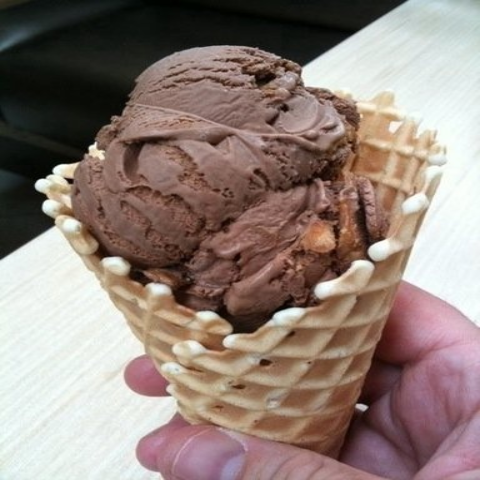} 
\includegraphics[width=0.32\linewidth] {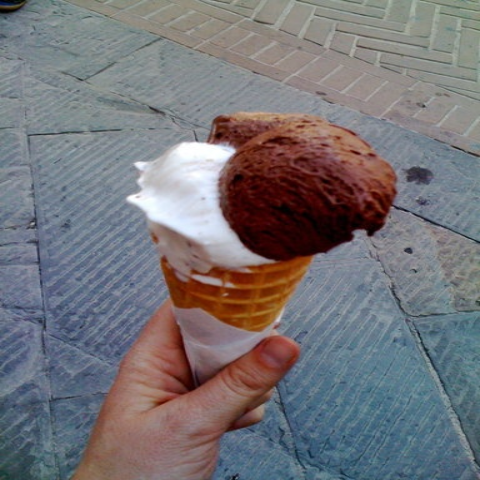}
\includegraphics[width=0.32\linewidth] {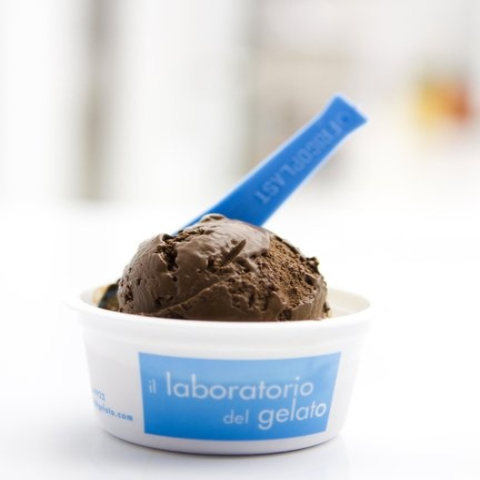} \\ 
\vspace{1mm}
\includegraphics[width=0.32\textwidth] {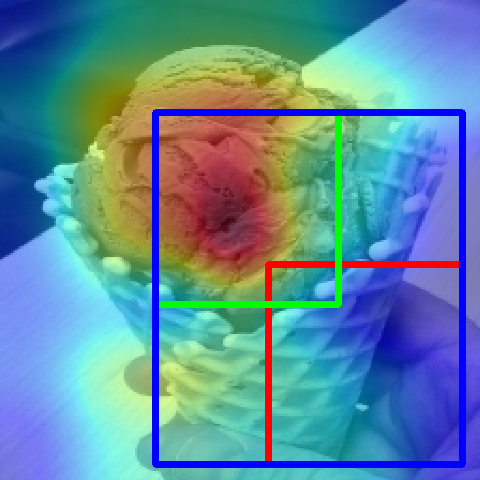}
\includegraphics[width=0.32\textwidth] {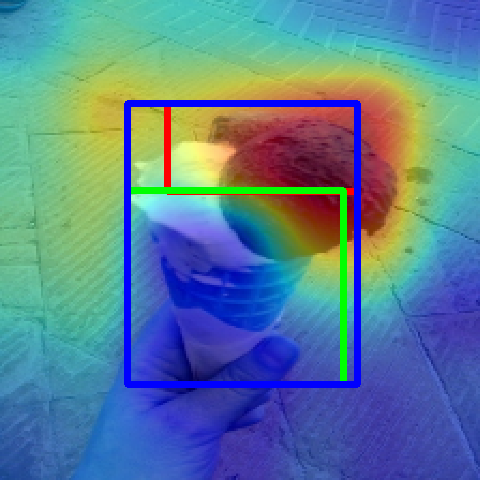}
\includegraphics[width=0.32\textwidth] {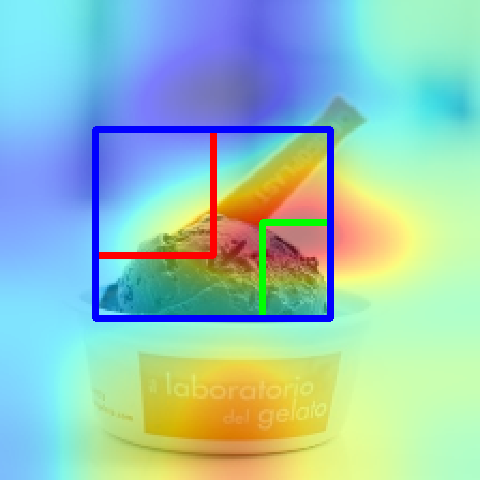}
\end{minipage}
}
\caption{Visualization of class activation maps using the Gradient-weighted Class Activation Mapping (Grad-CAM) \cite{ selvaraju2017grad} are illustrated on five datasets: (a, d) Stanford-40 \cite{YaoJKLGF11}, (b, e) PPMI-24 \cite{YaoF10}, (c, f) AUC-V2 \cite{H.M.Eraqi}, (g, h) Caltech-256 \cite{griffin2007caltech}, and (i) Food-101 \cite{BossardGG14}.
The top row describes three distinctive actions from: (a) Stanford-40, (b) PPMI-24, and (c) AUC-V2. The middle row depicts three examples of the same action  per dataset to demonstrate the effectiveness of the proposed SRs: (d) blowing bubble (Stanford-40), (e) playing Guitar (PPMI-24), and (f) drinking (AUC-V2). The last row shows (g) three generic object classes (Caltech-256), (h) three examples of the same object class Zebra (Caltech-256), and (i) three examples of the same ice-cream class (Food-101). Each example contains the original image (top) and corresponding activation map of salient regions (bottom) on which we have overlaid only three SRs for clarity. It contains two primary SRs (enclosed with \textcolor{red}{red} and \textcolor{green}{green} bounding-boxes) along with a secondary region (enclosed with \textcolor{blue}{blue} bounding-box) which is derived from those two primary SRs. More visualizations are provided in the supplementary document. Best view in color.}
\label{fig:GRAD_CAM}
\vspace{-1em}
\end{figure*}
\subsection{Performances on Stanford-40 and PPMI-24}
Most of the existing approaches use the provided bounding-box annotations while experimenting on Stanford-40. Whereas, our approach does not consider the bounding-box annotations at all; instead, it relies on the automatic detection of the proposed SRs. It has attained 97.83\% accuracy and 96.21\% mAP (Table \ref{table:Sota_comparison}). 
Our model has also improved mAP by 2.15\% compared to the SotA \cite{liu2018loss} that considers human mask loss using Inception-ResNet-v2 as a base network. The multi-branch attention model \cite{YanSLZ18} has achieved the best mAP of 90.70\% with the use of bounding-box information, whereas it is degraded to 85.20\% without it. Thus, our novel AG-Net has gained a significant margin (11.01\% in mAP) compared to \cite{YanSLZ18} by avoiding manually annotated data. The AP of each action in Stanford-40 is shown in Fig. \ref{fig:S_40_class}. 
The best AP of the color fusion model \cite{Lavinia2019NewCF} is 97.42\% for \emph{playing guitar}, whereas our method attains 99.47\% for the same. The AP of \emph{waving hand} in our proposal is 90.45\% which is also an improvement over the human mask loss (76.70\%) \cite {liu2018loss}, and multi-branch attention ($<$75.00\%) \cite{YanSLZ18}. It is clear from Fig. \ref{fig:S_40_class} that APs of the right-most nine actions (from \emph{playing violin} to \emph{looking through microscope}) are of 100\%. In Fig. \ref{fig:S_40_AP_comparison}, a comparative performance analysis is made using the top-20 actions as mentioned in \cite{Lavinia2019NewCF}. It is clear that our approach has produced significant improvement in those actions. 

The results on the PPMI-24 dataset are provided in Table \ref{table:Sota_comparison}. Our model achieves an mAP improvement of 16.05\% over the best SotA \cite{YanSZ17}. Though Zhao et al. \cite{ZhaoMC17} have achieved 92.90\% mAP for 12-class binary classification using the generalized symmetric pair model (GSPM), they have attained only 51.70\% mAP for 24-classes. In \cite{Hu2013recognising}, the exemplar-model has achieved 49.34\% accuracy. In their experiments, they have manually annotated the objects and six body parts. The AP of each interaction is shown in Fig. \ref{fig:PPMI_class}. For each interaction, our method outperforms the mutual context model \cite{yao2012recognizing}, as well as the color fusion model \cite{Lavinia2019NewCF}) by a significant margin. The right-most nine actions in Fig. \ref{fig:PPMI_class} (from \emph{play saxophone} to \emph{play violin}) have produced 100\% AP using our AG-Net. This shows the advantages of the proposed AG-Net. The confusion matrices of both datasets are provided in the supplementary document.
\subsection{Performance on Food-101}
Many SotA approaches use both the top-1 and top-5 accuracy while experimenting with this dataset. We follow the same. The result is shown in Table \ref{table:Sota_comparison}. The top-1 and top-5 accuracies attained by our AG-Net are 99.30\% and 99.87\%, respectively. This clearly shows that the proposed method outperforms the SotA by a significant margin. This justifies our proposed AG-Net that focuses on SRs to guide the network to attend more important regions for effective recognition. Our approach has gained 6.30\% top-1 accuracy improvement over GPipe \cite{GPipeFood}, and 1.16\% top-5 accuracy than WISeR \cite{MartinelFM18}. The authors in \cite{LiuXWL16} have achieved a 0.20\% improvement in accuracy with three attention layers (86.50\%) than with two attention layers (86.30\%). Our method has also attained 100\% top-1 accuracy for the best 41 food categories. The worst top-1 accuracy of 30 food categories is shown in Fig. \ref{fig:Food_class}, implying that the top-1 accuracy of \emph{steak} is 90.40\% and that of \emph{chicken curry} is 99.20\%. We also provide the mAP (98.93\% in Table \ref{table:overall}) value as an additional metric. The average precision (AP) of each food category is shown in the supplementary document.
\subsection{Performance on Caltech-256}
The accuracy of our AG-Net is 96.89\% on this diverse generic object recognition dataset. Like over other datasets, it significantly outperforms SotA methods (Table \ref{table:Sota_comparison}). It achieves 2.59\% gain over the previous best accuracy (94.30\%) \cite{ge2019weakly} that uses selective-joint fine-tuning with additional secondary training data from ImageNet. Whereas, their accuracy is 93.50\% with only the target data. A transfer learning-based regularization is performed in \cite{xuhong2018explicit} which achieves 87.90\%. AG-Net gives 5.54\% gain over a recent work \cite{zhang2019multi} that jointly learns visual, semantic and view consistency (VSVC).    
\begin{table}
\centering      
  \caption{Baseline Accuracy (\%) using only SotA backbone CNNs}
  \label{tab:commands}
  \begin{tabular}{l| c c c c  c}
    \toprule
    Base CNN & AUC-V1 & AUC-V2 & Stanford  & PPMI &  Food \\
    \midrule
    ResNet-50 & 90.63 &74.10 &76.46 &75.33 &80.04  \\ 
    Inception-V3 &94.70 & 69.56 &70.94 & 69.13 &78.34 \\
    DenseNet-121 &95.00 & 77.65 &78.64 & 81.29 &82.70\\
    NASNet-M &94.00 & 58.38 &75.20 & 70.41 &79.16  \\
    VGG-16 & 94.80 & 76.13 &63.00 & 72.78 &74.93  \\
    \bottomrule
  \end{tabular}
  \label{table:BL_CNN}
\end{table}
\begin{table}
\centering        
  \caption{Model Complexity using 36 SRs and with (+Attn) or without (-Attn) over Baseline (BL). It is Shown as Trainable Parameters in Millions (M), GFLOPS in Billions (B), Per-image Inference Time in Milliseconds (ms) and  Respective Training Time in Hours (hrs) using Different Datasets for 50 Epochs. Our AG-Net's Complexity is Compared to Different Base CNNs}
  \label{tab:commands}
  \begin{tabular}{l| p{0.5cm} p{0.7cm} p{0.68cm}| p{0.8cm} p{0.5cm} p{0.4cm}}
    \toprule
Model & Param & GFLOPs & Infer. & \multicolumn{3}{c} {Training time (hrs)}  \\
     & (M) & (B)  & (ms)  &   Stanford  &  PPMI & V2 \\

    \midrule
    ResNet-50 (BL) &23.62 & 7.77 &2.33 & 0.44   & 0.26    &1.56 \\
    AG-Net (-Attn) & 43.02 &9.36 & 4.45 & 2.70   &   1.59    &7.16\\
    AG-Net (+Attn) & 54.79 &10.42 & 5.20 & 2.88   &  1.69  &7.30\\
\midrule
    DenseNet-121 (BL) &6.99 & 5.68 &2.58 & 0.65   &   0.39   &2.04 \\
    AG-Net (-Attn) & 11.84 &5.71 & 4.68 & 2.90  & 1.90     &7.74\\
    AG-Net (+Attn) & 16.42 &6.10 & 4.81 & 3.06 &   1.97   &8.47\\
\midrule
    Inception-V3 (BL) &21.85 & 5.72 &2.43 & 0.54  & 0.32   &1.63 \\
    AG-Net (-Attn) & 41.25 &5.81 & 4.83 & 2.74 &  1.83   &7.76\\
    AG-Net (+Attn) & 53.02 &6.88 & 5.76 & 2.99 &  1.92    &7.97\\
\midrule 
    NASNet-M (BL) &4.28 & 1.15 &2.89 & 1.29   &  0.78   &4.04 \\
    AG-Net (-Attn) & 9.43 &1.17 & 4.63 & 3.05 &  1.95   &7.99\\
    AG-Net (+Attn) & 14.19 &1.58 & 5.74 & 3.21  &  2.10   &8.82\\
    \bottomrule
  \end{tabular}
  \label{table:complexity}
\end{table}
\subsection{Performance Summarization and Model Complexity}
The proposed AG-Net is an end-to-end pipeline in which SRs are generated on-the-fly by considering the spatial distribution of salient keypoints during the image augmentation process. These SRs are then used to pool features from the output of a base CNN and then attentionally aggregated (learning joint relationships) to generate a global descriptor, which is able to emphasize the importance of an SR by focusing on the misalignment of local parts or pattern. The SRs detection process is simple and computationally inexpensive. Our AG-Net has achieved SotA performances on all of the aforesaid six datasets consisting of diverse visual recognition tasks - human-objects interactions to fine-grained action to food image to varied object categories. This justifies the novelty and effectiveness of the proposed approach in recognizing diverse visual content.

The recent deep learning based approaches are significantly better than the earlier methods (Table \ref{table:Sota_comparison}) focusing on engineered keypoints-based local feature descriptors. In Table \ref{table:Sota_comparison}, these keypoints-based approaches are marked with $\textcolor{blue}\dagger$. In this work, we unify both deep learning and keypoints-based approaches in an innovative way to capture the contextual information representing fine-to-coarse changes in image content to enhance the discriminability potential of a SotA base CNN. As a result, our approach can be applied to diverse visual recognition tasks, as shown in Table \ref{table:V1-V2} and \ref{table:Sota_comparison} and has outperformed the SotA approaches over all the six different datasets. This signifies the effectiveness and competence of the proposed approach. To justify the wider applicability of our approach, we evaluate our AG-Net with different SotA backbone CNNs such as ResNet-50 \cite{he2016deep}, Inception-V3 \cite{szegedy2016rethinking}, DenseNet-121 \cite{huang2017densely},  NASNet-Mobile \cite{zoph2018learning} and VGG-16 \cite{simonyan2014very}. The performance is shown  in Table \ref{table:abl_CNN}. It is evident that the performance is very consistent on all backbones and is significantly better than the previous best. The accuracy using ResNet-50 and DenseNet-121 as a backbone is slightly better than the rest and could be due to the nature of their design. The performance of AG-Net using VGG-16 backbone is better than the existing methods that used VGG-16 base CNN on diverse datasets. For example, on AUC-V1, VGG-16 with regularization method \cite{BahetiGT18} has achieved 96.31\% and our AG-Net has attained 99.65\%. Also, we have reached 95.16\% on AUC-V2 compared to 76.13\% in \cite{H.M.Eraqi} that uses VGG-16 (Table \ref{table:V1-V2}).  Moreover, AG-Net attains 91.72\% mAP and 95.15\% accuracy on Stanford-40 using VGG-16 backbone. Whereas, multi-branch attention method in \cite{YanSLZ18} has achieved 90.70\% mAP on this dataset with leveraging bounding box annotations. However, their mAP is 85.20\% without annotation and is comparable to our experimental setup, i.e. without using the bounding box annotations. The keypoint-based VLAD \cite{YanSZ17} method has attained 88.50\% mAP on Stanford-40. This method has gained 81.30\% mAP on PPMI-24, and our method has achieved 92.51\% on PPMI-24 (Table \ref{table:Sota_comparison}). MSMVFA \cite{JiangMLL20} has reported 87.68\% top-1 accuracy on Food-101 using VGG-16 backbone. On the contrary, AG-Net has gained 98.07\% using the same. However, AG-Net with VGG-16 backbone requires comparatively more training time (extra 100 epochs) than other backbone CNNs. Moreover, the model complexity is also increased and is due to the architectural design of the VGG-16. The baseline accuracy of these backbone CNNs is presented in Table \ref{table:BL_CNN}. For this, we use the standard \textit{transfer learning} by fine-tuning it on the target dataset using the same data augmentation and hyper-parameters (Section \ref{sec:learning}). It is obvious that the performance of these backbone CNNs has been  significantly enhanced (Table \ref{table:abl_CNN}) after our novel SRs and attention mechanism have been added. This suggests that these two components are key ingredients for achieving SotA performance over the six diverse datasets. 

We have analyzed the computational complexity of our AG-Net, regarding the number of trainable parameters, GFLOPs, training and inference time, as shown in Table \ref{table:complexity}. We have computed these parameters for our AG-Net with all aforementioned backbone CNNs, as well as without and with attention mechanism. For example, there are 23.62M (millions) trainable parameters, 7.77B GFLOPs (billions) and 2.33 milliseconds (ms) for per-image inference time using 12GB NVIDIA Titan V GPU for the ResNet-50 as a base CNN. For our AG-Net (with 36 SRs), without our attention mechanism (-Attn), these values are: 43.02M, 9.36B and 4.45ms. Similarly, these values for our AG-Net with attention (+Attn) are 54.79M, 10.42B, and 5.20ms, respectively. The model can be trained for 50 epochs with a batch size of 8 within 1.56 hours (hrs) for the ResNet-50 base CNN. For AG-Net with 36 SRs, the training time is 7.16hrs for -Attn, and 7.30hrs for +Attn using AUC-V2 dataset (12,555 training images). Likewise, these values for the Stanford-40 dataset (4K training images) are 0.44, 2.70 and 2.88 hrs, respectively. Overall, the per-image inference time of our AG-Net is faster than the existing works such as FCAN ($\sim$150ms) \cite{LiuXWL16} and CPM (27ms) \cite{ge2019weakly}. It reflects the computational efficiency of our AG-Net.
\subsection{Visualization using Class Activation Map}
In order to visualize our proposed SRs driven attention to guide the AG-Net, we use the Gradient-weighted Class Activation Mapping (Grad-CAM) \cite{selvaraju2017grad} on example images representing inter-class (Fig. \ref{fig:GRAD_CAM}a-c) and intra-class (Fig. \ref{fig:GRAD_CAM}d-f) variations from three  datasets. The visual map shows the most crucial regions for visual recognition. The feature maps from the \emph{5c-branch2c} layer of the ResNet-50 model have been used for visualization. Each action/object contains the actual image and its corresponding class activation map over which three important SRs (out of 36) are overlaid to justify the novelty of our SR formulation method. All SRs guide the proposed AG-Net to focus on regional contexts during learning. Fig. \ref{fig:GRAD_CAM} reveals that SRs generated in our approach also emphasize contextual features in a similar fashion like Grad-CAM for further refinement in attention mechanism. It also shows that the SRs are consistently detected which are overlaid on the attention maps of the intra-class images (Fig. \ref{fig:GRAD_CAM}d-f) justifying the effectiveness of our simple on-the-fly keypoints-based SRs generation process. 
\begin{figure*}
\centering
\subfloat [climbing] {\includegraphics[width=0.115\textwidth] {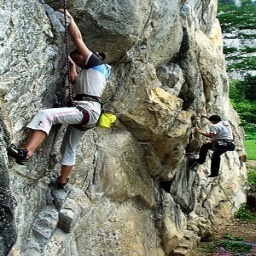} 
\includegraphics[width=0.115\textwidth]  {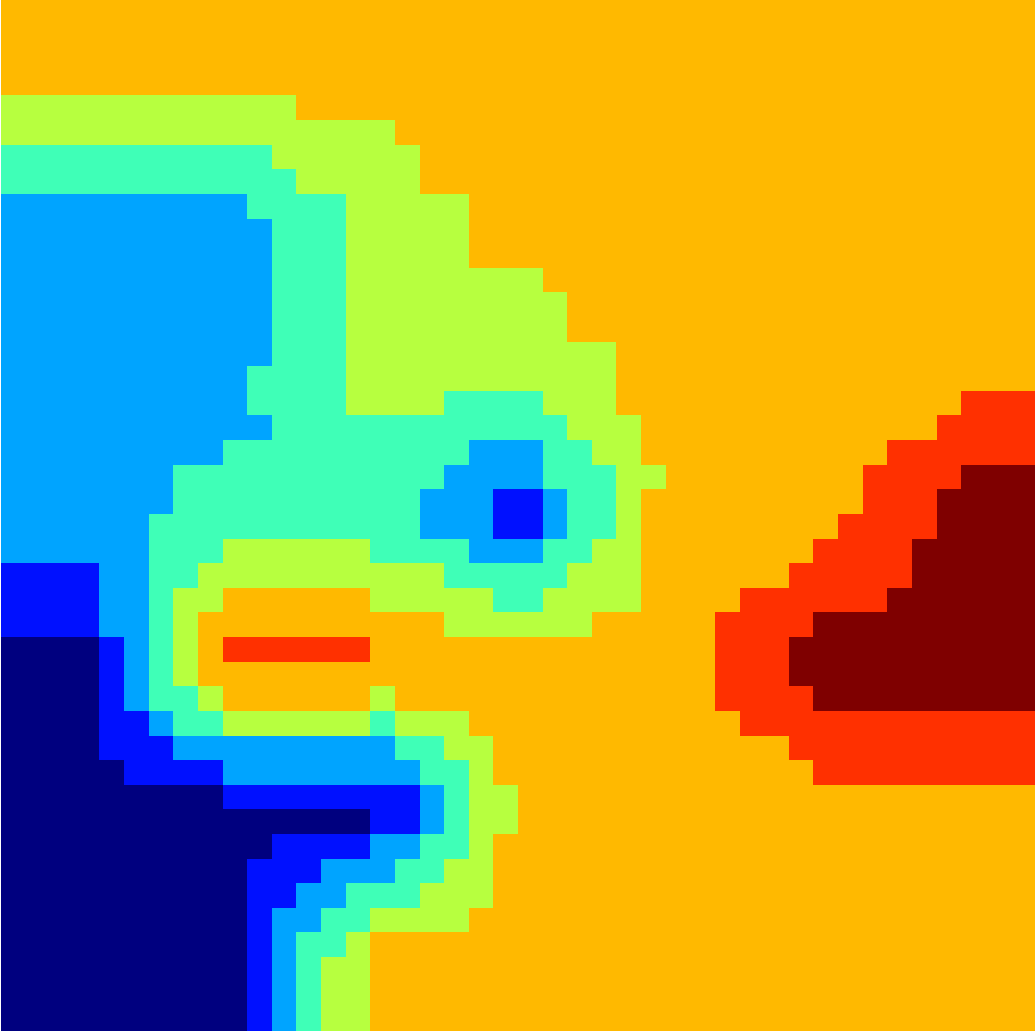}
\includegraphics[width=0.115\textwidth]  {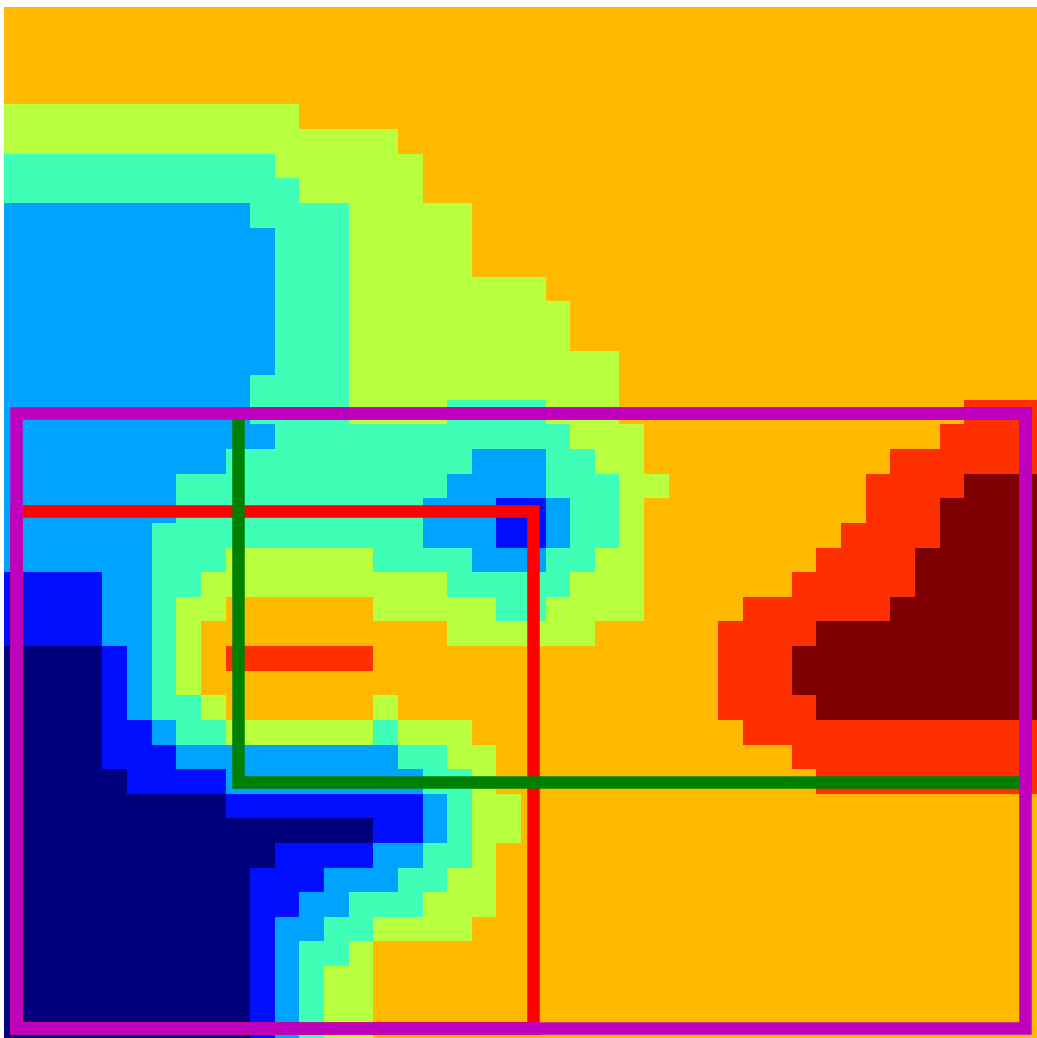}
\includegraphics[width=0.115\textwidth]  {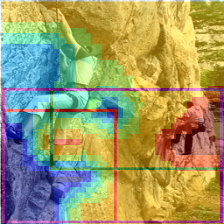} 
}\hfill
\subfloat [blowing bubble] {\includegraphics[width=0.115\textwidth] {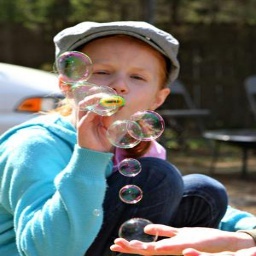} 
\includegraphics[width=0.115\textwidth]  {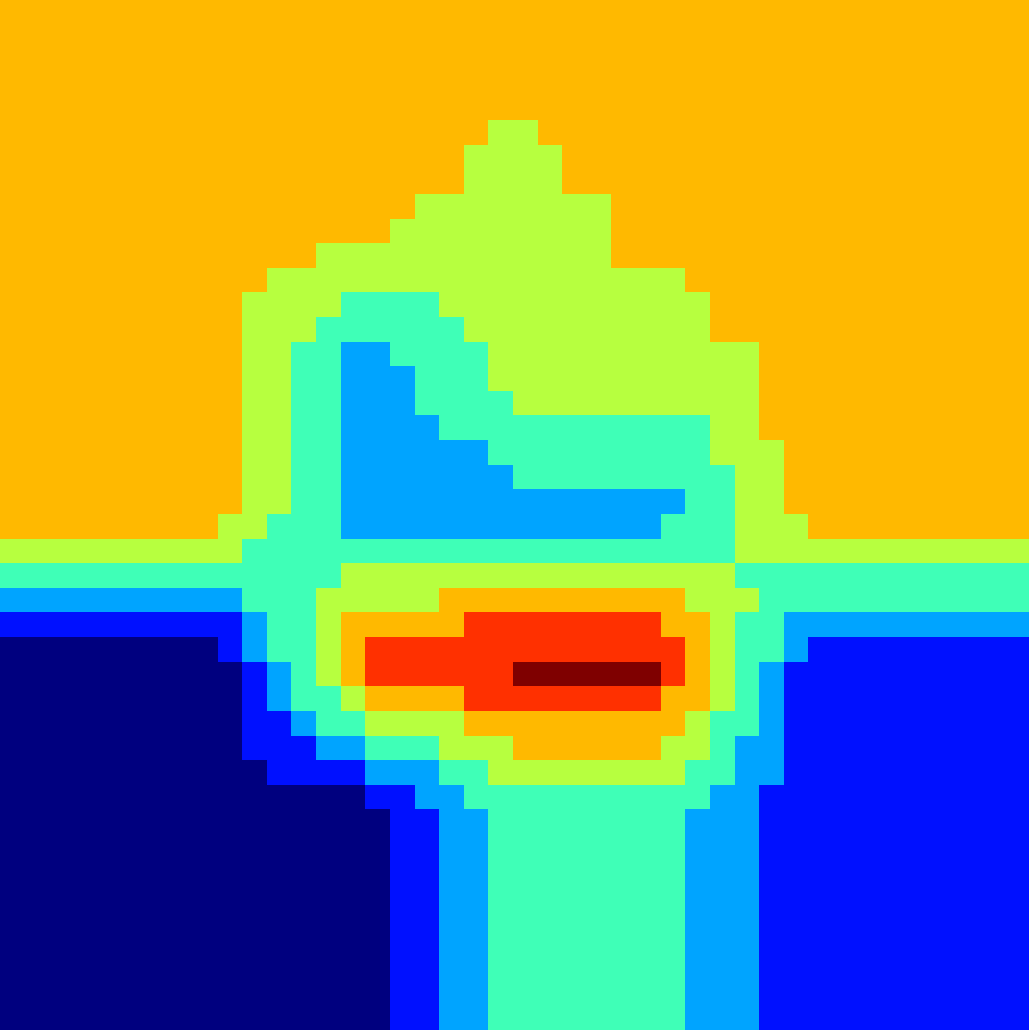}
\includegraphics[width=0.115\textwidth]  {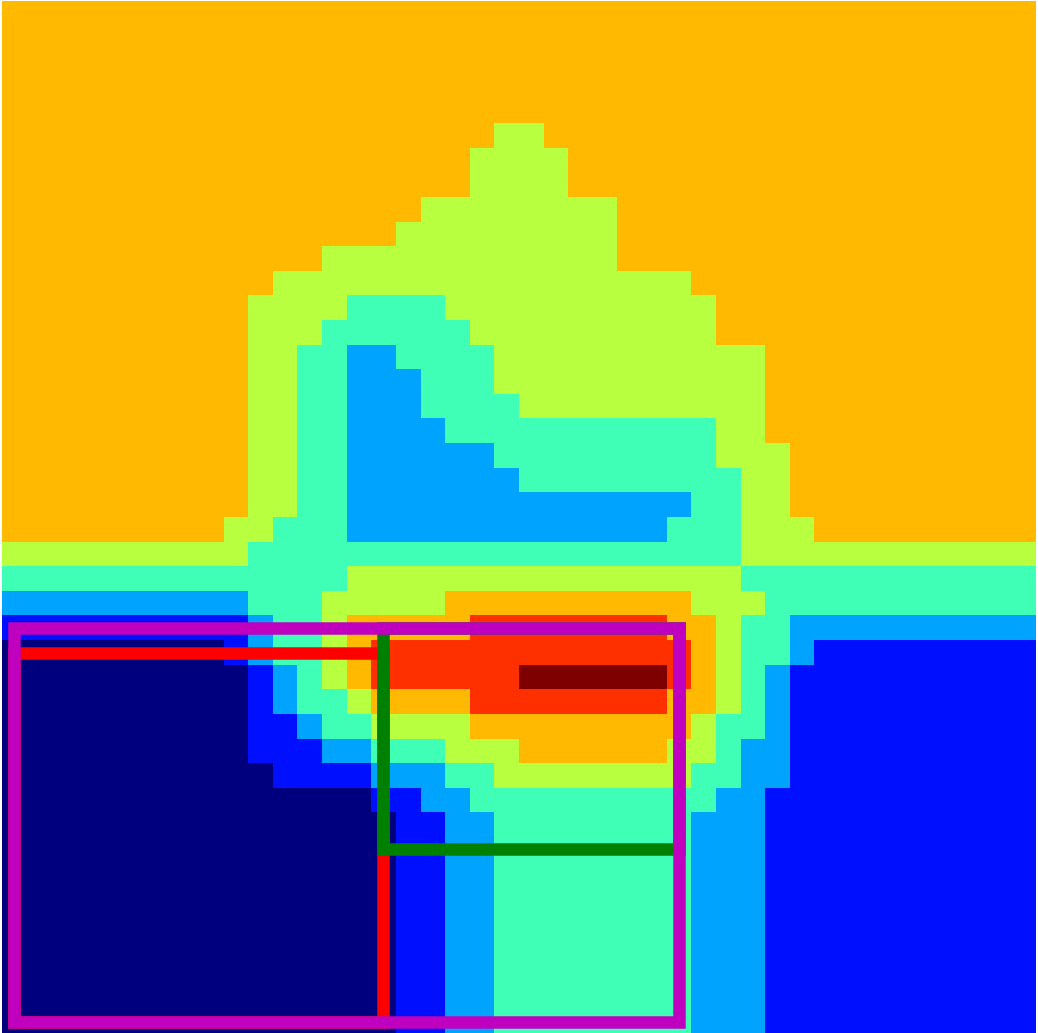}
\includegraphics[width=0.115\textwidth]  {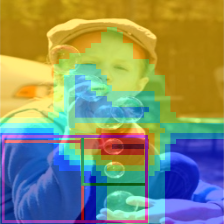}
}
\caption{Semantic regions (SRs) detection by clustering the CNN feature maps over the spatial dimension. Example images from Stanford-40 \cite{YaoJKLGF11} showing (a) climbing and (b) blowing bubble actions are given. The original image $\Rightarrow$ eight clustered primary SRs. For clarity, bounding boxes enclosing three SRs are provided; two primary SRs (\textcolor{green}{green} and \textcolor{red}{red}) and the corresponding derived secondary SR (\textcolor{magenta}{magenta}). These SRs are overlaid on the input image to demonstrate their effectiveness. 
Best view in color.}
\label{fig:CNN_cluster}
\vspace{-1em}
\end{figure*}
\section{Ablation Study} \label{Ablation}
We have also conducted a comprehensive ablation study on three datasets (Standford-40, PPMI-24, and AUC-V2) to analyze the effectiveness of the key components of our AG-Net, as well as the impact of the number of primary and secondary SRs, and their combination on the performance. All the training parameters and evaluation procedure are kept the same as described in Section \ref{sec:learning} above. Firstly, we evaluate the impact of the various key components on the proposed architecture. The results are given in Table \ref{table:abl_component}. It is evident that the main source of performance improvement is when our attention mechanism is combined with the generated SRs. For example, in the case of PPMI-24, the ResNet-50 base offers an accuracy of 75.33\%, which has been improved to 84.00\% with the 36 SRs. Next, the self-attention improves the accuracy up to 89.66\%, which is further significantly enhanced to 97.33\% with the inter-attention mechanism. Finally, the fusion of GAP and GMP in the classification layer improves the accuracy as 97.66\%. Thus, it is obvious that the SRs and the attention modules are the vital components for achieving SotA accuracy. 

Our SRs generation process is simple yet effective by considering the spatially distributed salient keypoints without requiring any object/part detectors, which are often explored by the existing approaches \cite{gkioxari2015actions,GkioxariGM15,mallya2016learning} to localize the bounding boxes of different body parts and objects. These bounding boxes are used as SRs for further processing to recognize actions and/or human-objects interactions. Therefore, to justify the appropriateness of our approach, we compare the performance of AG-Net considering the final SRs from mask R-CNN \cite{he2017mask} instead of generated by our keypoints-based approach. The mask R-CNN was trained on COCO dataset. The results are given in Table \ref{table:abl_component}.  
We have also experimented with SRs generated by clustering the convolutional features over the spatial dimension. The convolutional features from the last convolutional layer of the ResNet-50 backbone are used to generate the 8 primary and 28 secondary SRs like in the keypoints-based clustering using GMM (Section \ref{sec:sr}). Visual examples of such SRs generation method are shown in Fig. \ref{fig:CNN_cluster}. The results of our AG-Net using these SRs are given in Table \ref{table:abl_component} (last row) for the Stanford-40 (96.39\%), PPMI-24 (95.07\%) and AUC-V2 (87.39\%) datasets. It is found that the performance is better than the SRs generation process using the mask R-CNN (Stanford-40: 90.13\%, PPMI-24: 87.46\% and AUC-V2: 83.75\%). However, the performance using keypoints-based SRs is significantly better (Stanford-40: 97.83\%, PPMI-24: 97.66\% and AUC-V2: 94.56\%) than the CNN features clustering. This is mainly because the keypoints are of a large collection of local maxima or minima at each level of the pyramid of a given image and is well-known for identification tasks. Whereas, CNN features are more relevant to classification and categorization tasks since it has an excellent generalization ability. As a result, the SRs generated using keypoints represent more distinctive patches in comparison to CNN features, resulting in higher accuracy. Moreover, SRs' generation using keypoints is simple and fast, and is carried out on-the-fly in the data generator of our AG-Net. Whereas, the same process using CNN features requires offline practice as it requires a lot of processing power. This justifies our on-the-fly generation of the keypoints-based regions and their combination with our novel attention mechanism, resulting in effective captures of subtle visual variations by aggregating contextual features from most relevant image regions, and their importance in discriminating categories without requiring the objects/part detectors and/or distinguishable part annotations. 

Secondly, we study the impact of the number of primary and secondary SRs on our AG-Net's performance using ResNet-50 as a base CNN. The results are given in Table \ref{table:abl_SRsAll} in which the top-row and the middle-row reflect the accuracies (\%) using only primary SRs (P-SR) and secondary SRs (S-SR), respectively. The bottom-row contains combinations of both primary and secondary SRs. We limit the total number of SRs to a maximum of 36 to balance between the model's performance and complexity. In the top-row (P-SR), it is clear that the accuracy increases with the increasing number of primary SRs. However, the performance significantly improves in addition to secondary SRs (middle row). This suggests that the primary SRs only may not be sufficient to provide discriminative representation of the image content and justifies the significance of the secondary SRs. For example, if we consider the total number of SRs as 10, the accuracy for combining primary (4) and secondary (6) SRs (Stanford-40: 93.92\%, PPMI-24: 94.48\% and AUC-V2: 88.56\%) is significantly higher than using only 10 primary SRs (Stanford-40: 88.38\%, PPMI-24: 89.33\% and AUC-V2 85.79\%). It is worth to note that the model complexity is based on the total number of SRs. Thus, for the same number of SRs, the proposed combination of primary and secondary SRs are very useful in representing local-to-global structures in image content which guides the network to achieve significantly higher recognition accuracy. Interestingly, the accuracy increases with the increment of the respective primary and secondary SRs and gives the best accuracy (Stanford-40: 97.83\%, PPMI-24: 97.66\% and AUC-V2: 94.56\%) for the combination of 8 primary and 28 secondary SRs (unique pairs from 8 primary SRs). The performance decreases slowly with a further increment of the primary SRs with the total number of SRs fixed to 36. Therefore, 8 and 28 are the optimal number of the respective primary and secondary SRs. 
\begin{table}
\centering       
 \caption{Accuracy (\%) of AG-Net with Addition of Main Components  }
 \label{tab:commands}
 \begin{tabular}{l c c c}
\toprule
Components & Stanford-40  & PPMI-24 & AUC-V2\\
    \midrule
   Base ResNet-50 \cite{he2016deep} &76.46 & 75.33 &74.10\\
    +SRs(8 P-SRs + 28 S-SRs) &85.71  & 84.00 & 90.34\\
    +Intra-Attention & 87.53  & 89.66 &90.72 \\
     +Inter-Attention & 97.37 &97.33 &94.25\\
    + Classification (only GMP) & 97.46 &97.58  &94.54 \\
    + Classification (GAP \& GMP) &97.83 &97.66 & 94.56 \\
    \midrule
    SR (using Mask RCNN \cite{he2017mask}) & 90.13 & 87.46 & 83.75\\
    SR (using CNN feature clustering) & 96.39 & 95.07 & 87.39\\
    \bottomrule
  \end{tabular}
  \label{table:abl_component}
\end{table}
\begin{table}
\centering
  \caption{Accuracy (\%) of the AG-Net with Various Combinations of Primary (P-SR) and Secondary (S-SR) Semantic Regions. Total SR (T-SR) Are the Sum of Both P-SR and S-SR}
  \label{tab:commands}
  \begin{tabular}{c c c c| c c c}
    \toprule
   SR& P-SR & S-SR &T-SR & Stanford-40  & PPMI-24 & AUC-V2 \\
    \midrule
    \multirow{5}{*}{P-SR}& 4 &0  &4 &85.61 &86.70 &83.13 \\
     & 6 &0  &6 &85.56 &87.21 & 81.93 \\ 
   & 8 &0 &8 &88.67 & 87.67 &83.70 \\
    & 10 &0 &10 &88.38 & 89.33 & 85.79 \\
     & 16 &0 &16 &91.18 &90.51 &84.02 \\
    \midrule
   \multirow{7}{*}{S-SR} & 4 &6  &6 &91.96 &93.30 &85.97\\
    & 6 &15  &15 &96.04 &96.33 &92.95\\ 
    & 8 &28 &28 &97.18 & 97.70 &94.88 \\
     & 10 &26 &26 &96.90 & 97.70 &91.75\\
     & 16 &20 &20 &95.51 &96.33 &88.58\\
     & 10 &36 &36 &97.60 & 97.70 &93.37\\ 
     & 16 &36 &36 &97.09 &97.07 &94.51\\  
    \midrule
   \multirow{5}{*}{Both} & 4  &6  &10 &93.92 &94.48 & 88.56\\
     & 6  &15 &21 &96.61 &96.83 & 94.15 \\
    & \textbf{8}  &\textbf{28} &\textbf{36} &\textbf{97.83} &\textbf{97.66} & \textbf{94.56}\\
    & 10 &26 &36 &97.16 &97.51 & 92.16\\
   & 16 &20 &36 &96.51 & 97.02 & 88.62\\
    \bottomrule
  \end{tabular}
  \label{table:abl_SRsAll}
\end{table}
\section{Conclusion} \label{conclusion}
In this paper, a deep network called AG-Net with our novel attentional mechanism has been presented for visual recognition from still images. The SRs proposals are formulated using salient keypoints and their grouping using a GMM. The squeeze-and-excitation block with residual connection also enhances the channel-wise feature representation capability, and boost the performance with a little overhead. The attention module emphasizes various informative SRs to learn relevant region-level contexts attentively. The proposed method has been validated over six challenging diverse visual recognition datasets representing the driver action, human action, food dish, and generic object categories. These datasets are a mixture of fine-grained and distinct image categories for the visual classification task. In summary, our approach outperforms the existing SotA methods over these datasets by a significant margin. It shows the effectiveness of our robust proposal for both fine-grained and distinct visual classification.

The features representing SRs are extracted before the Softmax layer of a base CNN. Further enhancement could be achieved by combining SRs from multiple layers of the base CNN using an attention-driven approach. This would provide a better feature representation by learning to attend discriminative complementary information using SRs from multiple scales. In the near future, we plan to extend this salient region-based still image recognition to recognize human actions and activities in video sequences by modelling the spatiotemporal salient regions. 
\section*{Acknowledgement}
This research is supported by the UKIERI-DST grant CHARM (UKIERI-2018-19-10), and Research Investment Fund at Edge Hill University. The GPU used in this research is donated by the NVIDIA. We thank the Associate Editor and three anonymous reviewers for their constructive comments that have improved the quality of the paper.
\ifCLASSOPTIONcaptionsoff
  \newpage
\fi
%
\bibliographystyle{IEEEtran}
\bibliography{Ref}
\begin{IEEEbiography}[{\includegraphics[width=1in,height=1.25in,clip,keepaspectratio]{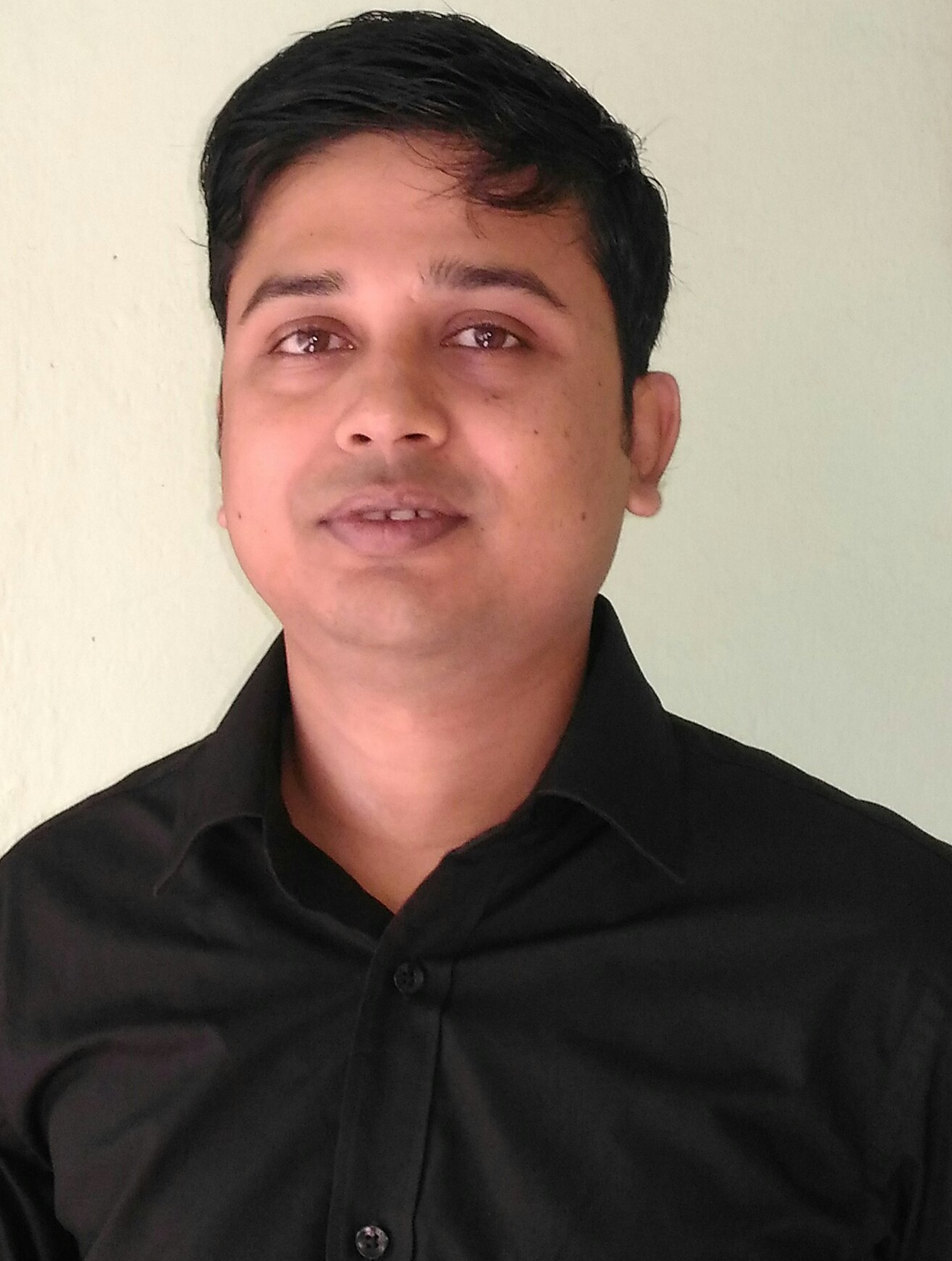}}] {Asish Bera} received the PhD degree from  the Jadavpur University, Kolkata, India, in 2018 and the M.Tech degree from from the IIEST Shibpur, India, in 2009. He is currently a Post-doctoral research associate at the Computer Science Department, Edge Hill University, UK. His current research interests include computer vision, deep learning, activity recognition, and biometrics. He is a member of IEEE.
\end{IEEEbiography}

\begin{IEEEbiography}[{\includegraphics[width=1in,height=1.25in,clip,keepaspectratio]{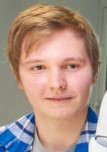}}]{Zachary Wharton} is currently an MRes student in the Department of Computer Science, Edge Hill University, UK. He obtained his Bachelor’s degree in Computing from Edge Hill University in 2019. His interests include computer vision, deep learning, human-robot interaction (HRI) and pattern recognition.
\end{IEEEbiography}

\begin{IEEEbiography}[{\includegraphics[width=1in,height=1.25in,clip,keepaspectratio]{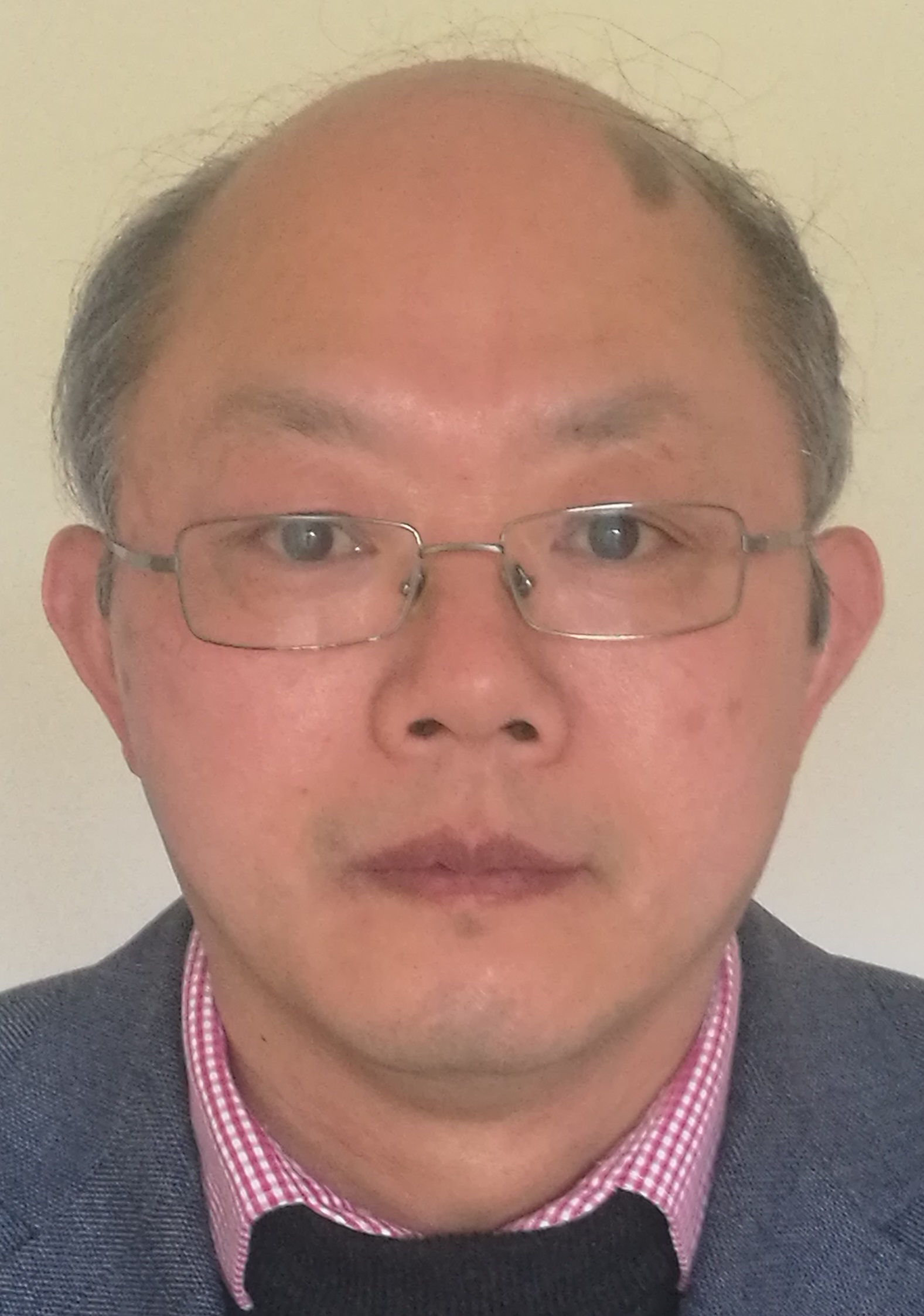}}]{Yonghuai Liu} is a professor and director of the Visual Computing Lab at Edge Hill University since 2018. He obtained his first PhD in 1997 from Northwestern Polytechnical University, P.R. China and second PhD in 2001 from The University of Hull, UK. He is an area/associate editor or editorial board member for a number of journals and conferences. He has published more than 180 papers in the top-ranked conferences and journals. His research interests lie in 3D computer vision, image processing, pattern recognition, machine learning, AI, and intelligent systems. He is a senior member of IEEE, Fellow of BCS, and Fellow of HEA of the UK.
\end{IEEEbiography}

\begin{IEEEbiography}[{\includegraphics[width=1in,height=1.25in,clip,keepaspectratio]{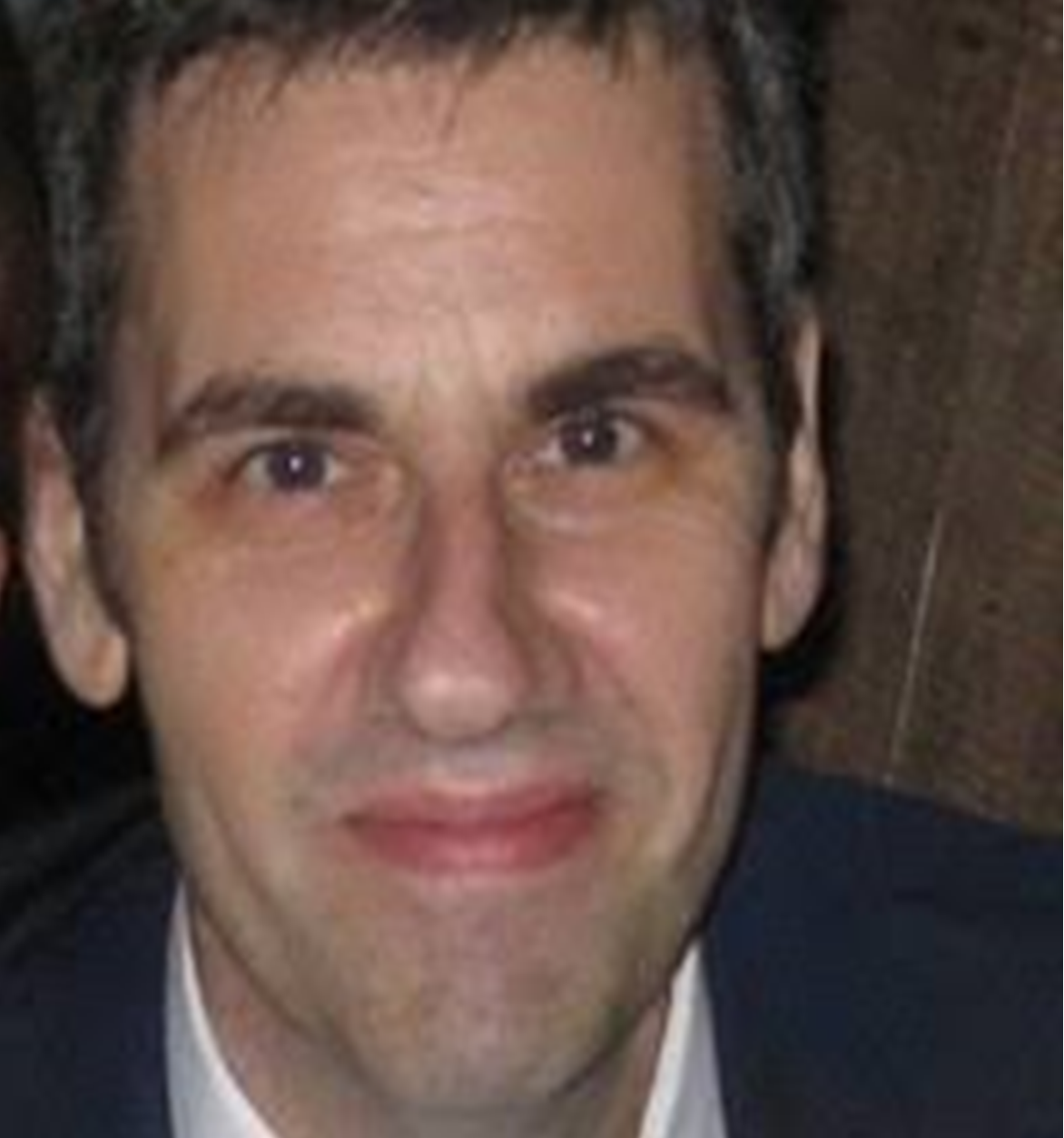}}]{Nik Bessis} received his BA from the T.E.I. Athens and his MA and PhD degrees from De Montfort University, UK. He is a full Professor (2010) and since 2015, the Head (Chair) of the Department of Computer Science at Edge Hill University, UK. He is a FHEA, FBCS and a senior member of IEEE. His research is on social graphs for network and big data analytics as well as developing data push and resource provisioning services in IoT, FI and inter-clouds. He is involved in a number of funded research and commercial projects in these areas. Prof Bessis has published over 300 works and won 4 best papers awards.
\end{IEEEbiography}

\begin{IEEEbiography}[{\includegraphics[width=1in,height=1.25in,clip,keepaspectratio]{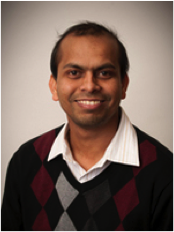}}]{Ardhendu Behera} received the PhD degree in Computer Science from the University of Fribourg, Switzerland and MEng degree in System Science and Automation from the Indian Institute Science (IISc) Bangalore, India. He is currently a Reader in Computer Vision \& AI in the Department of Computer Science, Edge Hill University, UK. He has worked as a Research Fellow and Senior Research Fellow in Computer Vision Group at the University of Leeds. He is a Fellow of HEA and member of {IEEE}, British Machine Vision Association, Applied Vision Association, British Computing Society, affiliated member of IAPR and ECAI. His main interests include computer vision, deep learning, human-robot social interaction, activity analysis and recognition.
\end{IEEEbiography}


\clearpage

\onecolumn

\begin{center}
\Large
\textbf{Supplementary Document}

\Large
\end{center}

\begin{figure*}[h]
\centering
\subfloat [Food-101: Apple-pie]
{\includegraphics[width=0.15\linewidth]{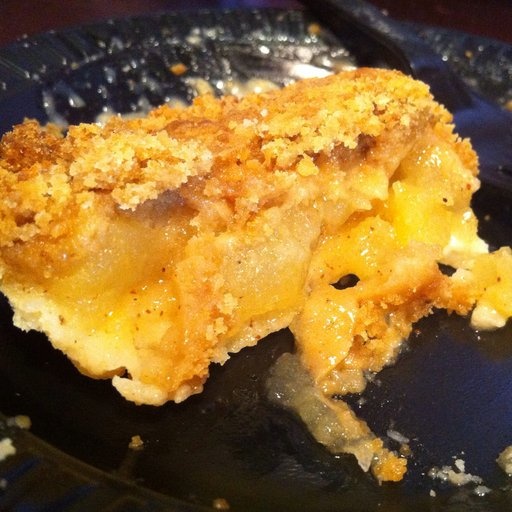} 
\includegraphics[width=0.15\textwidth]{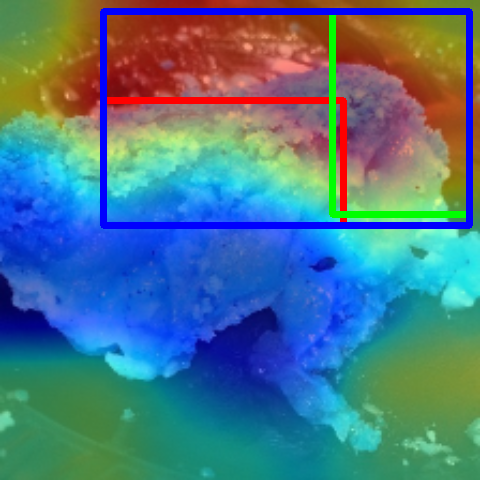}
}\hfill
\subfloat [Food-101: Ice-cream]
{\includegraphics[width=0.15\textwidth]{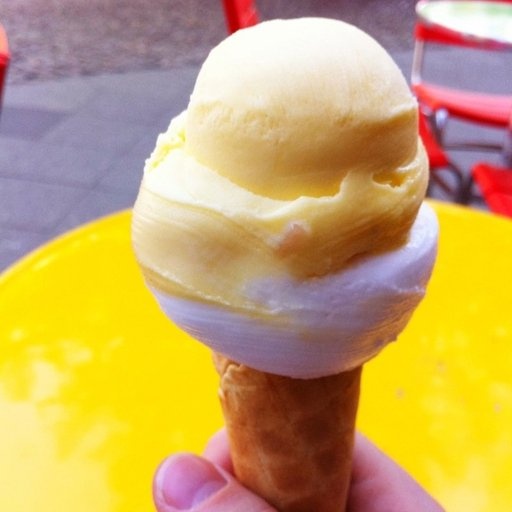}
\includegraphics[width=0.15\textwidth]{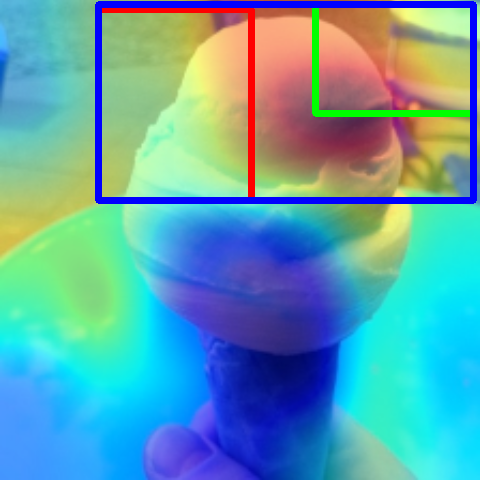}
}\hfill
\subfloat [Food-101: Strawberry shortcake]
{\includegraphics[width=0.15\textwidth]{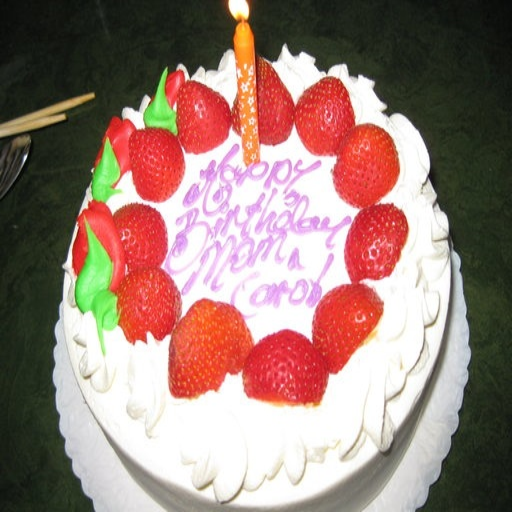}
\includegraphics[width=0.15\textwidth]{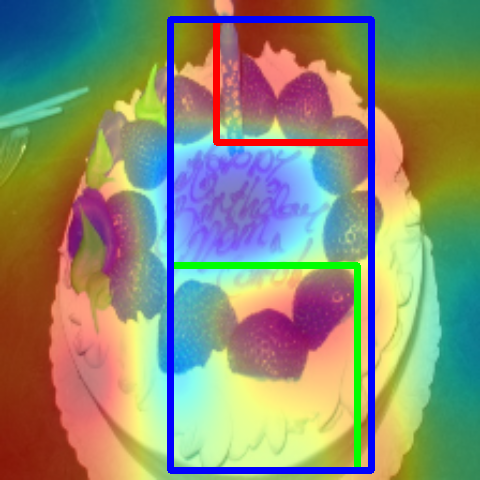} }
\\
\subfloat [AUC-V1: talking right] 
{\includegraphics[width=0.15\textwidth]{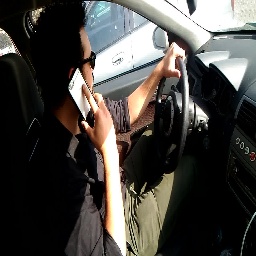}
\includegraphics[width=0.15\textwidth]{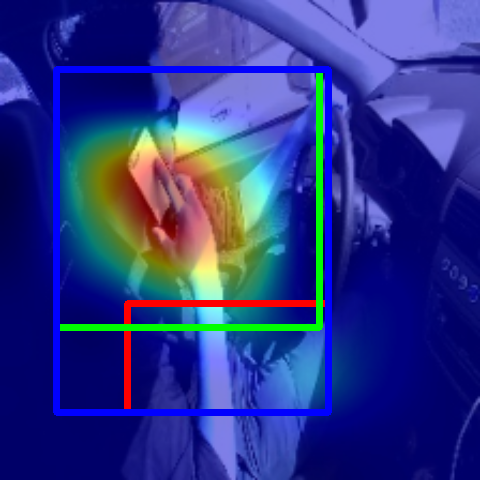}
}\hfill
\subfloat [AUC-V1: hair and makeup]
{\includegraphics[width=0.15\textwidth]{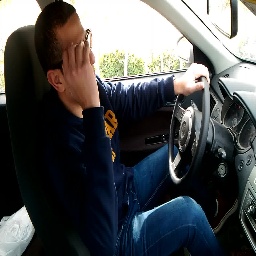}
\includegraphics[width=0.15\textwidth]{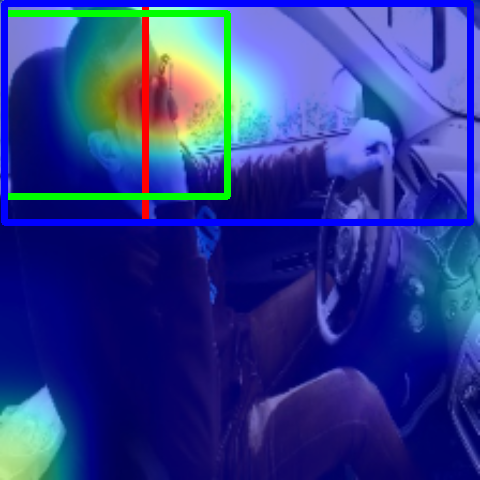}
}\hfill
\subfloat [AUC-V1: texting right]
{\includegraphics[width=0.15\textwidth]{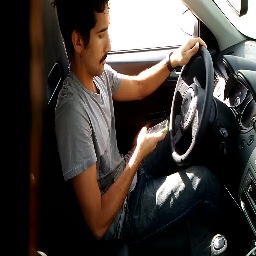}
\includegraphics[width=0.15\textwidth]{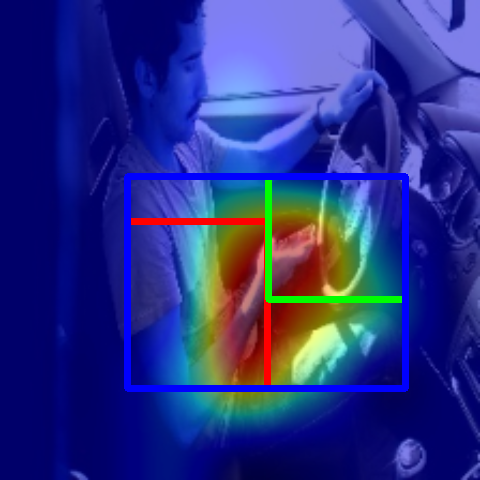}
}
\\  

 \subfloat[AUC-V1: texting-right]
 {{\includegraphics[width=0.15\linewidth]{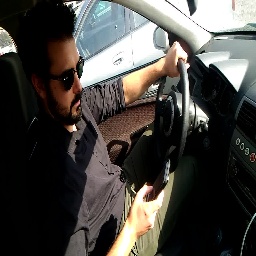} 
\includegraphics[width=0.15\textwidth]{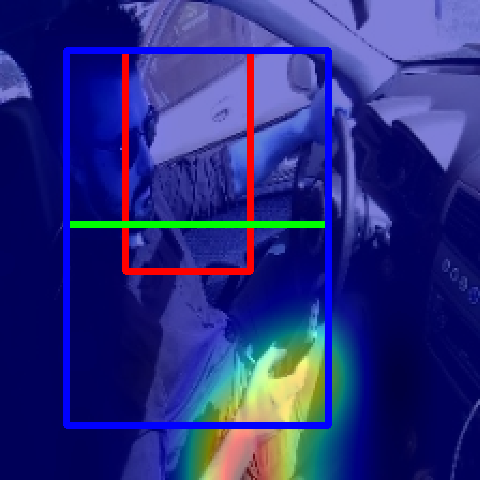}} \hspace{0.5 cm}\hfill
{\includegraphics[width=0.15\linewidth]{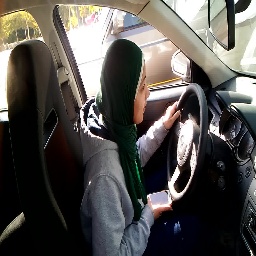}
\includegraphics[width=0.15\textwidth]{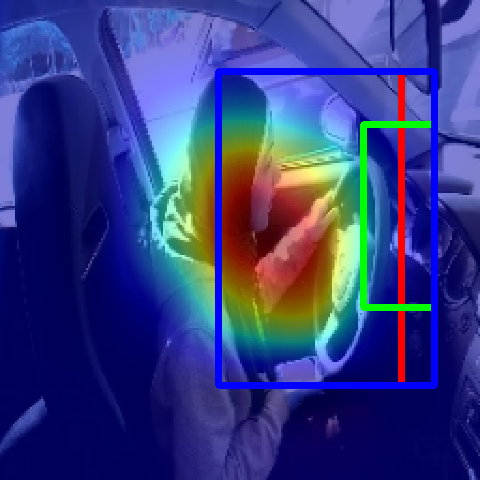}} \hspace{0.5 cm}\hfill
{\includegraphics[width=0.15\linewidth]{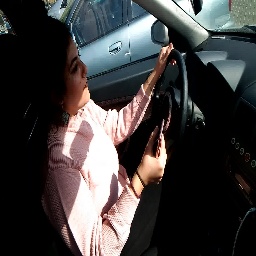}
\includegraphics[width=0.15\textwidth]{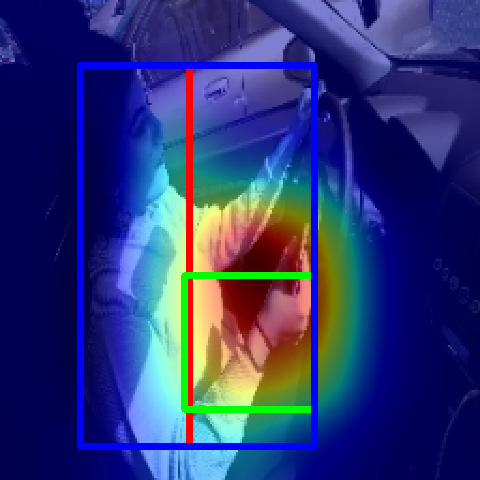}
}}\hspace{0.5 cm}\hfill
\\
\caption{ \textbf{Visualization of class activation maps} using the Gradient-weighted Class Activation Mapping (Grad-CAM) \cite{ selvaraju2017grad}. The top and middle rows illustrate \textbf{inter-class} variations with three different classes from two datasets: (a-c) Food-101 \cite{BossardGG14},  and (d-f) AUC-V1 \cite{AbouelnagaEM17}. The last row (g) represents the same action from AUC-V1 dataset. Each example contains the original image (left) and corresponding activation map of salient regions (right) on which we have overlaid only three SRs for clarity. It contains two primary SRs (enclosed with \textcolor{red}{red} and \textcolor{green}{green} bounding-boxes) along with a secondary region (enclosed with \textcolor{blue}{blue} bounding-box) which is derived from those two primary SRs. It is related to Fig.10 (Section V-F) in the paper. Best view in color. }
\label{fig:CAM_2}
\end{figure*}

\begin{figure*}[h]
\centering
\subfloat{\includegraphics[width=0.24\textwidth] {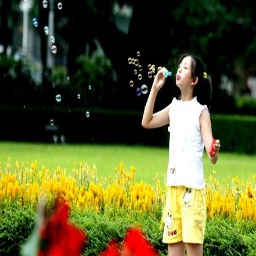} 
\includegraphics[width=0.24\textwidth]  {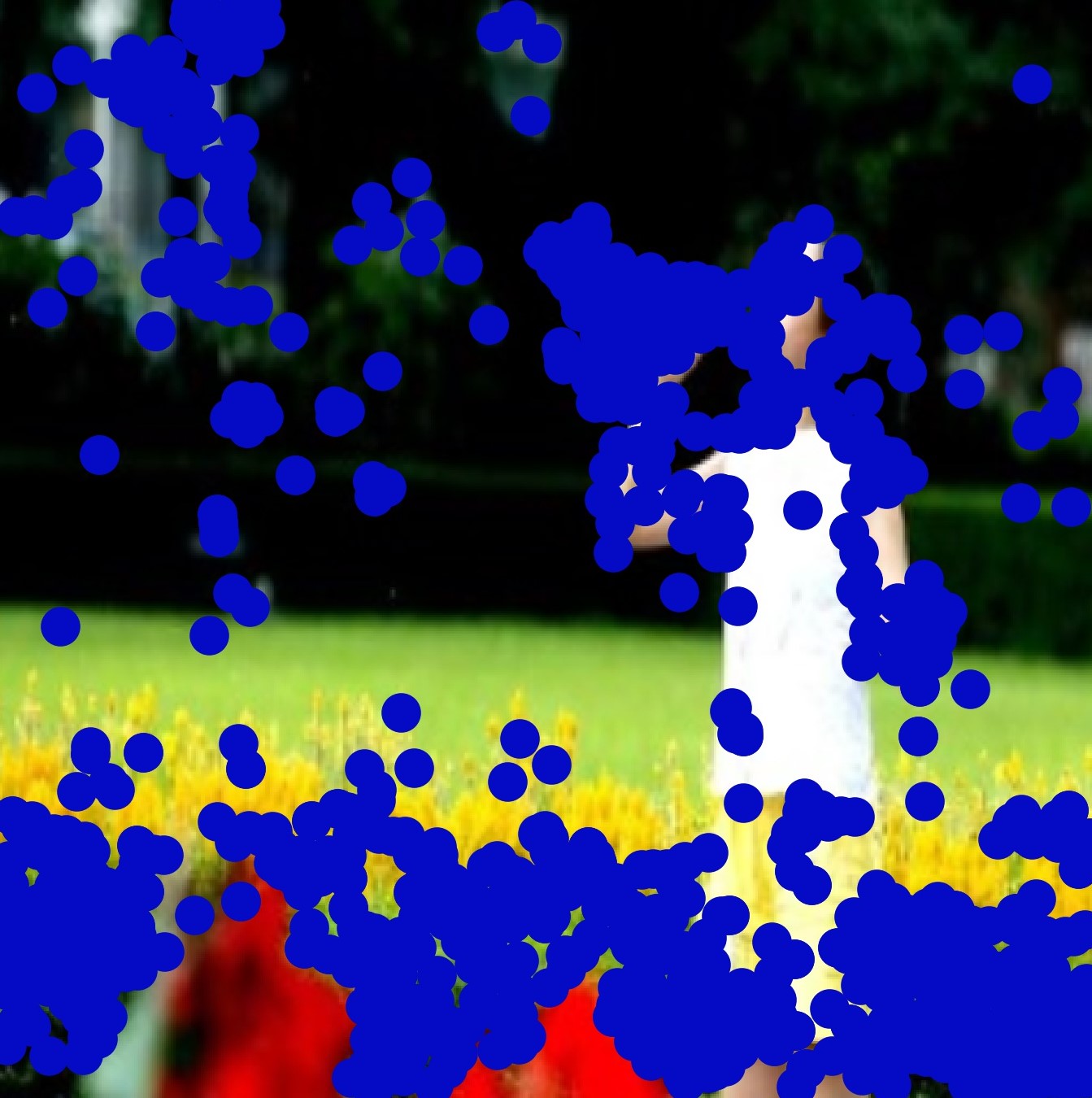}
\includegraphics[width=0.24\textwidth]  {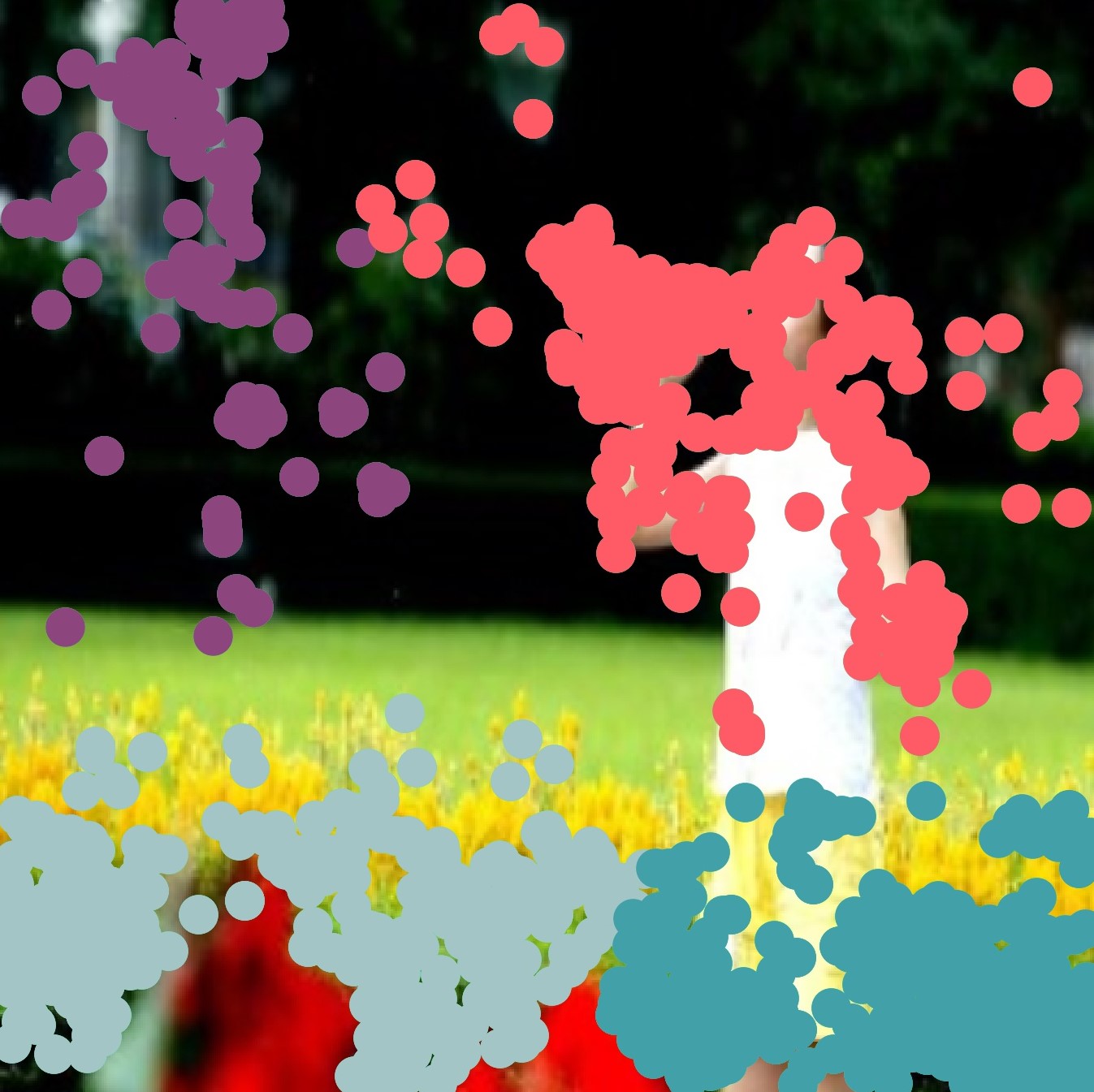}
\includegraphics[width=0.24\textwidth]  {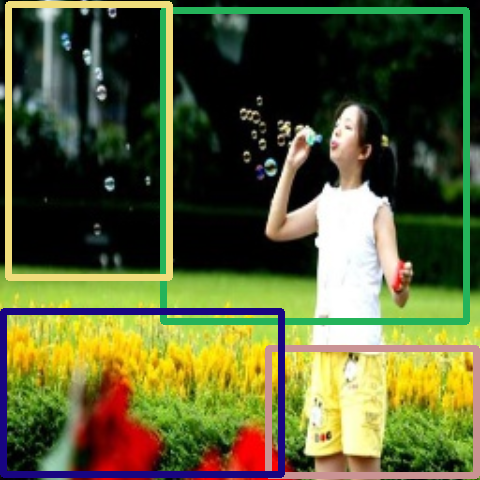}
}
\caption{Example images for blowing bubbles from Stanford-40 dataset \cite{YaoJKLGF11}. Using our keypoints-based clustering for detecting primary SRs. In this example, four primary semantic regions (SRs) are detected: original image $\Rightarrow$ detected SIFT keypoints $\Rightarrow$ clustered keypoints $\Rightarrow$ bounding boxes enclosing SRs (left to right). It is related to Fig.3 (Section III) in the paper. Best view in color.}
\label{fig:Primary_4SRs}
\end{figure*}
\begin{figure*}[h]
\centering
\subfloat{ \includegraphics[width=0.32\linewidth] {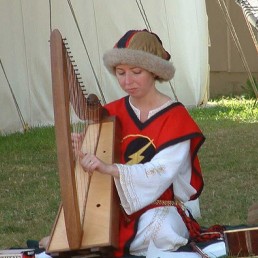}
\includegraphics[width=0.32\textwidth]  {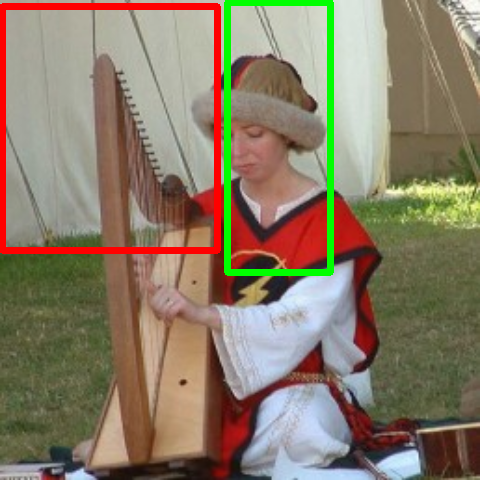} 
\includegraphics[width=0.32\textwidth]  {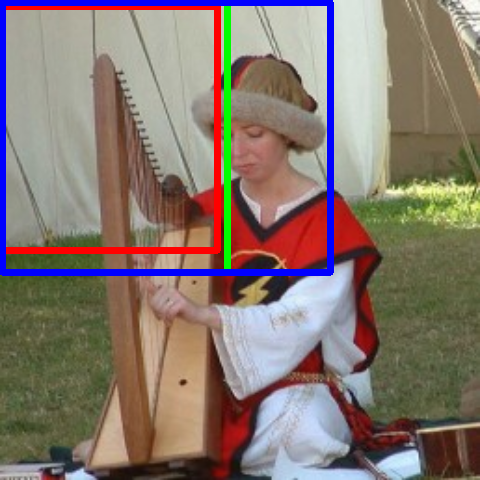}
}
\caption{Playing instrument Harp image example from PPMI-24 dataset \cite{YaoF10}. Detection of two primary SRs with which a secondary SR is generated. Original image $\Rightarrow$ detected two primary SRs (\textcolor{red}{red} and \textcolor{green}{green} bounding-boxes) $\Rightarrow$ Secondary SR (\textcolor{blue}{blue} bounding-box) with the primary SRs (left to right). The primary SRs describe a person (\textcolor{green}{green}) and Harp instrument (\textcolor{red}{red}) separately, which is combined in the secondary region to describe the action that a person is playing Harp. It is related to Fig.4 (Section III) in the paper. Best view in color.}
\label{fig:PR_SR}
\end{figure*}

\begin{figure*}
\centering
 \includegraphics[height=0.9\textheight] {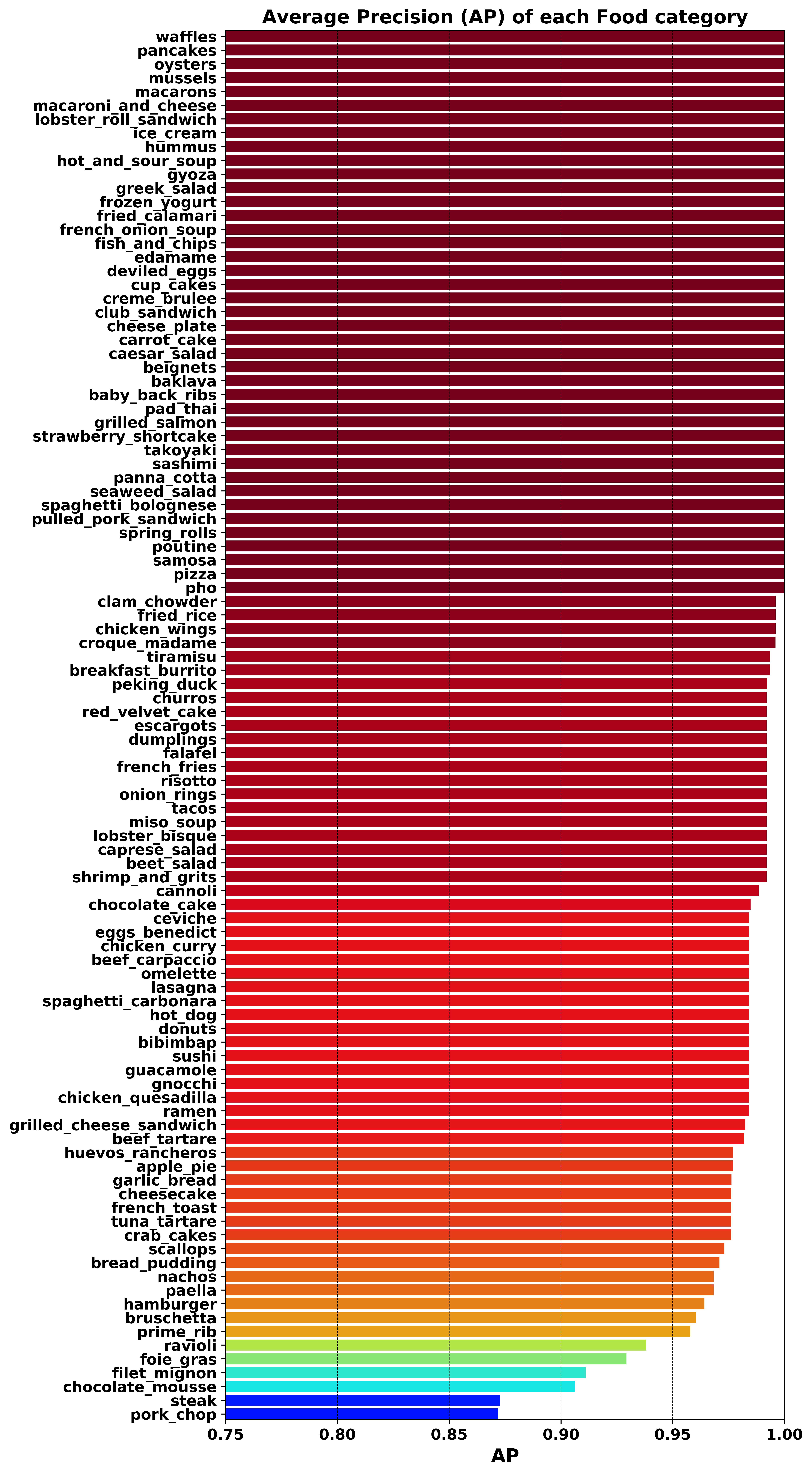} \caption{ \textbf{Average Precision (AP)} of each food category in Food-101 dataset \cite{BossardGG14} using our AG-Net. It is evident that the top 41 food classes are classified  with 100\% AP. The least AP is 87.18\% for the \emph{pork chop} food class. This has been described in Section V-C in the paper.} 
\label{fig:confusion matrix}
\end{figure*}

\begin{figure*}[h]
\centering
 \includegraphics[width=0.8\textwidth] {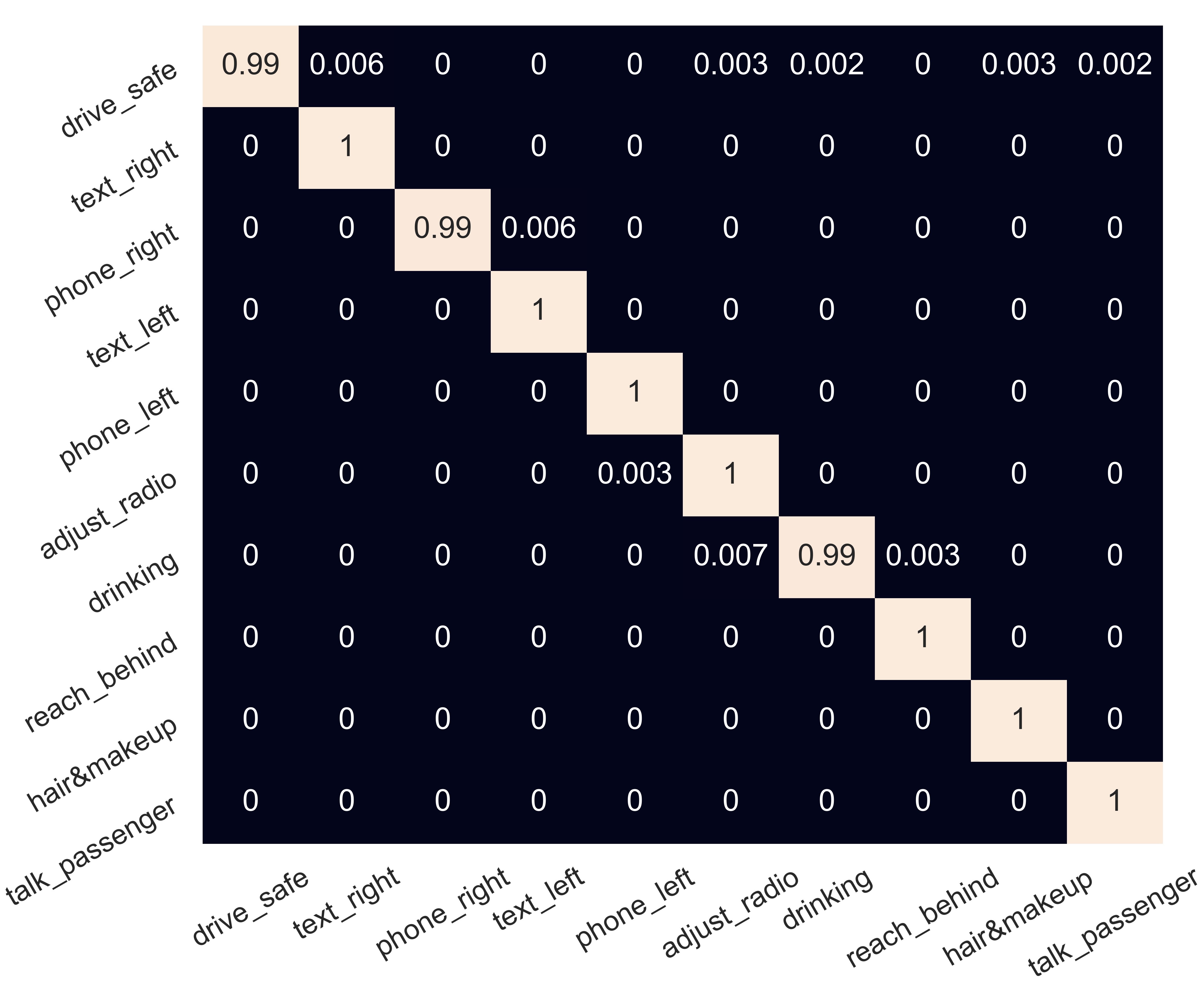} 
\caption{ \textbf{Confusion matrix} of the proposed AG-Net using AUC-V1 dataset \cite{AbouelnagaEM17}. This is related to Fig. 5a in the paper, in which class-wise accuracy has been presented. The x-axis represents predicted labels and y-axis denotes actual class labels of driving activities.
} 
\label{fig:CM_ V1}
\end{figure*}

\begin{figure*}
\centering
 \includegraphics[width=0.8\textwidth] {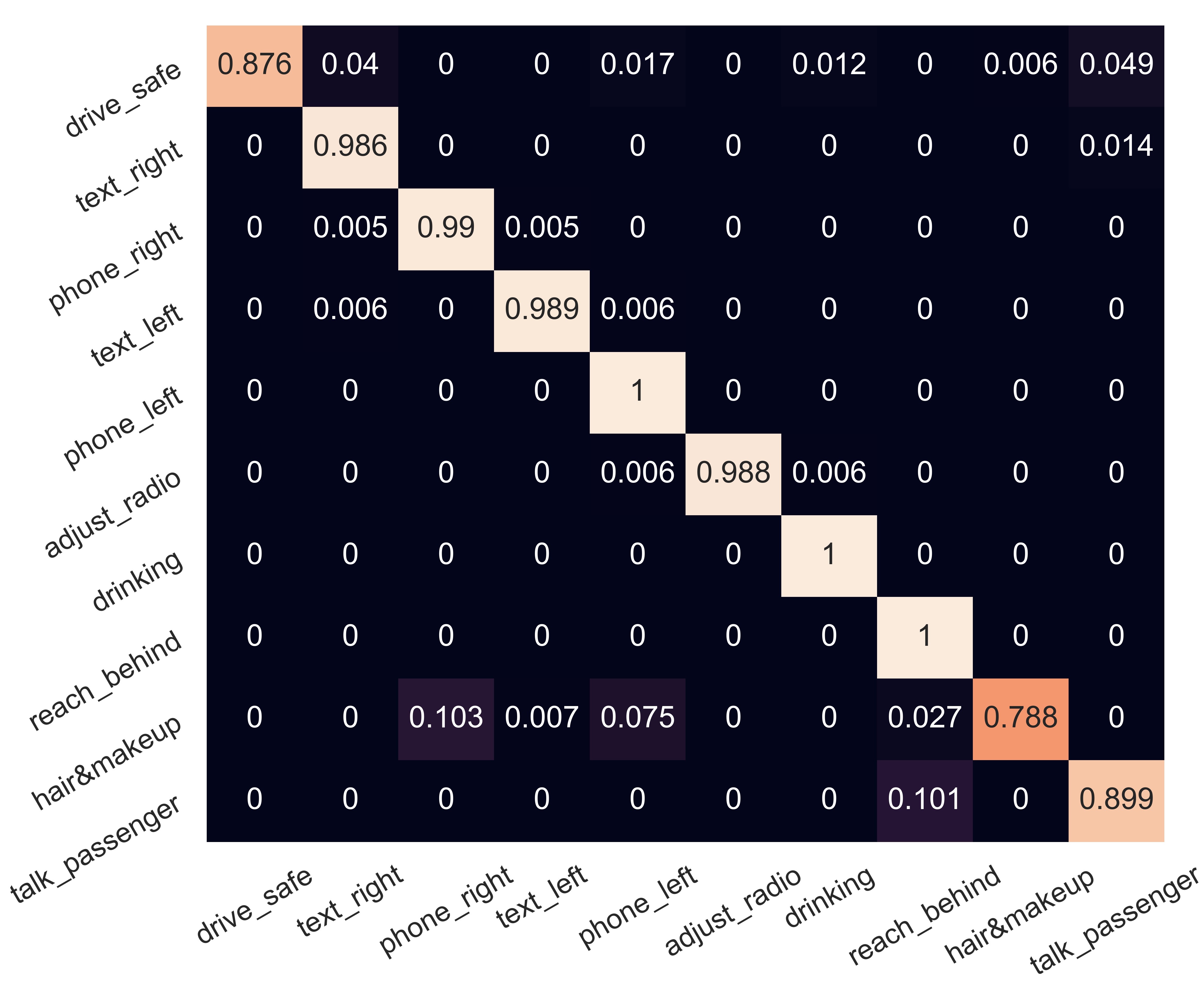}
\caption{\textbf{Confusion matrix} of the proposed AG-Net using AUC-V2 dataset which comprises of unique drivers-wise train-test split \cite{H.M.Eraqi}. This is related to Fig.5b in the paper, in which class-wise accuracy has been depicted. It is clear that \emph{C9: hair and makeup} and \emph{C0: driving safely} fine-grained activities are the low performer and is explained in Section V-A. 
} 
\label{fig:CM_V2}
\end{figure*}

\begin{figure*}
\centering
 \includegraphics[width=1.0\textwidth] {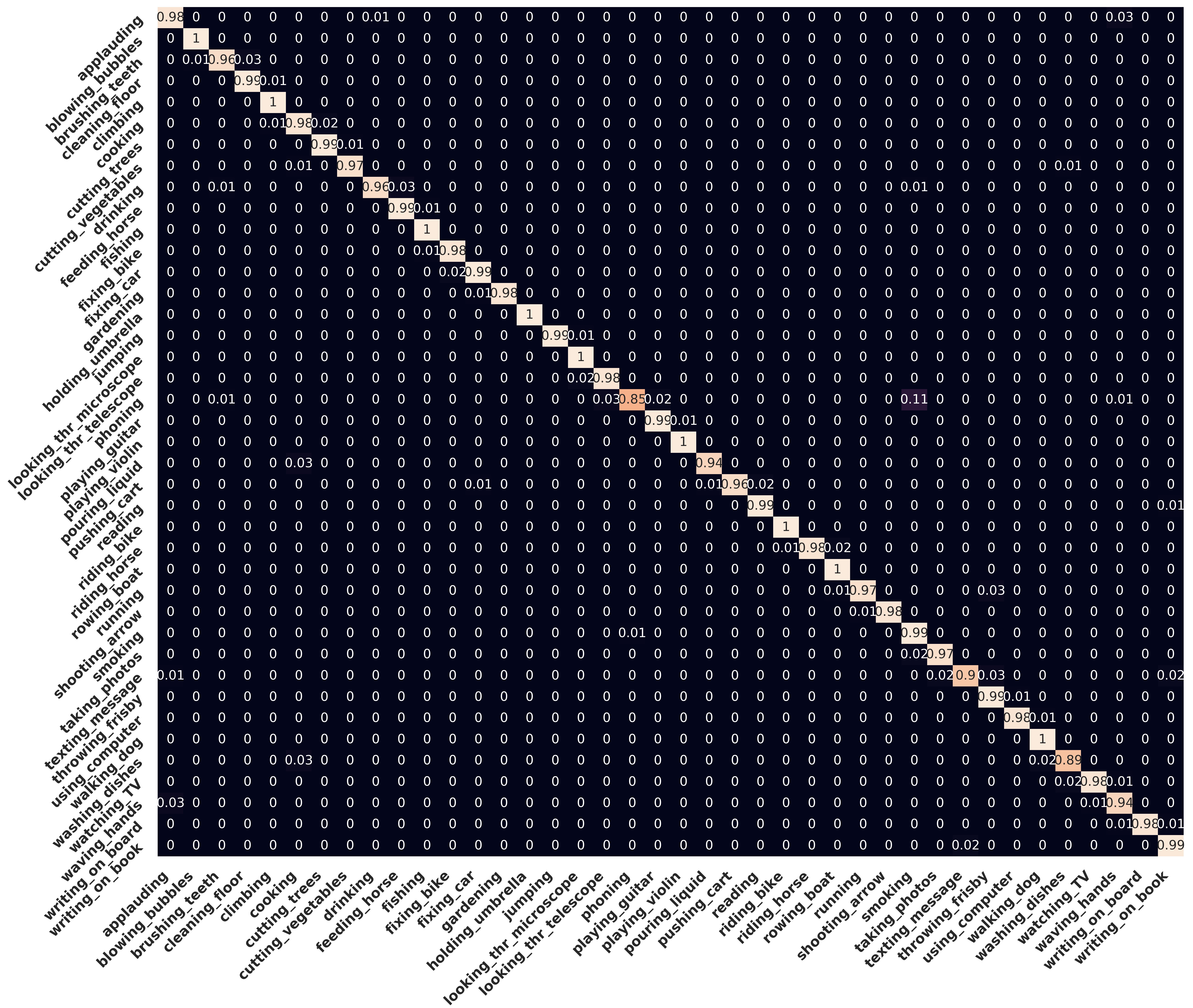}
\caption{\textbf{Confusion matrix} of the proposed AG-Net using Stanford-40 actions dataset \cite{YaoJKLGF11}. Nine actions offer 100\% classification accuracy and it is discussed in Section V-B in the paper. 
} 
\label{fig:CM_S_40}
\end{figure*}

\begin{figure*}
\centering
 \includegraphics[width=1.0\textwidth] {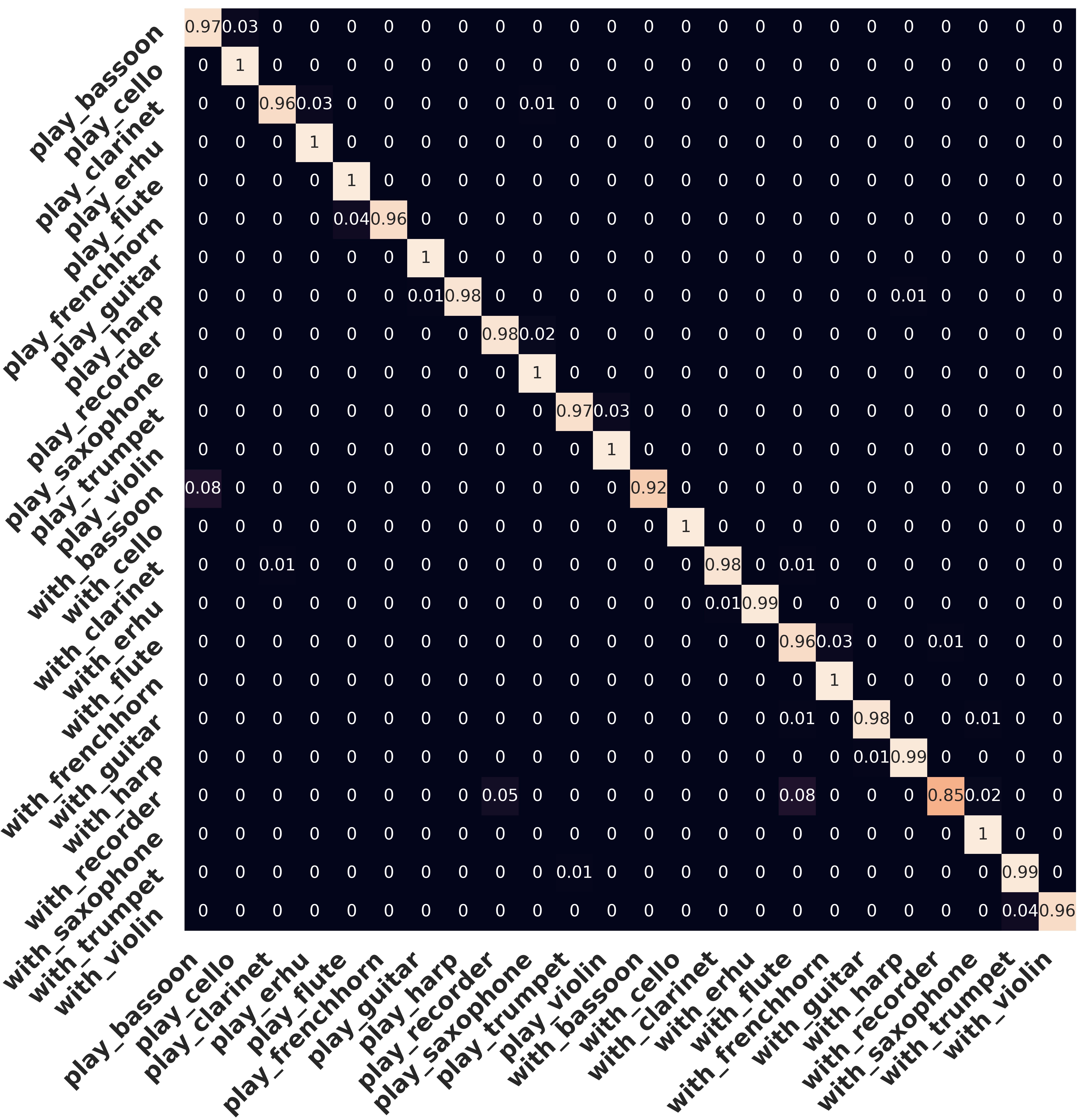}
\caption{ \textbf{Confusion matrix} of the proposed AG-Net using PPMI-24 dataset \cite{YaoF10}. It is clear that the predicted labels of nine human-instruments interactions are achieved with 100\% accuracy. The least accuracy (85\%) is achieved by \emph{with recorder} fine-grained interaction. It is presented in Section V-B in the paper.} 
\label{fig:CM_PPMI}
\end{figure*}

\end{document}